\documentclass[11pt]{amsart}

\usepackage{amsfonts,amssymb,amsmath}
\usepackage{graphicx}
\usepackage{booktabs}
\usepackage{subfigure}
\usepackage{multirow}
\usepackage{xcolor}
\definecolor{red}{RGB}{200,16,46}
\usepackage[colorlinks=true,
linkcolor=red,citecolor=red]{hyperref}
\usepackage{cleveref}
\usepackage{algorithm}
\usepackage[margin=1in]{geometry}
\usepackage{algorithmic}
\usepackage{siunitx}
\usepackage{listings}
\usepackage[inline]{enumitem}
\usepackage[foot]{amsaddr}
\setlist[enumerate]{leftmargin=.5in}
\setlist[itemize]{leftmargin=.5in}
\usepackage{amsopn}

\usepackage{listings}

\lstdefinestyle{bashstyle}{
  language=bash,
  basicstyle=\ttfamily\small,
  backgroundcolor=\color{gray!10},
  frame=single,
  rulecolor=\color{gray!40},
  showstringspaces=false,
  breaklines=true,
  keywordstyle=\color{blue!70!black},
  commentstyle=\color{green!50!black},
  stringstyle=\color{orange!60!black},
}

\usepackage{xcolor}
\usepackage[font=small,width=\textwidth]{caption}

\graphicspath{{figs}}

\newcommand{\defeq}{\ensuremath{\mathrel{\mathop:}=}}
\newcommand{\eqdef}{\ensuremath{=\mathrel{\mathop:}}}
\DeclareMathOperator*{\minopt}{minimize}

\newcommand{\decc}[1]{\num[round-mode=places,round-precision=3]{#1}}

\newcommand\sci[1]{\num[
scientific-notation=true,
round-precision=3,
fixed-exponent=1,
detect-weight=true,
detect-family=true,
round-mode=places,
retain-explicit-plus=true,
mode=text,fixed-exponent=0,
retain-explicit-plus=true,
output-exponent-marker=\text{e}]{#1}}

\title[VPreg: An Optimal Control Formulation for Diffeomorphic Image Registration]{VPreg: An Optimal Control Formulation for Diffeomorphic Image Registration based on the Variational Principle Grid Generation Method}

\author{Zicong Zhou${}^\dag$}
\address{$\dag$\,Department of Pharmacology and Neuroscience, University of North Texas Health Science Center, Fort Worth TX 76106, USA}
\email{zicong.zhou@unthsc.edu}
\thanks{This work was partly supported by the National Science Foundation (NSF) through the grants DMS-2009923 and DMS-2012825. Any opinions, findings, and conclusions, or recommendations expressed herein are those of the authors and do not necessarily reflect the views of the NSF}

\author{Baihan Zhao${}^\diamond$}
\address{$\diamond$\,Department of Mathematics, University of Texas at Arlington, Arlington TX 76019, USA}
\email{bxz6705@mavs.uta.edu}

\author{Andreas Mang${}^\ast$}
\address{$\ast$\,Department of Mathematics, University of Houston, Houston TX 77204, USA}
\email{andreas@math.uh.edu}

\author{Guojun Liao${}^\diamond$}
\email{liao@uta.edu}

\begin{document}

\begin{abstract}
This paper introduces VPreg, a novel diffeomorphic image registration method. This work provides several improvements to our past work on mesh generation and diffeomorphic image registration. VPreg aims to achieve excellent registration accuracy while controlling the quality of the registration transformations. It ensures a positive Jacobian determinant of the spacial transformation and provides an accurate approximation of the inverse of the registration map---a crucial property for many neuroimaging workflows. Unlike conventional methods, VPreg generates this inverse transformation within the group of diffeomorphisms rather than operating on the image space. The core of VPreg is a grid generation approach, referred to as \emph{Variational Principle} (VP), which constructs non-folding grids with prescribed Jacobian determinant and curl. These VP-generated grids guarantee diffeomorphic spatial transformations essential for computational anatomy and morphometry, and provide a more accurate inverse than existing methods. To assess the potential of the proposed approach, we conduct a performance analysis for 150 registrations of brain scans from the OASIS-1 dataset. Performance evaluation based on Dice scores for 35 regions of interest along with an empirical analysis of the properties of the computed spatial transformations demonstrates that VPreg outperforms state-of-the-art methods in terms of Dice scores, regularity properties of the computed transformation, and accuracy and consistency of the provided inverse map. We compare our results to ANTs-SyN, Freesurfer-Easyreg, and FSL-Fnirt.

\end{abstract}

\maketitle

\section{Introduction}\label{s:intro}

In the study of \emph{morphometry}~\cite{ashburner2000:voxel} or \emph{computational anatomy}~\cite{grenander1998:computational, thompson2002:framework}, we focus on measuring and quantifying geometric changes of anatomical structures resulting, e.g., from normal development, aging, or degenerative diseases. A fundamental tool that enables this analysis is diffeomorphic image registration. In image registration, we seek a geometric transformation $\pmb{\phi} : \Omega \to \Omega$, $\Omega \subset \mathbb{R}^d$, $d\in \{2,3\}$, that aligns points in one image with those in another~\cite{chen2023:image, fischer2008:illposed, modersitzki2003:numerical, modersitzki2009:fair,  sotiras2013:deformable}; more precisely, given two images---the (template or) moving image $M : \Omega \to \mathbb{R}$ and the (reference or) fixed image $F: \Omega \to \mathbb{R}$---we seek a mapping $\pmb{\phi}$ such that $(M \circ \pmb{\phi})(\pmb{\omega}) \approx F(\pmb{\omega})$ for all $\pmb{\omega} \in \Omega$. Image registration is a non-linear, ill-posed inverse problem that poses significant mathematical and computational challenges~\cite{fischer2008:illposed}. In \emph{diffeomorphic} image registration we restrict geometric transformations to the set of diffeomorphisms $\operatorname{Diff}(\Omega)$ from $\Omega$ to $\Omega$, i.e., bijections with a smooth inverse~\cite{younes2010:shapes}. As we will see, this is a key requirement in computational anatomy. Popular software packages for diffeomorphic registration include Demons~\cite{mansi2011:ilogdemons,vercauteren2008:symmetric,vercauteren2009:diffeo}, ANTs~\cite{avants2011:reproducible, avants2008:symmetric}, Deformetrica~\cite{bone2018:deformetrica}, DARTEL~\cite{ashburner2007:fast}, Fnirt~\cite{andersson2007:non, jenkinson2012:fsl}, or EasyReg~\cite{hoffmann2021:synthmorph, iglesias2023:ready}.

In cohort studies, registration allows us to align the images in a common reference frame and perform statistical analysis to, e.g., identify biomarkers associated with a disease. Techniques such as \emph{voxel-based morphometry} ({\bf VBM})~\cite{mechelli2005:voxel, antonopoulos2023:systematic, ashburner2000:voxel}, \emph{deformation-based morphometry} ({\bf DBM})~\cite{chung2001:unified, gaser2001:deformation}, and \emph{tensor-based morphometry} ({\bf TBM})~\cite{hua2008:tensor, leow2006:longitudinal, koikkalainen2011:multi, lepore2008:generalized} have been widely adopted in the neuroimaging community to discover and aid understanding of the differences among patient populations. Here, we are given a set of images $\{M_{i} : \Omega \to \mathbb{R}\}_{i=1}^n$ of the brain anatomy and a pre-selected reference image $F$ (typically referred to as the \emph{atlas image}; for instance the MNI152/MNI305 template~\cite{evans1993:3d}). We find spatial transformations $\{\pmb{\phi}_{i} : \Omega \to \Omega\}_{i=1, \ldots, n}$ such that $M_{i}\circ\pmb{\phi}_i \approx F$ for all $i = 1,\ldots,n$. VBM constitutes the statistical analysis conducted on the registered images $\{M_{i} \circ \pmb{\phi}_i)\}_{i=1, \ldots, n}$, whereas DBM is the statistical analysis conducted on the spatial transformations $\{\pmb{\phi}_{i}\}_{i=1}^n$. Likewise, TBM extends conceptual ideas underpinning DBM to longitudinal data of individual patients. TBM has been instrumental in detecting subtle neurological changes~\cite{hua2008:tensor, shi2015:studying, vemuri2015:accelerated, hua2008:3d}, which can be crucial for diagnosing the early stages of certain diseases~\cite{zhou2025:macro} and their pathological developments~\cite{dennis2016:tensor}. This is for example enabled by statistical analysis on the \emph{Jacobian determinants} ({\bf JD}) of the spatial transformations $\{\pmb{\phi}_{i}\}_{i=1}^n$~\cite{lepore2008:generalized}. In addition, many of these studies require not only the forward maps $\pmb{\phi}$ by their inverses $\pmb{\phi}^{-1}$ to relate information back to either the atlas space or the patient space, respectively. As a consequence, $\pmb{\phi}$ has to be diffeomorphic. We elaborate more below.

Thanks to recent advances in computational mathematics, large-scale studies such as ENIGMA~\cite{thompson2020:enigma} and the Human Connectome Project~\cite{elam2021:human} have been made possible. Our work intends to provide a novel framework to aid these efforts. We hypothesize that \emph{controlling the properties of the computed diffeomorphisms and its inverse opens up new avenues for computational anatomy}.

\subsection{Outline of the Method}

In the present work, we formulate diffeomorphic image registration as a partial differential equation ({\bf PDE}) constrained optimization problem. We minimize the distance between the deformed moving image $M \circ \pmb{\phi}$ and the fixed image $F$ whilst controlling the map $\pmb{\phi}$ by introducing hard constraints on the determinant of the Jacobian $\det \nabla \pmb{\phi}$ and $\nabla \times \pmb{\phi}$. Our work is motivated by a variational grid generation method. Given an arbitrary target mesh $\pmb{\phi}_t$, this method allows us to generate the ``closest'' non-folding mesh $\pmb{\phi}$ with prescribed Jacobian determinant and curl~\cite{zhou2023:vpgrid}. We will show that our formulation for diffeomorphic image registration yields well behaved diffeomorphic maps $\pmb{\phi}$, provides excellent registration accuracy, and avoids the solution of complicated space-time problems. The proposed framework integrates a novel, effective reformulation of our grid generation method to also provide the inverse of the computed registration map $\pmb{\phi}$. We show empirically that these inverse maps are much more consistent with the forward map than those generated by other methods.

\subsection{Related Work}

In the present work we follow up on our past work on diffeomorphic image registration~\cite{zhou2022:recent} and grid generation~\cite{zhou2023:vpgrid, zhou2024:construction}. Traditionally, deformable image registration is formulated as a variational optimization problem akin to many traditional inverse problem formulations: We balance \begin{enumerate*}[label={\it(\alph*)}] \item a functional measuring the discrepancy between the model prediction (in our case the transformed moving image $M \circ \pmb{\phi}$) and a reference dataset (in our case the fixed image $F$) and \item a regularization functional to address the ill-posedness of the problem\end{enumerate*}. Formally, given input images $M, F \in \mathcal{I}$, $\mathcal{I} \subset \{ I : \Omega \to \mathbb{R}\}$, we seek a spatial transformation $\pmb{\phi} \in \text{Map}(\Omega)$, $\text{Map}(\Omega) \subset \{\pmb{\psi} : \Omega \to \Omega\}$, that satisfies
\begin{equation}\label{e:imgreg}
	\minopt_{\pmb{\phi}\, \in\, \text{Map}(\Omega)}\,\, \mathcal{D}(M,F,\pmb{\phi}) + \alpha\mathcal{R}(\pmb{\phi}).
\end{equation}

\noindent Here, $\mathcal{D} : \mathcal{I} \times \mathcal{I} \times \text{Map}(\Omega) \to \mathbb{R}$ measures the proximity between $M \circ \pmb{\phi}$ and $F$. This distance can be defined in various ways. Examples include the Mean Squared Error ({\bf MSE}), Mutual Information ({\bf MI}), or Normalized Cross-Correlation ({\bf NCC})~\cite{modersitzki2009:fair}. The second term $\mathcal{R} :  \text{Map}(\Omega) \to \mathbb{R}$ is a regularization model that prescribes desirable properties for $\pmb{\phi}$. These traditionally include norms stipulating smoothness requirements on $\pmb{\phi}$~\cite{fischer2003:curvature, fischer2002:fast} or are based on physical principles~\cite{christensen1997:volumetric, crum2005:anisotropic, burger2013:hyperelastic}.

A key concern of our work is to generate diffeomorphic transformations $\pmb{\phi}$. The need to generate diffeomorphisms arises from the applications we target with our work. For the morphometry studies outlined above, the features we work with need to make sense. If we study anatomical changes over time or across subjects based on measures derived from the computed maps $\pmb{\phi}$, these maps should not introduce folding or singularities. In addition, in many cohort studies we need to bring information from the atlas space to the patient space and vice versa (e.g., for atlas based segmentation or morphometry); this requires access to the inverse of $\pmb{\phi}$; for the inverse to be meaningful, $\pmb{\phi}$ has to be a diffeomorphism.

In general, the model outlined above does not guarantee that the computed maps are diffeomorphisms (with a few exceptions; e.g.,~\cite{burger2013:hyperelastic}). One strategy to safeguard against non-diffeomorphic maps $\pmb{\phi}$ is to add hard and/or soft constraints to the variational problem~\cite{burger2013:hyperelastic, haber2007:image, rohlfing2003:volume, haber2004:numerical}. We follow a similar approach in the present work. An alternative strategy founded on principles in Riemannian geometry is to introduce a pseudo-time variable $t$ and parameterize the sought after map $\pmb{\phi}$ in terms of a smooth, time-dependent or stationary velocity field $\pmb{v}$~\cite{younes2020:subriemannian, trouve1995:infinite, dupuis1998:variational, miller2001:group, younes2007:jacobi, vercauteren2008:symmetric}. This led to various approaches based on optimal control formulations governed by ordinary ({\bf ODE}s)~\cite{beg2005:computing, trouve1995:infinite, dupuis1998:variational} or PDEs~\cite{mang2015:inexact, mang2016:constrained, mang2017:lagrangian, mang2017:semilagrangian, hart2009:optimal, mang2024:claire}. Introducing time-dependent dynamics poses significant computational challenges, especially in cases where these are modeled by PDEs; to make these methods computationally tractable and useful in practice, one needs to design effective numerical methods~\cite{mang2015:inexact, mang2017:lagrangian, mang2017:semilagrangian} and/or deploy them on dedicated hardware~\cite{mang2016:distributed, brunn2020:multinode, mang2024:claire, brunn2021:claire, brunn2021:fast, mang2019:claire}. In the present work, we avoid these challenges by formulating the problem as a PDE-constrained optimization problem that does not involve any time-dependent dynamics. This significantly reduces the computational burden; we only require elliptic solves that can be implemented effectively using spectral methods.

Likewise to many existing PDE-constrained formulations for diffeomorphic image registration, we use an \emph{optimize-then-discretize} approach to derive the optimality conditions~\cite{mang2015:inexact, mang2016:constrained, mang2017:semilagrangian, hart2009:optimal, mang2024:claire}. Methods that rely on automatic differentation for optimization can for instance be found in~\cite{bone2018:deformetrica, hartman2023:elastic, franccois2021:metamorphic}. Moreover, the recent success of machine learning in numerous scientific disciplines has led to the emergence of methods that replace traditional variational approaches by learning~\cite{chen2023:image, amiri2025:physics, shen2021:accurate, hoffmann2021:synthmorph, iglesias2023:ready, krebs2019:learning}.

A key aspect of our work is to generate diffeomorphic maps $\pmb{\phi}$ that have prescribed properties. In particular, our formulation controls the determinant of the Jacobian and the curl of $\pmb{\phi}$. The work that is most closely related to ours is~\cite{lee2011:analysis}. Works that control the determinant of the Jacobian include~\cite{burger2013:hyperelastic, haber2007:image, rohlfing2003:volume, haber2004:numerical}; other models that control the Jacobian are based on incompressible~\cite{mang2015:inexact,  ruhnau2007:optical, chen2011:image, hinkle2009:4dmap, mansi2011:ilogdemons} or near-incompressible~\cite{mang2016:constrained, mang2019:claire, mang2024:claire} flows.

Lastly, as outlined above, one key requirement is to generate maps that can take us from the moving image $M$ to the fixed image $F$ and vice versa, i.e., we have access to $\pmb{\phi}$ that takes us from $M$ to $F$ and its inverse $\pmb{\phi}^{-1}$. One way of accomplishing this is to formulate the problem such that it is invariant to permutations of the input image, i.e., the map that takes us from $M$ to $F$ does not change if we swap $M$ and $F$. This is referred to as symmetric or inverse consistent image registration~\cite{leow2005:inverse, christensen2002:consistent, avants2011:reproducible,liao2025:icondiffnet, avants2008:symmetric, vercauteren2008:symmetric}. This can be accomplished in various ways; for instance we can jointly estimate $\pmb{\phi}$ and $\pmb{\phi}^{-1}$. In our approach, we go a different route and compute an approximation to $\pmb{\phi}^{-1}$ in a post-processing step.

\subsection{Contributions}

In the present work, we follow up on our past contributions in diffeomorphic image registration~\cite{zhou2022:recent} and grid generation~\cite{zhou2023:vpgrid, zhou2024:construction}.

The main contributions of the present work are
\begin{enumerate}
\item We introduce a novel variational approach for diffeomorphic image registration that allows us to precisely control the properties of the computed diffeomorphic transformations $\pmb{\phi}$.
\item Compared to our past work~\cite{zhou2022:recent, zhou2024:construction}, we introduce reformulations of the problem that yield more effective numerical algorithms. We also provide additional algorithmic details underpinning our past work.
\item Our framework for diffeomorphic image registration integrates a grid generation method that allows us to effectively compute approximations to the inverse of the spatial transformations $\pmb{\phi}$ that maps the moving image $M$ to the fixed image $F$.
\item We provide a detailed evaluation of our method and compare its performance against several prominent packages for diffeomorphic image registration. Our results indicate that our framework yields well behaved diffeomorphisms with precise control without sacrificing registration accuracy. We also report results that empirically show that the inverse maps we compute are more consistent with the our forward map than is the case for existing methods.
\end{enumerate}

We coin our framework for diffeomorphic image registration VPreg. The codes of our algorithms are going to be released at \url{https://github.com/zicongzhou818} after acceptance of this article.

\subsection{Outline of the Paper}
The rest of the paper is organized as follows. In \Cref{s:methods}, we present the problem formulation and our numerical approach. This includes a recapitulation of the foundational method for mesh generation that underpins our work (\Cref{s:vp}). Based on this variational formulation, we present a related problem formulation for diffeomorphic image registration (\Cref{s:vpcontrol}). We follow up with reformulations of the diffeomorphic registration problem (\Cref{s:vpregel}) and the grid generation method (\Cref{s:lmvp}) that yield more effective numerical methods. In \Cref{s:results}, we provide experiments. In \Cref{s:conclusion}, we conclude this paper. We provide theoretical derivations and details in \Cref{s:append}.

\section{Methods}\label{s:methods}

In this section, we introduce the overall methodology. Our work is based on an approach we coined the ``Variational Principle ({\bf VP}) Grid Generation Method'' \cite{zhou2023:vpgrid}. We revisit this problem formulation in \Cref{s:vp}. Based on the underlying principles, we have developed an optimal control formulation for diffeomorphic image registration in~\cite{zhou2022:recent}. We outline this approach in \Cref{s:vpcontrol}. We introduce a reformulation of the problem in \Cref{s:vpregel}. This reformulation allows us to significantly reduce the computational complexity. Subsequently, \Cref{s:lmvp} introduces a novel approach to compute the inverse of the spatial transformation found during the registration using~\cite{zhou2024:construction}. We conclude this exposition by presenting our overall framework that combines the latter two building blocks into an optimal control formulation for diffeomorphic image registration that allows us to precisely control the curl and the determinant of the Jacobian of the spatial transformation and its inverse (see \Cref{s:vpreg}).

\subsection{The Variational Principle (VP) Grid Generation Method}\label{s:vp}

Here, we are going to recapitulate the grid generation method described in~\cite{zhou2022:recent}. This method forms the basis of our diffeomorphic image registration formulation. For simplicity of presentation, we limit the exposition to $d=3$; the case for $d=2$ follows similar arguments. Let the simply-connected and bounded set $\Omega \subset \mathbb{R}^3$ denote the domain and let $\pmb{\omega}=(x,y,z)\in\Omega$. Let $H^{2}_{0}(\Omega)^3$ denote the Sobolev space of vector-valued $H^2$-functions defined on $\Omega$ that vanish on $\partial\Omega$. Moreover, let $\pmb{\phi} : \Omega \to \Omega$ denote a mapping from $\Omega$ to $\Omega$, let $\pmb{u} : \Omega \to \mathbb{R}^3$ denote a displacement field, and let $\pmb{\operatorname{id}}_{\Omega}(\pmb{\omega}) = \pmb{\omega}$ denote the identity map. We define the class of transformations we seek to compute as elements of the set
\begin{equation}\label{e:maps}
\operatorname{Map}(\Omega)
\defeq \left\{\pmb{\phi}: \Omega \to \Omega \mid  \pmb{\phi}=\pmb{\omega} + \pmb{u}(\pmb{\omega}) \text{ where }\pmb{u}(\pmb{\omega})\in H^{2}_{0}(\Omega)^3\right\}.
\end{equation}

\noindent Suppose we are given a strictly positive scalar function $f_t : \Omega \to \mathbb{R}_{++}$ and a vector-valued function $\pmb{g}_t : \Omega \to \mathbb{R}^3$ satisfying
\begin{equation}\label{e:fg}
\int_{\Omega} f_t(\pmb{\omega})\,\text{d}\pmb{\omega} = |\Omega|
\quad\text{and}\quad
\nabla \cdot  \pmb{g}_{t}(\pmb{\omega})= 0,
\end{equation}

\noindent respectively. These conditions will force the target functions $f_t$ to behave like the JD and $\pmb{g}_t$ like curl.

Let $f_t$ and $\pmb{g}_t$ be the prescribed JD and curl. We define the squared $L^2$-distance
\begin{equation}\label{e:ssd}
\frac{1}{2}\int_{\Omega} (\det\nabla\pmb{\phi} - f_t)^2+\|\nabla \times\pmb{\phi}- \pmb{g}_t\|^2 \,\mathrm{d}\pmb{\omega}.
\end{equation}

\noindent to measure the proximity between $\det\nabla\pmb{\phi}$ and $f_t$ and $\nabla \times\pmb{\phi}$ and $\pmb{g}_t$, respectively.

Suppose we are given an initial diffeomorphism $\pmb{\phi}_{o}\in \operatorname{Map}(\Omega)$. In the grid generation method, we seek a diffeomorphism $\pmb{\phi} \in \operatorname{Map}(\Omega)$ of the form $\pmb{\phi}=\pmb{\phi}_{m} \circ \pmb{\phi}_{o}=\pmb{\phi}_{m}(\pmb{\phi}_{o})$, where an intermediate transformation $\pmb{\phi}_{m}$ left-translates the given $\pmb{\phi}_{o}$ to $\pmb{\phi}$. We assume $\pmb{\phi}_m$ is a perturbation of the identity, i.e., we model it as $\pmb{\phi}_m(\pmb{\omega})=\pmb{\operatorname{id}}_{\Omega}(\pmb{\omega})+\pmb{u}(\pmb{\omega})$ with displacement vector field $\pmb{u} \in H^2_0(\Omega)^3$ that minimizes the distance in \cref{e:ssd} with control functions $f$ for JD and $\pmb{g}$ for curl, respectively. This problem can be formulated as a variational PDE-constrained optimization problem of the form
\begin{equation}\label{e:ssdcont}
\begin{aligned}
    \minopt_{\pmb{\phi}_{m}, \,\,f,\,\, \pmb{g}}\;\;
    &\frac{1}{2}\int_{\Omega}
    (\det\nabla\pmb{\phi} - f_t)^2+\|\nabla \times\pmb{\phi}- \pmb{g}_t\|^2\,\text{d}\pmb{\omega}\\
    \begin{aligned}
    \text{subject to} \\{}\\
    \end{aligned}
    \;\;&
    \begin{aligned}
        \nabla \cdot  \pmb{\phi}_{m} &= f+2  && \text{in}\;\; \Omega,\\
        \nabla \times \pmb{\phi}_{m} &= \pmb{g}&& \text{in}\;\; \Omega,
    \end{aligned}
\end{aligned}
\end{equation}

\noindent where $\pmb{\phi} = \pmb{\phi}_m \circ \pmb{\phi}_{o}$ and $\pmb{\phi}_{m} = \pmb{\operatorname{id}}_{\Omega}$ on $\partial\Omega$. We explain the particular form of the constraints in \cref{e:ssdcont} in \Cref{s:explanation-constraints}.

Using the identity $\nabla \times (\nabla \times \pmb{\phi}_{m}) = \nabla(\nabla \cdot \pmb{\phi}_m) - \Delta \pmb{\phi}_m$ we can eliminate the first constraint to obtain
\begin{equation}\label{e:ssdC}
\Delta \pmb{\phi}_{m} =\nabla f-\nabla \times \pmb{g} \eqdef \pmb{C}(f,\pmb{g}) \text{ in } \Omega,
\end{equation}

\noindent where $\pmb{C}$ is the control function accountable for both JD and curl that is derived from the divergence-curl system using first-order derivatives. Notice that we control $\pmb{\phi}_{m}$ instead of $\pmb{\phi}$. We refer to \cite{zhou2022:recent} for the derivation of the necessary optimality conditions associated with \cref{e:ssd}.

The solution pool of VP is the set of diffeomorphic transformations that lies in $\operatorname{Map}(\Omega)$ defined in~\cref{e:maps}; we define
\begin{equation}\label{e:sol}
\operatorname{Sol}(\Omega) \defeq \operatorname{Diff}(\Omega) \cap \operatorname{Map}(\Omega).
\end{equation}

\subsubsection{Justification of the Constraints}\label{s:explanation-constraints}
Let $\pmb{\phi}\in\operatorname{Sol}(\Omega)$ with $\pmb{\phi}(\pmb{\omega}) = \pmb{\operatorname{id}}_{\Omega}(\pmb{\omega})+\pmb{u}(\pmb{\omega})$, $\pmb{u}\in H^{2}_{0}(\Omega)^3$, $\pmb{u}(\pmb{\omega})=(u_{1}(\pmb{\omega}),u_{2}(\pmb{\omega}),u_{3}(\pmb{\omega}))$, $\pmb{\omega}=(x,y,z)\in\Omega$, denote a \emph{given} transformation. We have $\nabla\cdot\pmb{\phi} = 3 + \nabla\cdot\pmb{u}$ and $\det\nabla\pmb{\phi} = 1 + \nabla\cdot\pmb{u} + \det\nabla\pmb{u} + \psi(\pmb{u})$ with tail term $\psi: \mathbb{R}^3 \to \mathbb{R}$,
\[
\psi(\pmb{u})
= u_{1x}u_{2y}u_{3z} + u_{1z}u_{2x}u_{3y} + u_{1y}u_{2z}u_{3x}
- u_{1x}u_{2z}u_{3y} - u_{1y}u_{2x}u_{3z} - u_{1z}u_{2y}u_{3x}.
\]

\noindent Consequently,
\begin{equation}\label{e:div}
\nabla\cdot\pmb{\phi}= \det\nabla\pmb{\phi} + 2 - \det\nabla\pmb{u} - \psi(\pmb{u}).
\end{equation}

\noindent Assuming that the displacement is small, we can neglect the nonlinear, high-order terms in $\pmb{u}$. That is, suppose $\pmb{u} = \epsilon \pmb{\tilde{u}}$ with $\epsilon >0$ small, then
\[
\det\nabla\pmb{\phi}= 1+\nabla\cdot(\epsilon\pmb{\tilde{u}})+\det\nabla(\epsilon\pmb{\tilde{u}})+\psi(\epsilon\pmb{\tilde{u}}),
\]

\noindent where $\nabla\cdot(\epsilon\pmb{\tilde{u}}) = \mathcal{O}(\epsilon)$ but $\det\nabla(\epsilon\pmb{\tilde{u}})=\mathcal{O}(\epsilon^{3})$ and $\psi(\epsilon\pmb{\tilde{u}})=\mathcal{O}(\epsilon^{3})$. By ignoring $\det\nabla(\epsilon\pmb{\tilde{u}})$ and $\psi(\epsilon\pmb{\tilde{u}})$, \cref{e:div} becomes $\nabla\cdot\pmb{\phi}\approx \det\nabla\pmb{\phi} + 2$. Consequently, we can control the determinant of the Jacobian by controlling the divergence of $\pmb{\phi}$, which explains the form of the constraint in \cref{e:ssdcont}. We refer to the study in~\cite{zhou2022:unique} for additional insights.

Next, we explore some of the properties of the described framework for grid generation.

\subsubsection{Inverse Consistency and Transitivity of VP}

\begin{figure}
\centering
\includegraphics[width=\textwidth]{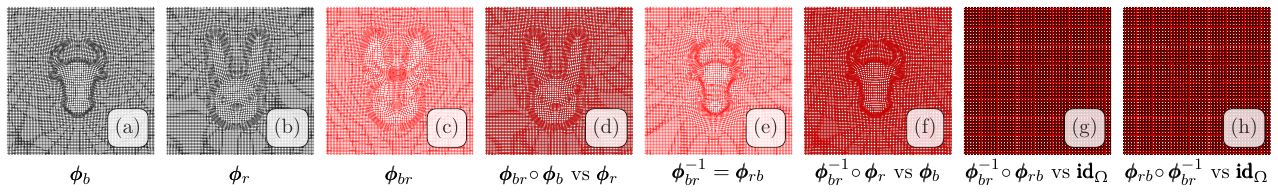}
\includegraphics[width=\textwidth]{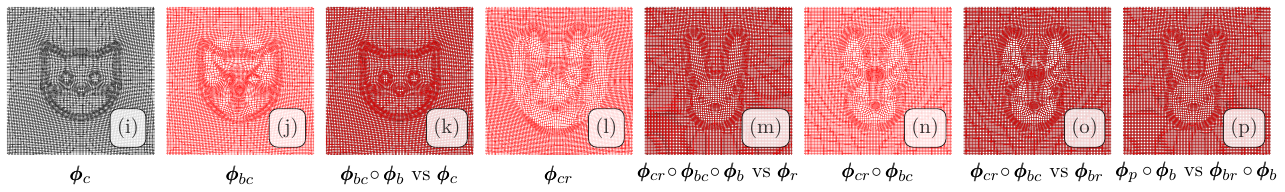}
\caption{We show three initial maps $\pmb{\phi}_b$ (a), $\pmb{\phi}_r$ (b), and $\pmb{\phi}_c$ (i) in black. The remainder of the figures illustrate maps generated by the VP. The figures labeled with ``vs'' (i.e., (d), (f)--(h), (k), (m)--(p)) compare maps overlaid in red to some reference mesh shown in black. The less the black grid is visible, the more accurate are our computations. The top row shows results for \emph{inverse consistency}. In (c) we show the grid associated with the mapping $\pmb{\phi}_{br}$ that maps $\pmb{\phi}_{b}$ to $\pmb{\phi}_r$. In (e) we show the inverse of $\pmb{\phi}_{br}$ generated via the VP. The remaining figures show compositions of transformations that either should yield grids that are similar to the initial maps given in (a) and (b) (see (d) and (f)) or the identity map $\pmb{\operatorname{id}}_{\Omega}$ (see (g) and (h)), respectively. The bottom row illustrates the \emph{transitivity}. We show the maps that take $\pmb{\phi}_{b}$ to $\pmb{\phi}_c$ and $\pmb{\phi}_{c}$ to $\pmb{\phi}_r$ in (j) and (l), respectively. The remainder of the figures ((k) and (m)--(n)) show various compositions and compare them to maps generated by the VP. The map $\pmb{\phi}_p$ in (p) is defined as $\pmb{\phi}_{cr}\circ\pmb{\phi}_{bc}$.}\label{f:invcons-trans}
\end{figure}

We empirically illustrate the inverse consistency and transitivity of diffeomorphic maps generated via the VP framework introduced above. We consider three initial grids: \begin{enumerate*}[label={\it(\alph*)}]
\item a bull grid $\pmb{\phi}_{b}$ (\Cref{f:invcons-trans}(a)),
\item a rabbit grid $\pmb{\phi}_{r}$ (\Cref{f:invcons-trans}(b)), and
\item a cat grid $\pmb{\phi}_{c}$ (\Cref{f:invcons-trans}(i))\end{enumerate*}.
These grids are elements of $\operatorname{Sol}(\Omega)$ and are generated by the VP.

First, we test \emph{inverse consistency} using $\pmb{\phi}_{b}$ and $\pmb{\phi}_{r}$. We construct a map $\pmb{\phi}_{br}$ that maps $\pmb{\phi}_b$ to $\pmb{\phi}_r$ and its inverse $\pmb{\phi}_{rb}$ using the VP. We verify whether the left and right composition of $\pmb{\phi}_{br}$ and $\pmb{\phi}_{rb}$ yield maps that are close to $\pmb{\operatorname{id}}_{\Omega}$. The results are shown in \Cref{f:invcons-trans}(g) and \Cref{f:invcons-trans}(h). These plots suggest that the computed diffeomorphisms are indeed close to being inverses of one another.

Second, assuming $\pmb{\phi}_{c}$ is a middle point between $\pmb{\phi}_{b}$ and $\pmb{\phi}_{r}$ we check the \emph{transitivity} of the VP. We generate a map $\pmb{\phi}_{bc}$ that maps $\pmb{\phi}_{b}$ to $\pmb{\phi}_{c}$ and a map $\pmb{\phi}_{cr}$ that maps $\pmb{\phi}_{c}$ to $\pmb{\phi}_{r}$ by the VP. Then, the composition $\pmb{\phi}_{bc} \circ \pmb{\phi}_{cr}$ should yield a grid that is close to $\pmb{\phi}_{br}$ \Cref{f:invcons-trans}(g). We show this in \Cref{f:invcons-trans}(n) and \Cref{f:invcons-trans}(o). Consequently, in \Cref{f:invcons-trans}(p), $\pmb{\phi}_{p}=\pmb{\phi}_{bc} \circ \pmb{\phi}_{cr}$ and $\pmb{\phi}_{br}$ should yield similar results as they are composed on $\pmb{\phi}_{b}$, which both the results will be close to $\pmb{\phi}_r$.

\subsection{Optimal Control Formulation for Diffeomorphic Image Registration}\label{s:vpcontrol}

In this section, we present an optimal control formulation that is founded on the principles we have outlined in~\Cref{s:vp}. This approach has been originally proposed in \cite{zhou2022:recent}. Overall, our work is founded on the fundamental idea that we would like to precisely control the properties of the diffeomorphic map that matches two images. In particular, we aim at modeling admissible transformations as elements of $\operatorname{Sol}(\Omega)$ (see \cref{e:sol}).

This approach has several advantages: \begin{enumerate*}[label={\it(\alph*)}]
\item As stated above, we can precisely control properties of the computed diffeomorphism (in particular, JD and curl).
\item Our formulation builds upon function compositions, which allows for a straightforward implementation of hierarchical multi-resolution and multi-scale schemes and/or re-gridding strategies.
\item We can initialize our formulation with crude (potentially, non-diffeomorphic) maps obtained by fast, inaccurate algorithms and generate nicely behaved smooth diffeomorphic maps\end{enumerate*}.

Our approach is formulated as follows. Let $M : \Omega \to \mathbb{R}$ be the moving image that is to be registered to the fixed image $F : \Omega \to \mathbb{R}$. We assume these images are compactly supported on the fixed domain $\Omega \subset \mathbb{R}^{d}$ of size $|\Omega|$. To measure the discrepancy between $M \circ \pmb{\phi}$ and $F$ we consider the Mean Squared Error ({\bf MSE})
\begin{equation}\label{e:mse}
\text{MSE}(\pmb{\phi}) \defeq \dfrac{1}{2|\Omega|}\int_{\Omega} \left(M(\pmb{\phi}(\pmb{\omega})) - F(\pmb{\omega})\right)^2 \text{d}\pmb{\omega}
\end{equation}

In our formulation, we seek a diffeomorphic map $\pmb{\phi}\in\operatorname{Sol}(\Omega)$ that minimizes the MSE measure in~\cref{e:mse} controlled by $f : \Omega \to \mathbb{R}_{++}$ and $\pmb{g} : \Omega \to \mathbb{R}^d$ as follows:
\begin{equation}\label{e:msecont}
\begin{aligned}
\minopt_{\pmb{\phi}, \,\,f,\,\, \pmb{g}}&\;\; \dfrac{1}{2|\Omega|}\int_{\Omega} (M(\pmb{\phi}(\pmb{\omega})) - F(\pmb{\omega}))^2\, \text{d}\pmb{\omega}\\
\begin{aligned}
\text{subject to}\\{}\\
\end{aligned} &\;\;
\begin{aligned}
\nabla \cdot \pmb{\phi} & = f+2 && \text{in}\;\;\Omega,\\
\nabla \times \pmb{\phi} & = \pmb{g} && \text{in}\;\;\Omega.
\end{aligned}
\end{aligned}
\end{equation}

\noindent Likewise to \Cref{s:vp}, we eliminate one constraint to obtain the PDE operator
\begin{equation}\label{e:mseC}
\Delta \pmb{\phi} =\nabla f - \nabla \times \pmb{g} \eqdef \pmb{C}(f,\pmb{g}) \quad \text{in}\;\; \Omega,
\end{equation}

\noindent where $\pmb{C}$ is the control function accountable for both JD and curl.

We derive the optimality conditions for this formulation in \Cref{s:appendvpcont}. We outline the associated algorithm in \Cref{alg:vpcont}. The algorithm was originally introduced in~\cite{zhou2022:recent}, but has never been described in detail. Notice that the iterative scheme consists of two stages. In the first stage, we compute updates associated with the auxiliary variable $\pmb{C}$. In the second stage, we compute updates associated with the controls $f$ and $\pmb{g}$. That is, we decompose the sought-after map $\pmb{\phi}$ into two maps---the map $\pmb{\phi}_{\text{global}}$ and $\pmb{\phi}_{\text{local}}$. The map $\pmb{\phi}_{\text{global}}$ is computed based on an iterative procedure associated with $\pmb{C}$. The map $\pmb{\phi}_{\text{local}}$ is computed via an iterative procedure for updating $f$ and $\pmb{g}$. The final map $\pmb{\phi}$ is given by the composition of $\pmb{\phi}_{\text{global}}$ and $\pmb{\phi}_{\text{local}}$, i.e., $\pmb{\phi} = \pmb{\phi}_{\text{global}} \circ \pmb{\phi}_{\text{local}}$. There are two main reasons for this two-stage design:
\begin{enumerate*}[label={\it(\alph*)}]
\item Motivated by the empirically observed performance of VP (see example 4.1 in \cite{zhou2023:vpgrid}) in the context of grid generation, we observed that the iterative updates based on the controls $f$ and $\pmb{g}$ tend to converge with a larger gradient-step size compared to the updates associated with the auxiliary variable $\pmb{C}$.
\item Conversely, we observed in the context of the image registration problem, that iterative updates of the auxiliary variable $\pmb{C}$ in \cref{e:msedelc2} will stagnate once a shapes are ``globally'' aligned. Local deformation patterns associated with fine structures in the images are mostly driven by updates based on the controls $f$ and $\pmb{g}$ in \cref{e:msedelfg}; however, these updates are not as effective as $\pmb{C}$ in \cref{e:msedelc2} in ``globally'' aligning the shapes inside the images.
\end{enumerate*}
We attribute these observations to the fact that the gradient with respect to the auxiliary variable $\pmb{C}$ in \cref{e:msedelc2} involve a second-order derivative operator, the Laplacian, which results in a smooth  (i.e., ``global'') gradient step; but the gradients with respect to $f$ and $\pmb{g}$ in \cref{e:msedelfg} only involve first-order derivatives; they can capture ``high-frequency updates.'' Overall, this observation lead to a two-step algorithm; first, we compute a global, ``smooth'' alignment by iterating on the auxiliary variable $\pmb{C}$; subsequently, we refine the computed map by iterating on $f$ and $\pmb{g}$, respectively.

\begin{figure}
\centering
\includegraphics[width=\textwidth]{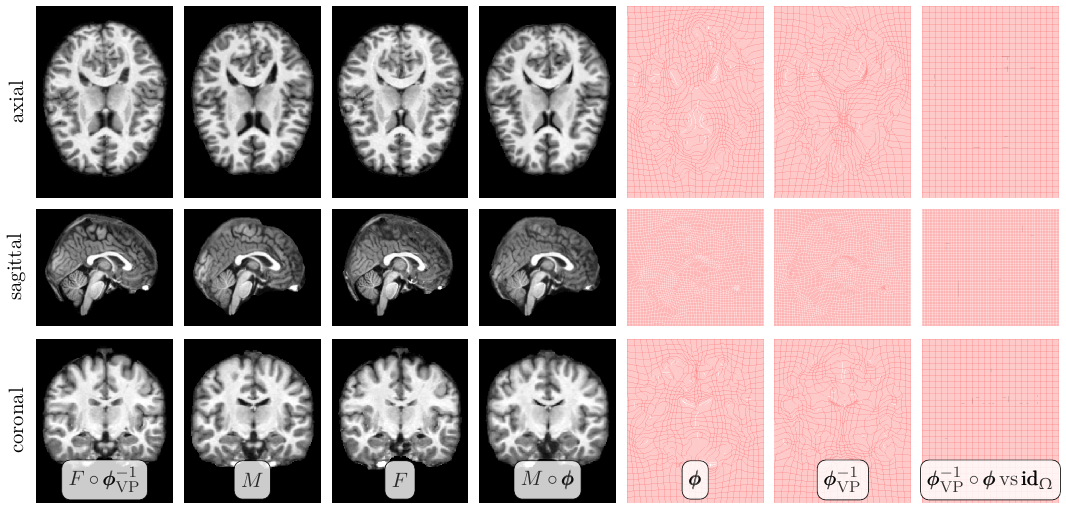}
\caption{Representative registration results. Each row (from top to bottom) shows a coronal, axial, and sagittal view of the computed results. The map $\pmb{\phi}$ is the diffeomorphic solution found by our optimal control approaches; the map $\pmb{\phi}_{\text{VP}}^{-1}$ is the inverse of $\pmb{\phi}$ that is constructed by the VP. We show (from left to right) $F$ composed with $\pmb{\phi}^{-1}_{\text{VP}}$, the moving image $M$, the fixed image $F$, $M$ composed with $\pmb{\phi}$, the diffeomorphic map $\pmb{\phi}$, the inverse $\pmb{\phi}_{\text{VP}}$, and the composition of $\pmb{\phi}$ and $\pmb{\phi}_{\text{VP}}$. For the latter, we overlay the resulting map (in red) to the identity transformation $\pmb{\operatorname{id}}_{\Omega}$ (in black).}\label{f:mri3d}
\end{figure}

We show representative results for the approach outlined above in \Cref{f:mri3d}. Here, $\pmb{\phi}$ is the diffeomorphic map found by solving \cref{e:msecont}; $M \circ \pmb{\phi}$ is the deformed moving image. We also compute the inverse of $\pmb{\phi}$ based on the VP described in \Cref{s:vp}. To illustrate the accuracy of our method, we compose $\pmb{\phi}$ by $\pmb{\phi}^{-1}_{\text{VP}}$. We show the resulting grid in red superimposed on a black grid for $\pmb{\operatorname{id}}_{\Omega}$. The most important observations are that
\begin{enumerate*}[label={\it(\alph*)}]
\item the computed maps are diffeomorphic;
\item the image $M \circ \pmb{\phi}$ is in excellent agreement with $F$.
\item the image $F \circ \pmb{\phi}^{-1}_{\text{VP}}$ is in excellent agreement with $M$; and
\item the composition of $\pmb{\phi}^{-1}_{\text{VP}}$ and $\pmb{\phi}$ is close to $\pmb{\operatorname{id}}_{\Omega}$
\end{enumerate*}.

In the next section, we are going to address some of the drawbacks associated with the numerical approach and problem formulation outlined in this section.

\subsection{A Penalty Approach for Diffeomorphic Image Registration}\label{s:vpregel}

As we have mentioned above, the iterative scheme outlined in \Cref{alg:vpcont} consists of two stages---one to compute $\pmb{\phi}_{\text{global}}$ and one to compute $\pmb{\phi}_{\text{local}}$. In each stage, we iterate until convergence. In each iteration, we have to solve two Poisson equations. While the implementation of these Poisson solves is done in an effective way via spectral (diagonalization) methods or multigrid techniques (we use pseudo-spectral method with a Fourier basis to invert the Laplacian operator), these two Poisson solves constitute the main computational bottleneck of our numerical scheme. In this section, we develop a numerical scheme that is more effective.

In the formulation in~\cref{e:ssd} the target transformation $\pmb{\phi}_{t}$ is not known; but with prescribed JD and curl, namely, ${f}_t$ and $\pmb{g}_t$, $\pmb{\phi}=\pmb{\phi}_{t}$ can be constructed. In the control formulation in~\cref{e:mse} and~\cref{e:msecont}, the spatial transformation $\pmb{\phi}$ is also not known; but by controlling $f$ and $\pmb{g}$ mimicking JD and curl, $\pmb{\phi}=\pmb{\phi}_{t}$ can be found. In this section, we want to establish a mechanism that drives $\pmb{\phi}$ to $\pmb{\phi}_{t}$ while $f$ and $\pmb{g}$ are driven to the unknown ${f}_t$ and $\pmb{g}_t$, where $f_t$ and $\pmb{g}_t$ are explicitly represented by other terms.

In our reformulation of the problem, we treat the hard constraints \cref{e:msecont} as soft penalties. That is, we introduce quadratic penalties that penalize deviations from the constraints in \cref{e:msecont}. In addition, we introduce soft constraints that penalize the deviation for $f_t$ from $f$ and $\pmb{g}$ from $\pmb{g}_t$. Likewise to \cref{e:msecont}, we consider \cref{e:mse} to measure the discrepancy between $M \circ \pmb{\phi}$ and $F$. Overall, we arrive at the following objective function. We seek the map $\pmb{\phi}\in\operatorname{Sol}(\Omega)$ that minimizes
\begin{equation}\label{e:vpregeltarget}
\begin{aligned}
\mathcal{V}(\pmb{\phi},f,\pmb{g},f_t,\pmb{g}_t)
&= \text{MSE}(\pmb{\phi})
+ \int_{\Omega}(\nabla\cdot\pmb{\phi} - f_t-2)^{2}d\pmb{\omega}
+ \int_{\Omega}(f_t-f)^{2}d\pmb{\omega}\\
& \quad + \int_{\Omega}\|\nabla \times\pmb{\phi}- \pmb{g}_t\|^{2}d\pmb{\omega}
+\int_{\Omega}\|\pmb{g}_t-\pmb{g}\|^{2}d\pmb{\omega}
\end{aligned}
\end{equation}

\noindent with $\pmb{\phi}=\pmb{\operatorname{id}}_{\Omega}$ on $\partial \Omega$ as the unknown control functions ${f}$ and $\pmb{g}$ approach the unknown target control functions ${f}_t$ and $\pmb{g}_t$.

We derive the associated optimality conditions for \cref{e:vpregeltarget} in \cref{s:appendvpregel}. The associated algorithm is summarized in \Cref{alg:vpcontelalg}. This algorithm still has two stages. However, at each outer iteration of each stage we only have to solve one Poisson equation as opposed to two Poisson equations in \Cref{alg:vpcont} (we have removed one elliptic solve from our problem per outer iteration).

\subsection{Computing the Inverse Map}\label{s:lmvp}

In this section, we introduce a strategy to compute the inverse of a given map $\pmb{\phi} \in \operatorname{Sol}(\Omega)$. This approach builds upon the formulation in \Cref{s:vp}. Being able to compute the inverse map is a critical requirement in many applications of diffeomorphic image registration. Our approach will guarantee that $\pmb{\phi}^{-1}$ is also an element of $\operatorname{Sol}(\Omega)$. Recall that in the formulation in \Cref{s:vp} properly prescribing the target JD and curl through $f_t$ and $\pmb{g}_t$ is key to construct the target grid $\pmb{\phi}_t$. For constructing the inverse of a given grid $\pmb{g}_{o}$ this is not the case; we know precisely what $f_t$ and $\pmb{g}_t$ should be, namely, $f_t = \det\nabla\pmb{\operatorname{id}}_{\Omega} = 1$ and $\pmb{g}_t = \nabla\times\pmb{\operatorname{id}}_{\Omega}=\pmb{0}$. We also know that the target grid is given by $\pmb{\phi}_t = \pmb{\operatorname{id}}_{\Omega}$. We can use this strategy within our diffeomorphic image registration framework to find the inverse of the computed map. In particular, given the original transformation $\pmb{\phi}_{o}$ and the target transformation $\pmb{\phi}_{t} (= \pmb{\operatorname{id}}_{\Omega}) \in \operatorname{Sol}(\Omega)$, we seek a diffeomorphic transformation $\pmb{\phi} = \pmb{\phi}_{m} \circ \pmb{\phi}_{o} = \pmb{\phi}_{m}(\pmb{\phi}_{o}) \in \operatorname{Sol}(\Omega)$ that is close to $\pmb{\phi}_{t}$. Here, $\pmb{\phi}_{m}$ is an intermediate transformation that left-translates $\pmb{\phi}_{o}$ to $\pmb{\phi}$.

We introduce the unknown Lagrange multipliers $\lambda_{f}$ and $\pmb{\lambda}_{\pmb{g}}=(\lambda_{\pmb{g}1},\lambda_{\pmb{g}2},\lambda_{\pmb{g}3})$ on $\Omega$ to handle the constraints associated with the curl and JD of $\pmb{\phi}$. We measure the discrepancy between $\pmb{\phi}$ and $\pmb{\phi}_t$ using a squared $L^2$ penalty. We also introduce regularizers to control $f$ and $\pmb{g}$. The resulting Lagrangian functional is given by
\begin{equation}\label{e:lmvp}
\begin{aligned}
\mathcal{L}(\pmb{\phi},\lambda_{f},\pmb{\lambda}_{\pmb{g}},f,\pmb{g})
&= \frac{1}{2}\int_{\Omega} \|\pmb{\phi}-\pmb{\phi}_{t}||^{2}\,\text{d}\pmb{\omega}
+ \int_{\Omega}[\lambda_{f}(\det\nabla\pmb{\phi} - f)]\,\text{d}\pmb{\omega}
+ \frac{1}{2}\int_{\Omega}f^{2}\,\text{d}\pmb{\omega}\\
& \quad + \int_{\Omega}[\pmb{\lambda}_{\pmb{g}}\cdot(\nabla \times\pmb{\phi}- \pmb{g})]\,\text{d}\pmb{\omega}
+ \frac{1}{2}\int_{\Omega}\|\pmb{g}\|^{2}\,\text{d}\pmb{\omega},
\end{aligned}
\end{equation}

\noindent where $f(\pmb{\omega})>0$, $\pmb{g}(\pmb{\omega})$ are control functions on $\Omega$ that satisfy~\cref{e:fg}. We derive the optimality conditions for \cref{e:lmvp} in \Cref{s:appendlmvp}. The algorithm is summarized in \Cref{alg:lmvpalg}.

Similarly to the use of VP in the context of image registration, this approach is used to find the inverse of the image registration map. That is, let $\pmb{\phi}_{o}$ denote the original map. Then, we set $\pmb{\phi}_{t}=\pmb{\operatorname{id}}_{\Omega}$ and construct the intermediate transformation $\pmb{\phi}_{m}$ such that $\pmb{\phi}_{m} \circ \pmb{\phi}_{o}$ is close to $\pmb{\operatorname{id}}_{\Omega}$. Consequently, $\pmb{\phi}_{m}$ approximates $\pmb{\phi}_{o}^{-1}$.

The optimality conditions associated with \cref{e:lmvp} allow us to significantly simplify the computational steps required to construct the sought after grids. In the original VP formulation presented in \Cref{s:vp} we have to solve two Poisson equations at each iteration when we attempt to find the inverse of a given map $\pmb{\phi}_{o}$. Our reformulation of the problem allows us to significantly reduce the time-to-solution; we can express all key unknowns explicitly.

Next, we will expose how we integrated the proposed framework into VPreg; we also demonstrate that it yields an effective method for recovering the inverse transformations.

\subsection{Diffeomorphic Image Registration Framework}\label{s:vpreg}

The proposed approach for diffeomorphic image registration is composed of two main components. We use the problem formulation in \Cref{s:vpregel} to find the sought after diffeomorphic map $\pmb{\phi}$ that transforms the moving image $M$ to the target image $F$. In addition, we consider the formulation introduced in \Cref{s:lmvp} to construct $\pmb{\phi}^{-1}$. We term the resulting method ``VPReg.'' Integrating these two reformulations of the original approaches to solve the diffeomorphic registration problem (see \Cref{s:vpcontrol}) and the original grid generation method (see \Cref{s:vp}) yields a more effective method for diffeomorphic image registration while still precisely controlling the properties of the computed diffeomorphism.

One key challenge in dealing with distance measures is that many imaging modalities (in particular magnetic resonance imaging) do not provide consistent data; the data exhibit intensity shifts and distortions, noise perturbations, and other inconsistencies across different imaging sessions, imaging studies, or imaging sites. Many of the available distance measures are sensitive to these intensity perturbations. This makes their use in imaging studies delicate. Consequently, many image processing pipelines utilize a range of image intensity normalization techniques such as shifting intensities to $[0,1]$, max-min-matching, histogram-matching, etc. In the present work, we follow the idea of conversion to a z-score: Given an image $M$ defined on $\Omega$ with voxel coordinates $(i,j,k) \in \Omega$, we denote by $\mu=\sum_{i,j,k}M(i,j,k) / |\Omega|$ the mean value and by $\sigma = (\sum_{i,j,k}(\mu-M(i,j,k))^{2})^{1/2} / (|\Omega|-1)$ the standard deviation. The z-score of $M$ is computed as
\begin{equation}\label{e:zs}
M_{z}(i,j,k)=\frac{M(i,j,k)-\mu}{\sigma}.
\end{equation}

This rescaling and shift operation is a widely applied statistical tool that converts different measurements with distinct units to a reference distribution with z-score mean $0$. We observed empirically that applying this transformation improved the performance of our method (made it less sensitive to intensity drifts).

We summarize the proposed algorithm in \Cref{alg:VPregalg}.

\begin{algorithm}
\caption{Proposed diffeomorphic image registration algorithm (VPreg).\label{alg:VPregalg}}
\begin{algorithmic}[1]
\STATE \textbf{input:} images $M$, $F$
\STATE initialize $\pmb{\phi} \gets \pmb{\operatorname{id}}_{\Omega}$
\STATE $M_{z}, F_{z} \gets$ convert $M$, $F$ to their z-score representations
\STATE $\pmb{\phi} \gets$ solve registration problem with inputs $M_{z}$, $F_{z}$ (see \Cref{alg:vpcontelalg})
\STATE $M_{\pmb{\phi}} \gets$ interpolate $M \circ \pmb{\phi}$
\STATE $\pmb{\phi}^{-1} \gets$ execute grid generation method (see \cref{alg:lmvpalg}) with inputs $\pmb{\phi}_{o} \gets \pmb{\phi}$, $\pmb{\phi}_{t} \gets \pmb{\operatorname{id}}_{\Omega}$
\STATE $F_{\pmb{\phi}^{-1}} \gets $ interpolate $F \circ \pmb{\phi}^{-1}$
\STATE \textbf{output:} $M_{\pmb{\phi}}$, $F_{\pmb{\phi}^{-1}}$, $\pmb{\phi}$, $\pmb{\phi}^{-1}$
\end{algorithmic}
\end{algorithm}

\section{Results}\label{s:results}

We explore the performance of the proposed method on neuroimaging data. We first discuss the dataset. Subsequently, we present some measures we consider to assess registration performance. Then we report detailed results for a representative pair of images. We conclude with a study for a cohort of patients.

We compare the performance of the proposed method to ANTs (in particular, antsRegistrationSyN)~\cite{avants2011:reproducible, avants2008:symmetric}, Easyreg (Freesurfer)~\cite{iglesias2023:ready,hoffmann2021:synthmorph}, and Fnirt (FSL)~\cite{jenkinson2012:fsl,andersson2007:non}. ANTs and Easyreg provide not only the forward map $\pmb{\phi}$ but also the inverse map $\pmb{\phi}^{-1}$ (or an approximation thereof) as an output. Fnirt requires a post processing step to generate the inverse (by calling \texttt{invwarp}). Since one of the primary concerns of this work is to control the properties of the computed transformation and by that generate well-behaved transformation maps, we will not only report metrics for registration accuracy but also metrics to assess the quality of the computed registration map and its inverse. We report the parameters and settings for the baseline methods in \Cref{a:baseline}.

\subsection{Data}

We report registration results for the public dataset OASIS-1~\cite{marcus2007:oasis, oasis-web}; this dataset consists of over 600 T1 MRI 3D volumes of different individuals. The age of the healthy subjects ranges from 18 to 60. The age of those diagnosed with dementia is 61 and above. We expect the registration methods to be more sensitive to mild, subtle, or early neural pathological changes. For our experiments, we purposely selected 35 subjects between the ages of 18 to 34 so that their variabilities caused by neurodegenerative disease are expected to be relatively small. For each subject in the dataset, there are two versions of raw data, one in the scanner image space and another is already linearly registered onto some template image space. In order to study the differences in nonlinear registration methods, we consider the second version of the raw data that are linearly pre-registered to a reference dataset (i.e., they are spatially normalized). Each dataset provides two levels of ROI annotations, which we can use for evaluating the alignment of corresponding anatomical structures. The first level includes four ROIs: cortex, subcortial gray matter ({\bf GM}), white matter ({\bf WM}), and cerebrospinal fluid ({\bf CSF}). The second level includes 35 ROIs.

\subsection{Measures of Registration Performance}

Here, we present some measures that we will use as criteria to assess the performance of the considered methods. Given two images $M$, $F$, and numbers of non-zero voxels in $M$, $F$ and $M \cap F$, denoted by $|M|$, $|F|$ and $|M\cap F|$, we consider the DICE score
\begin{equation}\label{e:dice}
\text{DICE}(M,F)= \frac{2|M\cap F|}{|M|+|F|}.
\end{equation}

\noindent This measure takes on values in $[0,1]$.

In addition, we consider the relative MSE based on \cref{e:mse} given by
\begin{equation}\label{e:mserate}
\text{MSE-ratio}
= \frac{\text{MSE}(M(\pmb{\phi}),F)}{\text{MSE}(M,F)}.
\end{equation}

Moreover, we consider a probabilistic similarity measure. Let $P_{M}(i)$, $P_{F}(j)$ be marginal probabilities (intensity histogram distributions) of $M$, $F$, and $P_{M,F}(i,j)$ denote the conditional probability. Then, the mutual information between two images $M$ and $F$ is given by
\begin{equation}\label{e:mi}
\text{MI}(M,F)
= \int_{m(i)\in M} \int_{f(j)\in F} P_{M,F}(i,j) \log\left(\frac{P_{M,F}(i,j)}{P_{M}(i)P_{F}(j)}\right) \,\text{d}i \text{d}j.
\end{equation}

This measure takes on values in $(0,+\infty)$. The relative MI increment based on \cref{e:mi} is given by
\begin{equation}\label{e:mirate}
\text{MI-Incr}
= \frac{\text{MI}(M(\pmb{\phi}),F)-\text{MI}(M,F)}{\text{MI}(M,F)}.
\end{equation}

We also report measures for the regularity of the computed transformations based on JD. We additionally assess how accurate the inverse transformations are by comparing the composition of the computed map and its inverse to the identity transformation.

Aside from reporting averages, we also include box-whisker plots to provide summary statistics. These box-whisker plots show the minimum (Oth percentile; whisker at bottom), the maximum (100th percentile; whisker at top), the medial (50th percentile; line in middle), and the first and third quartile (25th and 75th percentile; box), respectively.

\subsection{Representative Registration Results}

\begin{figure}
\centering
\subfigure[$M4$]{\includegraphics[width=3.0cm,height=4cm]{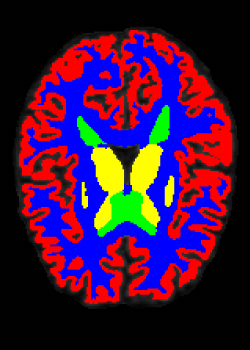}}
\subfigure[ANTs $F4(\pmb{\phi}^{-1})$]{\includegraphics[width=3.0cm,height=4cm]{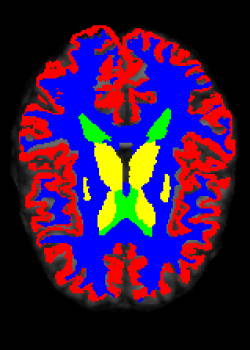}}
\subfigure[Easyreg $F4(\pmb{\phi}^{-1})$]{\includegraphics[width=3.0cm,height=4cm]{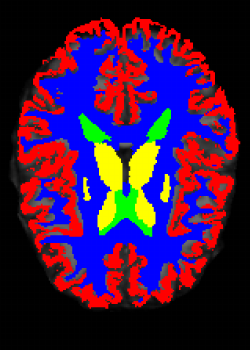}}
\subfigure[Fnirt $F4(\pmb{\phi}^{-1})$]{\includegraphics[width=3.0cm,height=4cm]{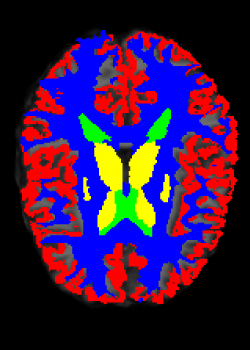}}
\subfigure[VPreg $F4(\pmb{\phi}^{-1})$]{\includegraphics[width=3.0cm,height=4cm]{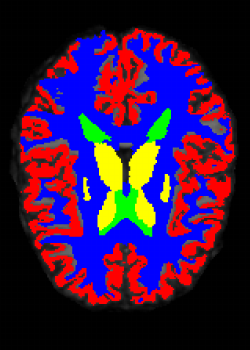}}

\subfigure[$M$]{\includegraphics[width=3.0cm,height=4cm]{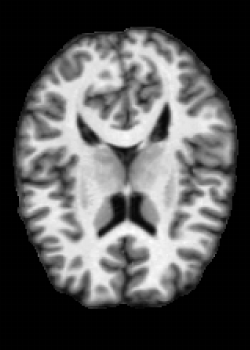}}
\subfigure[ANTs $F(\pmb{\phi}^{-1})$]{\includegraphics[width=3.0cm,height=4cm]{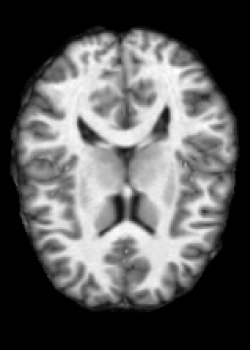}}
\subfigure[Easyreg $F(\pmb{\phi}^{-1})$]{\includegraphics[width=3.0cm,height=4cm]{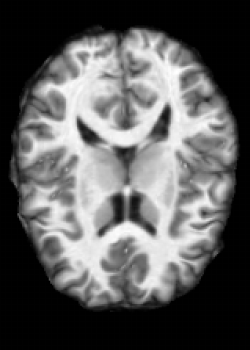}}
\subfigure[Fnirt $F(\pmb{\phi}^{-1})$]{\includegraphics[width=3.0cm,height=4cm]{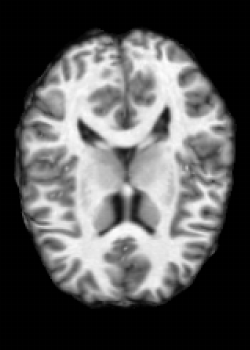}}
\subfigure[VPreg $F(\pmb{\phi}^{-1})$]{\includegraphics[width=3.0cm,height=4cm]{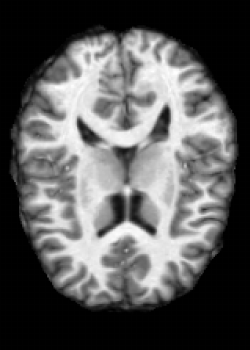}}

\subfigure[$F$]{\includegraphics[width=3.0cm,height=4cm]{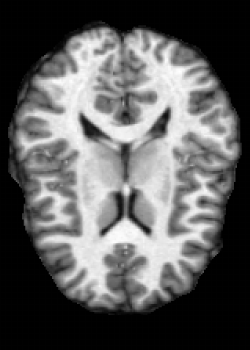}}
\subfigure[ANTs $M(\pmb{\phi})$]{\includegraphics[width=3.0cm,height=4cm]{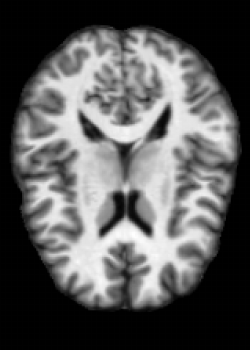}}
\subfigure[Easyreg $M(\pmb{\phi})$]{\includegraphics[width=3.0cm,height=4cm]{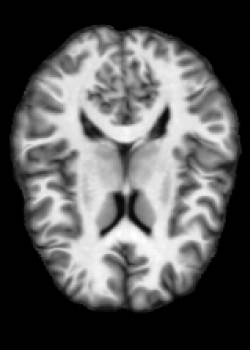}}
\subfigure[Fnirt $M(\pmb{\phi})$]{\includegraphics[width=3.0cm,height=4cm]{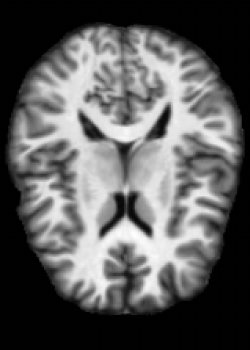}}
\subfigure[VPreg $M(\pmb{\phi})$]{\includegraphics[width=3.0cm,height=4cm]{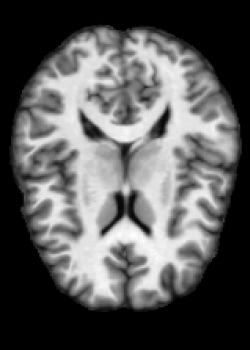}}

\subfigure[$F4$]{\includegraphics[width=3.0cm,height=4cm]{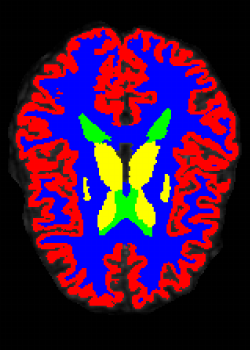}}
\subfigure[ANTs $M4(\pmb{\phi})$]{\includegraphics[width=3.0cm,height=4cm]{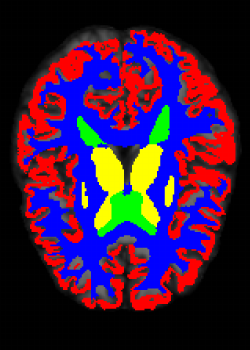}}
\subfigure[Easyreg $M4(\pmb{\phi})$]{\includegraphics[width=3.0cm,height=4cm]{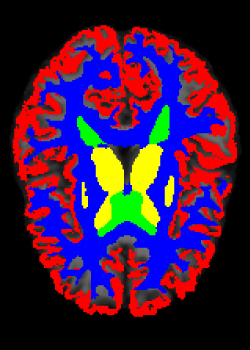}}
\subfigure[Fnirt $M4(\pmb{\phi})$]{\includegraphics[width=3.0cm,height=4cm]{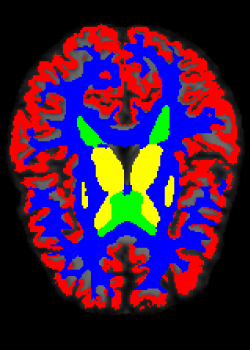}}
\subfigure[VPreg $M4(\pmb{\phi})$]{\includegraphics[width=3.0cm,height=4cm]{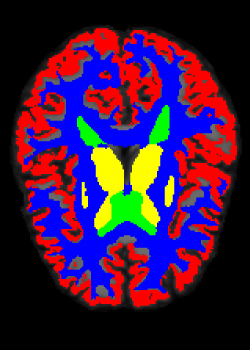}}
\caption{Axial slice of two representative images from the OASIS-1 dataset (registration from subject-0012 to subject-0140). Columns (left to right): Original data, ANTs, Easyreg, Fnirt, and VPreg. First row: four ROIs on $M$. Second row: $M$ compared to $F \circ \pmb{\phi}^{-1}$. Third row: $F$ compared to $M \circ \pmb{\phi}$. Bottom row: four ROIs on $F$.}\label{f:mri3dvs}
\end{figure}

\begin{figure}
\centering
\includegraphics[width=3.20cm,height=4.1cm]{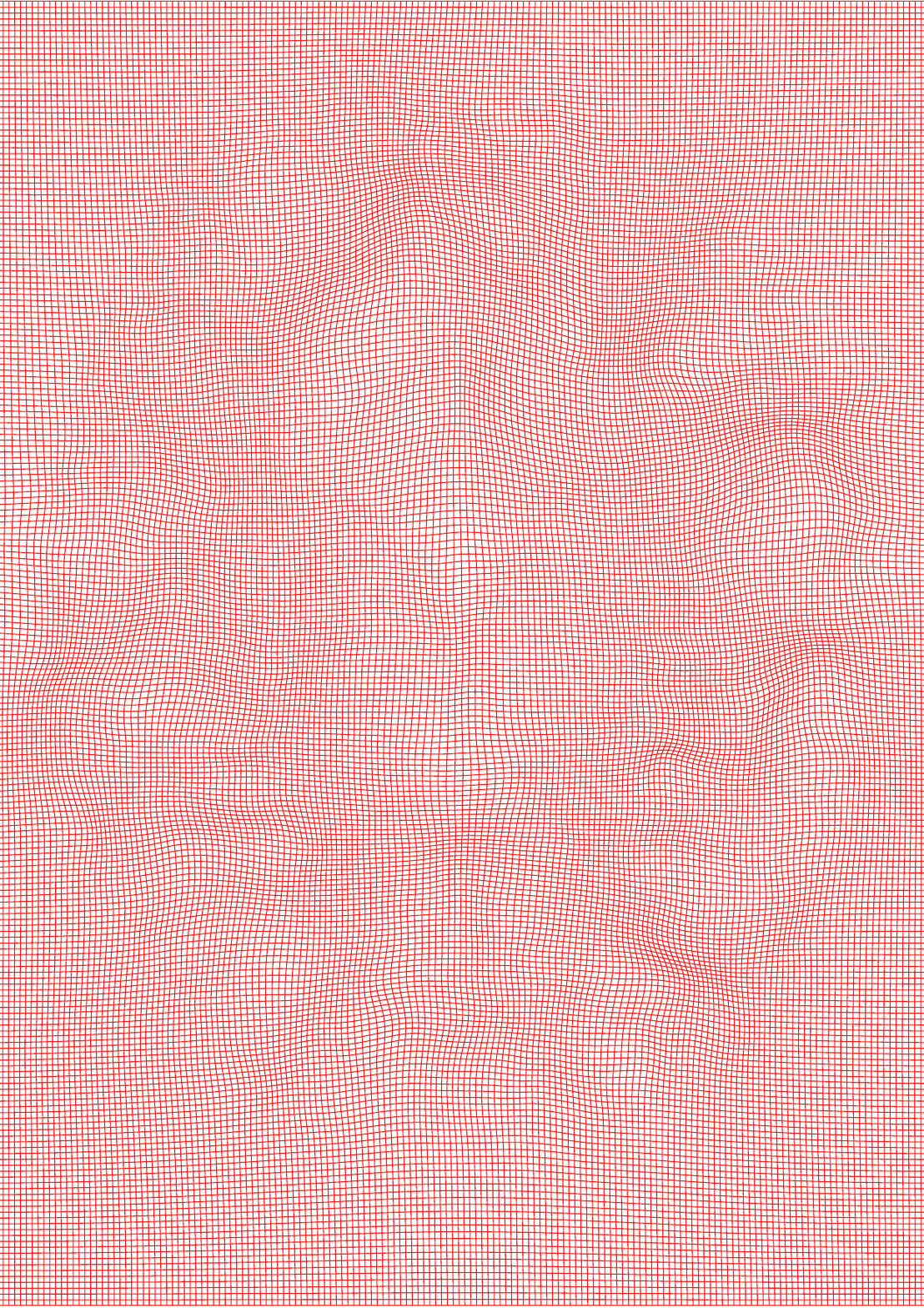}
\includegraphics[width=3.20cm,height=4.1cm]{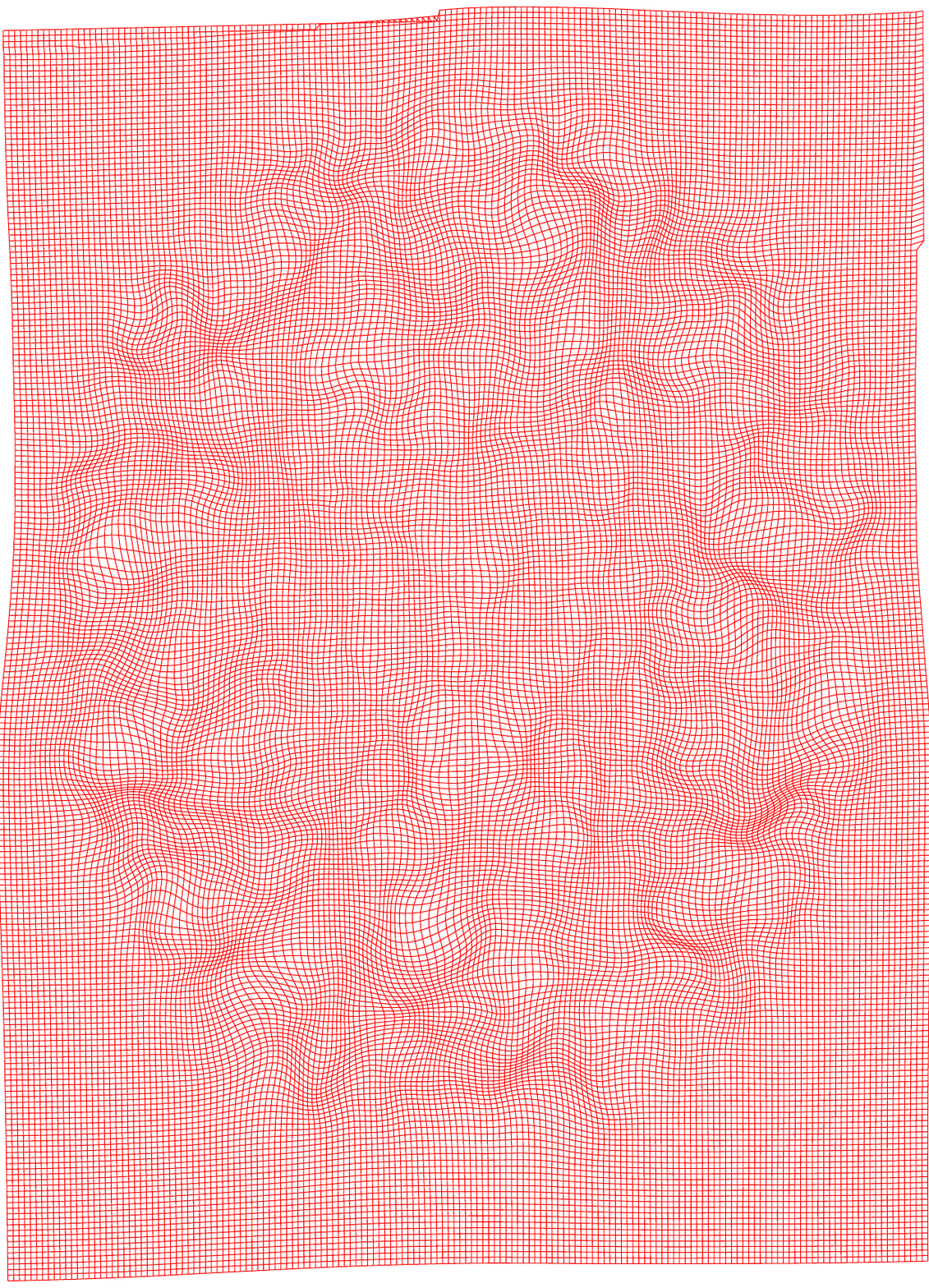}
\includegraphics[width=3.20cm,height=4.1cm]{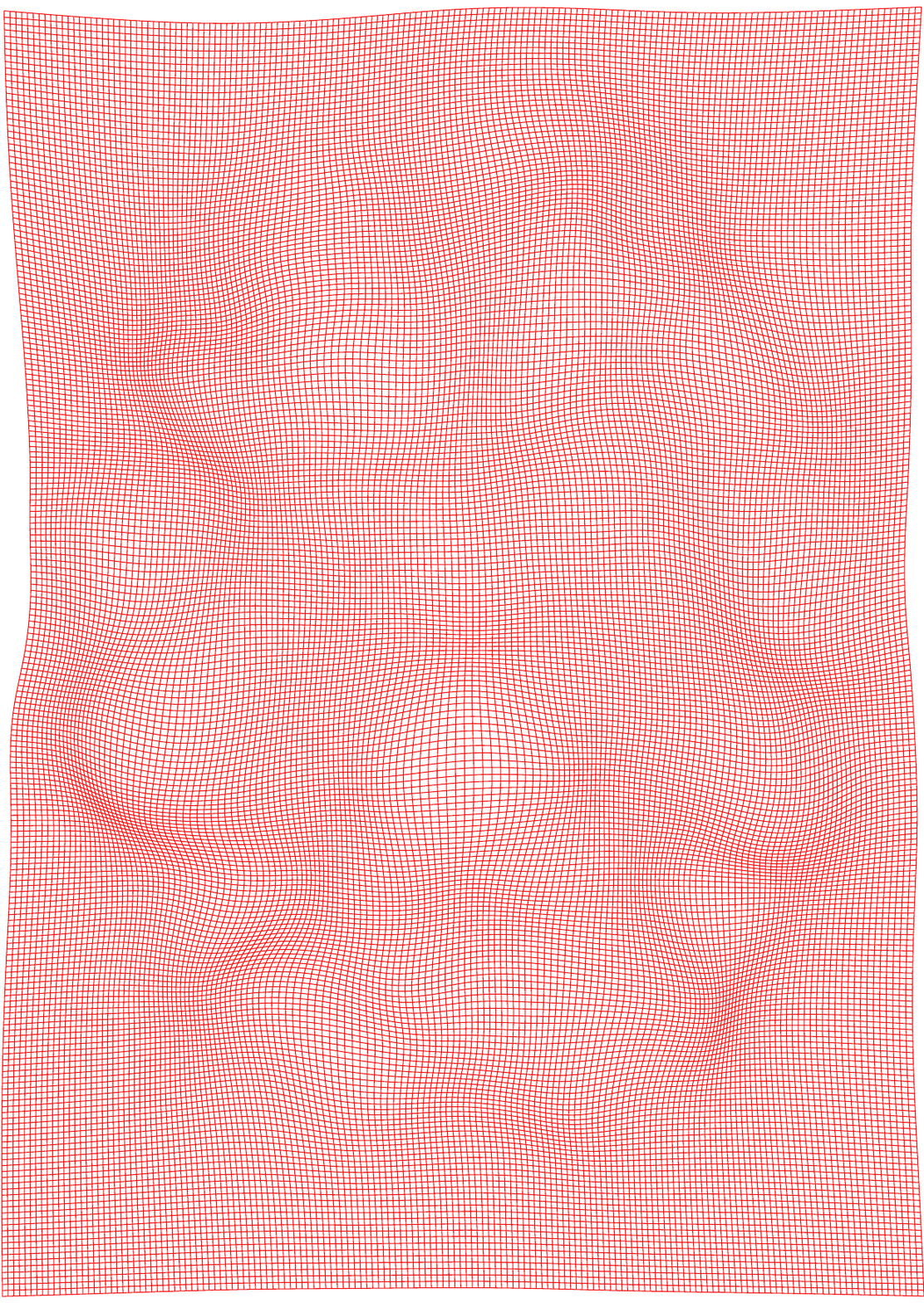}
\includegraphics[width=3.20cm,height=4.1cm]{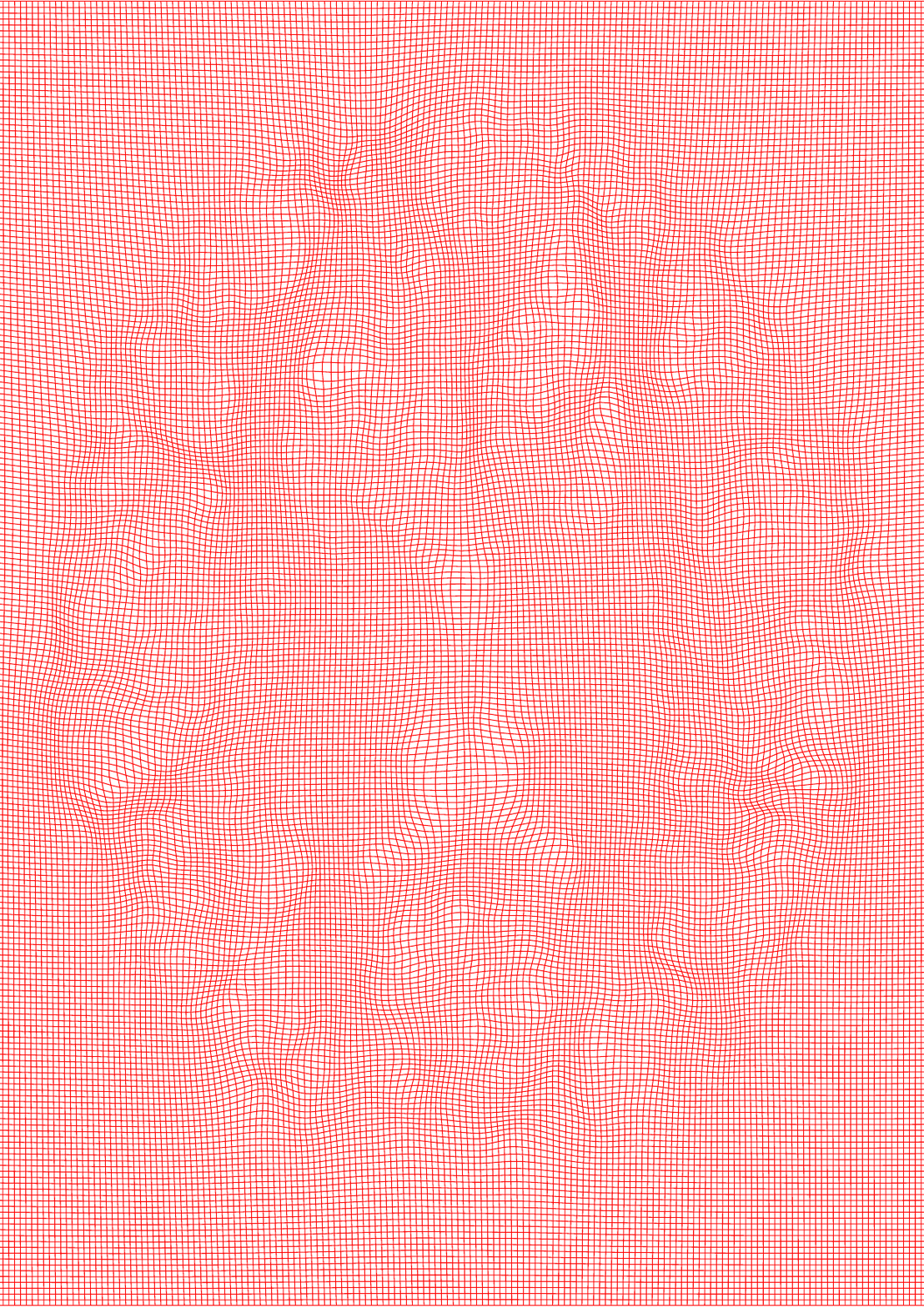}

\vspace{0.1cm}
\includegraphics[width=3.20cm,height=4.1cm]{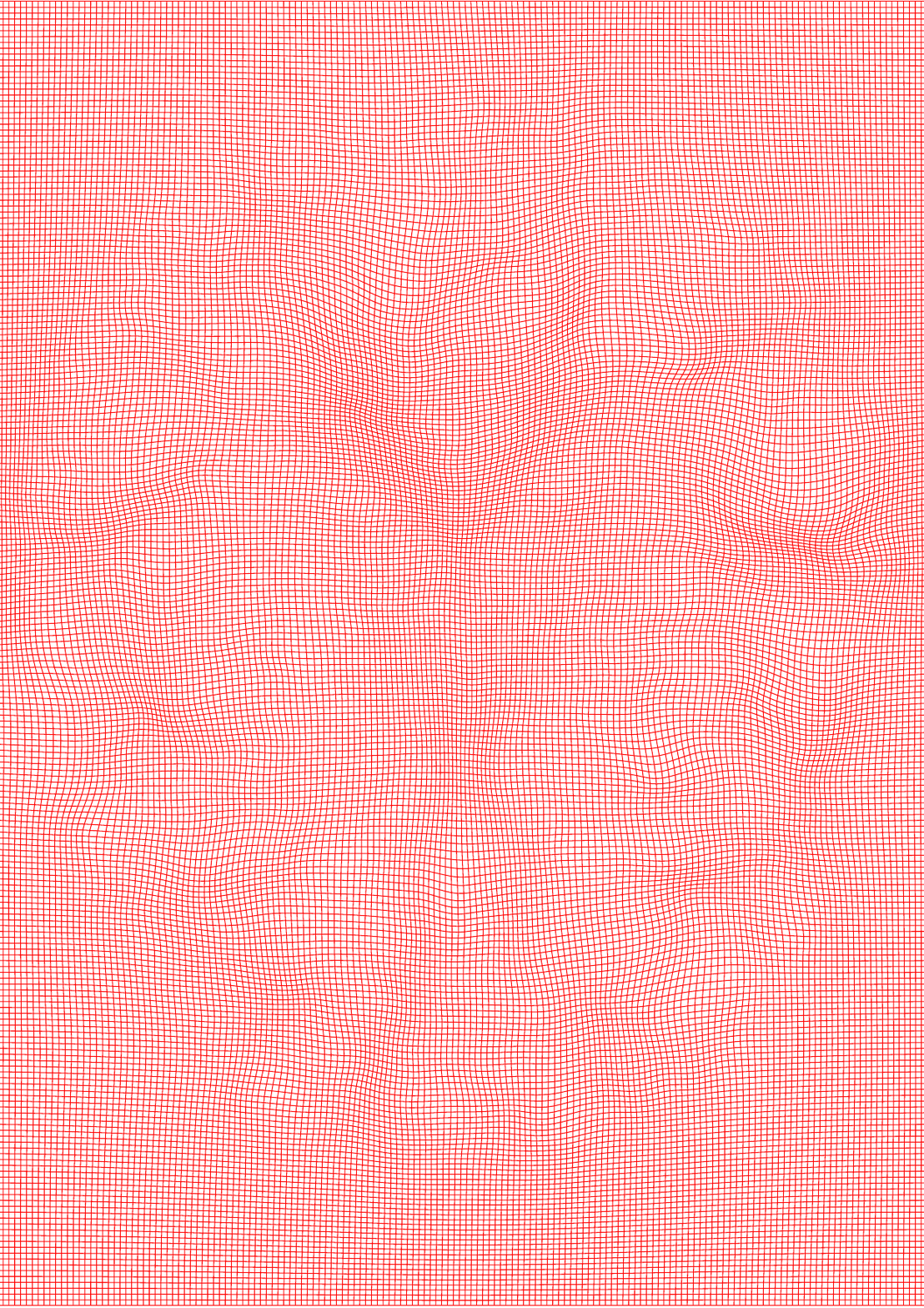}
\includegraphics[width=3.20cm,height=4.1cm]{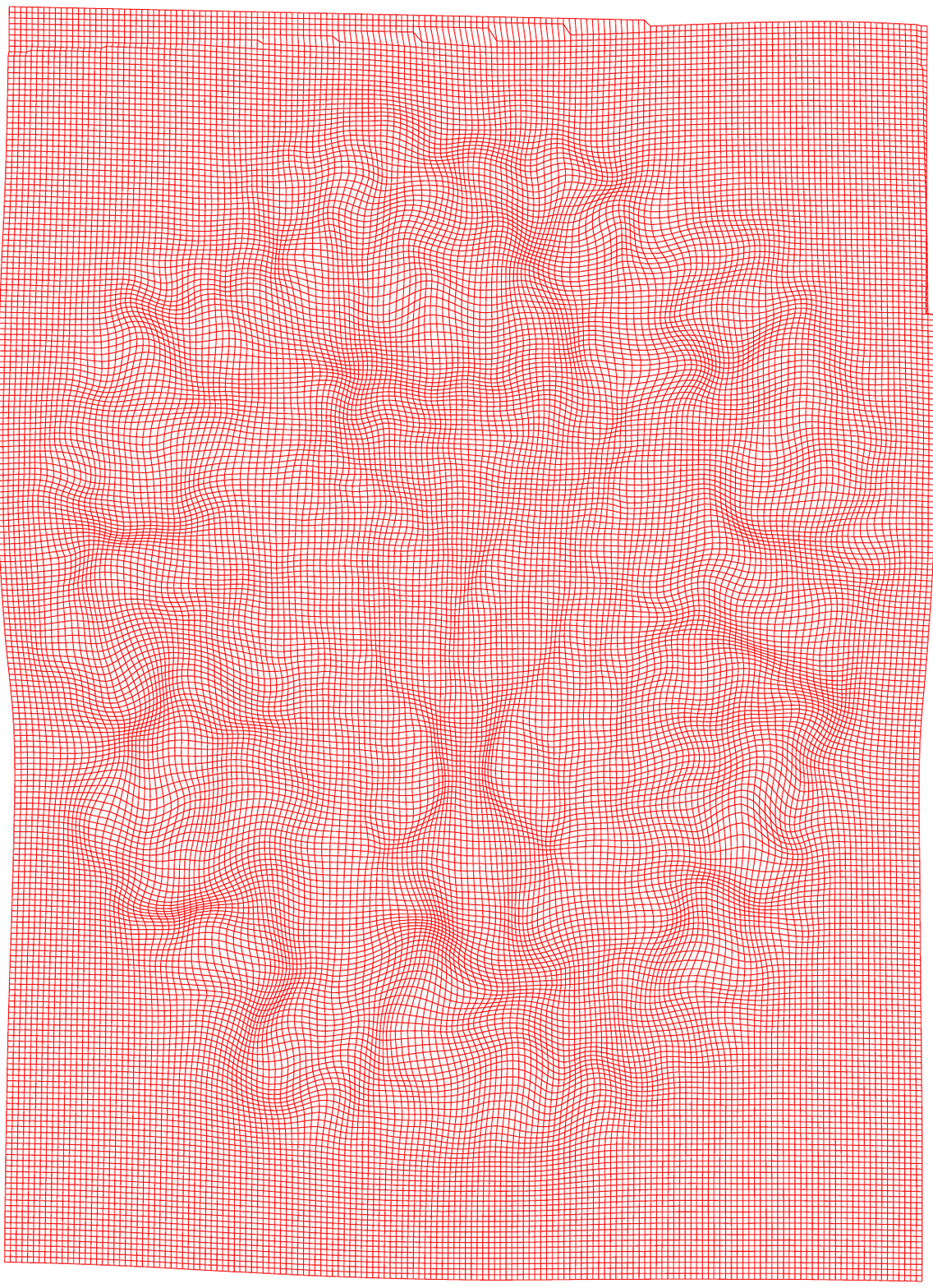}
\includegraphics[width=3.20cm,height=4.1cm]{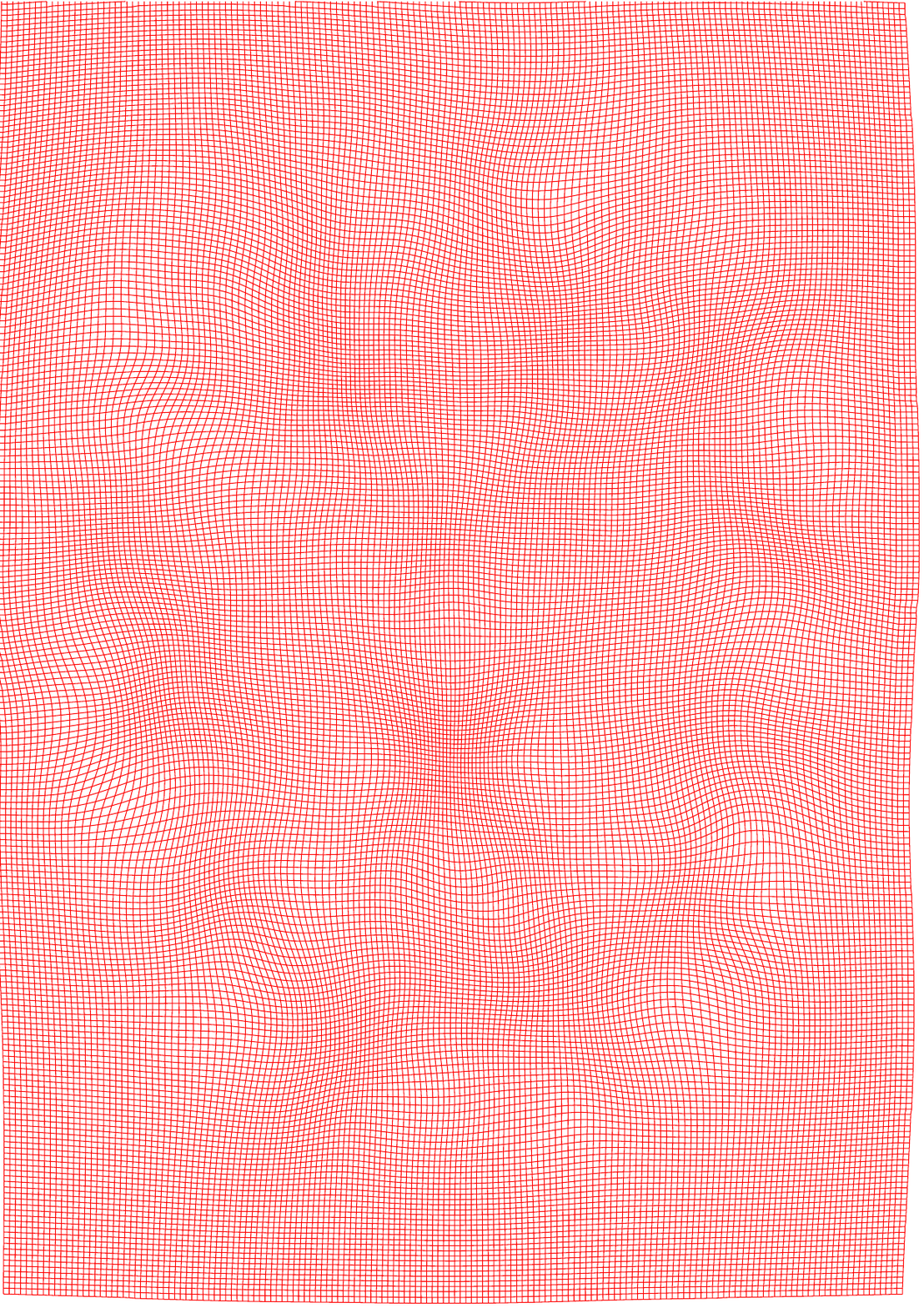}
\includegraphics[width=3.20cm,height=4.1cm]{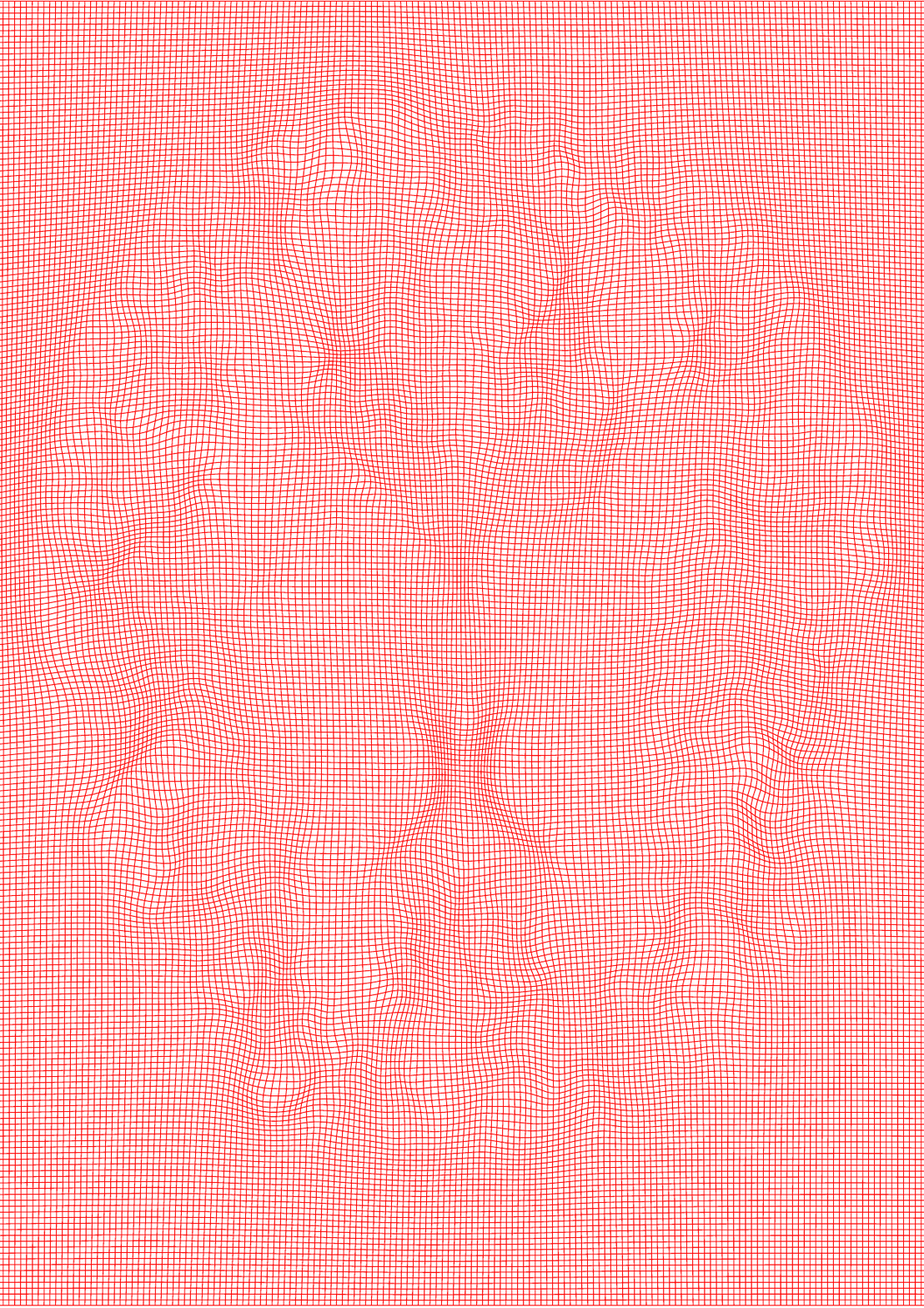}

\vspace{0.1cm}
\includegraphics[width=3.20cm,height=4.1cm]{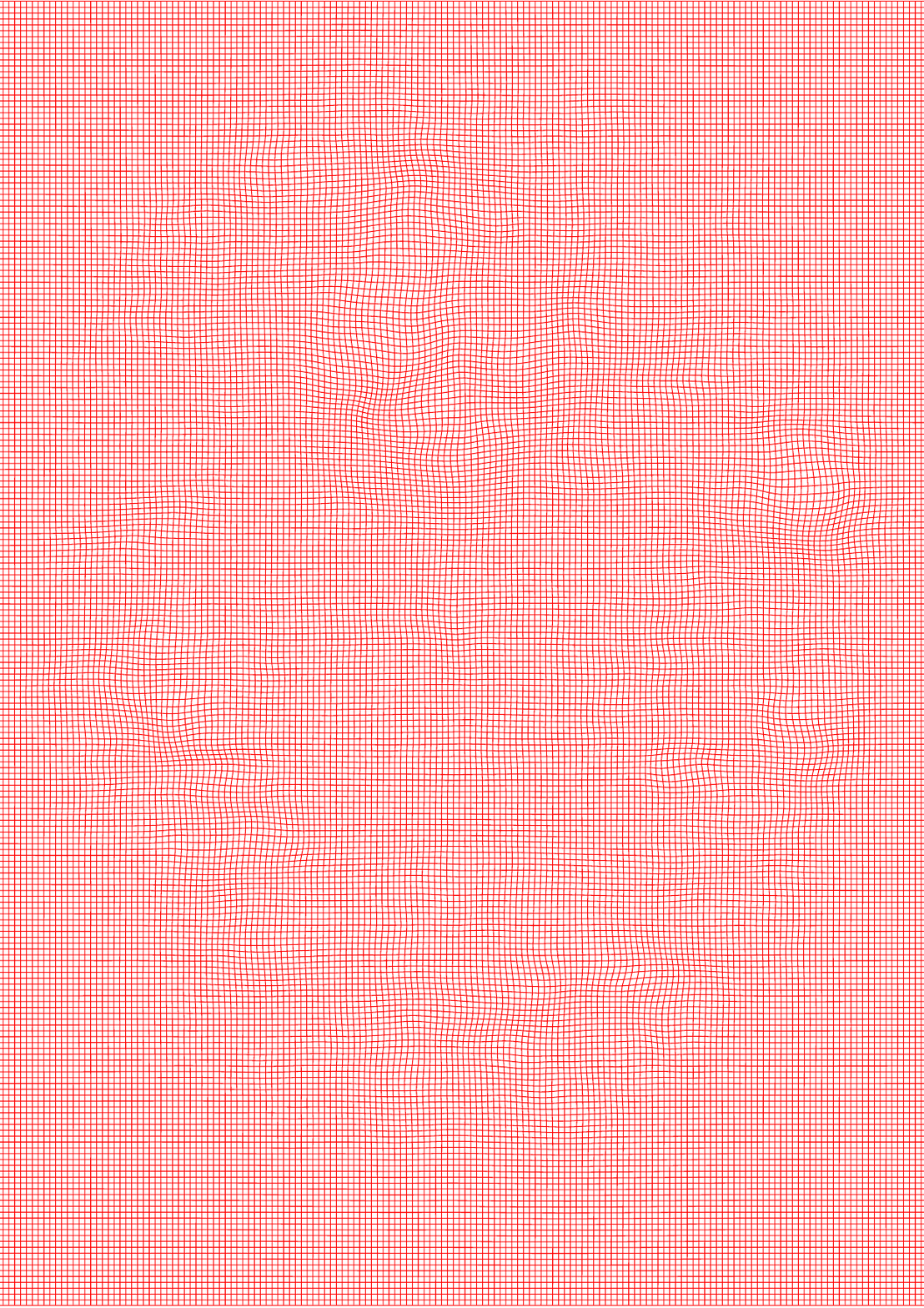}
\includegraphics[width=3.20cm,height=4.1cm]{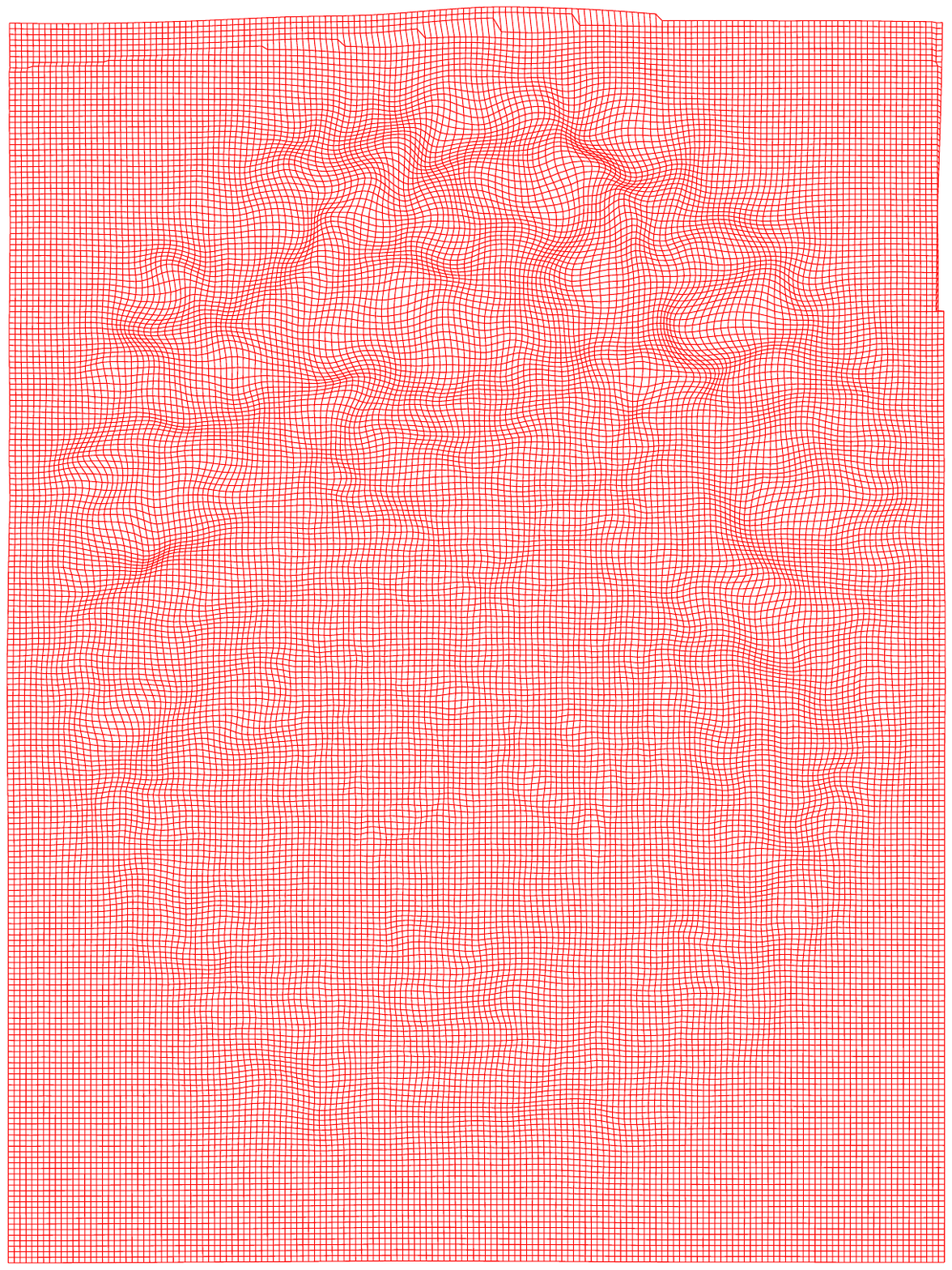}
\includegraphics[width=3.20cm,height=4.1cm]{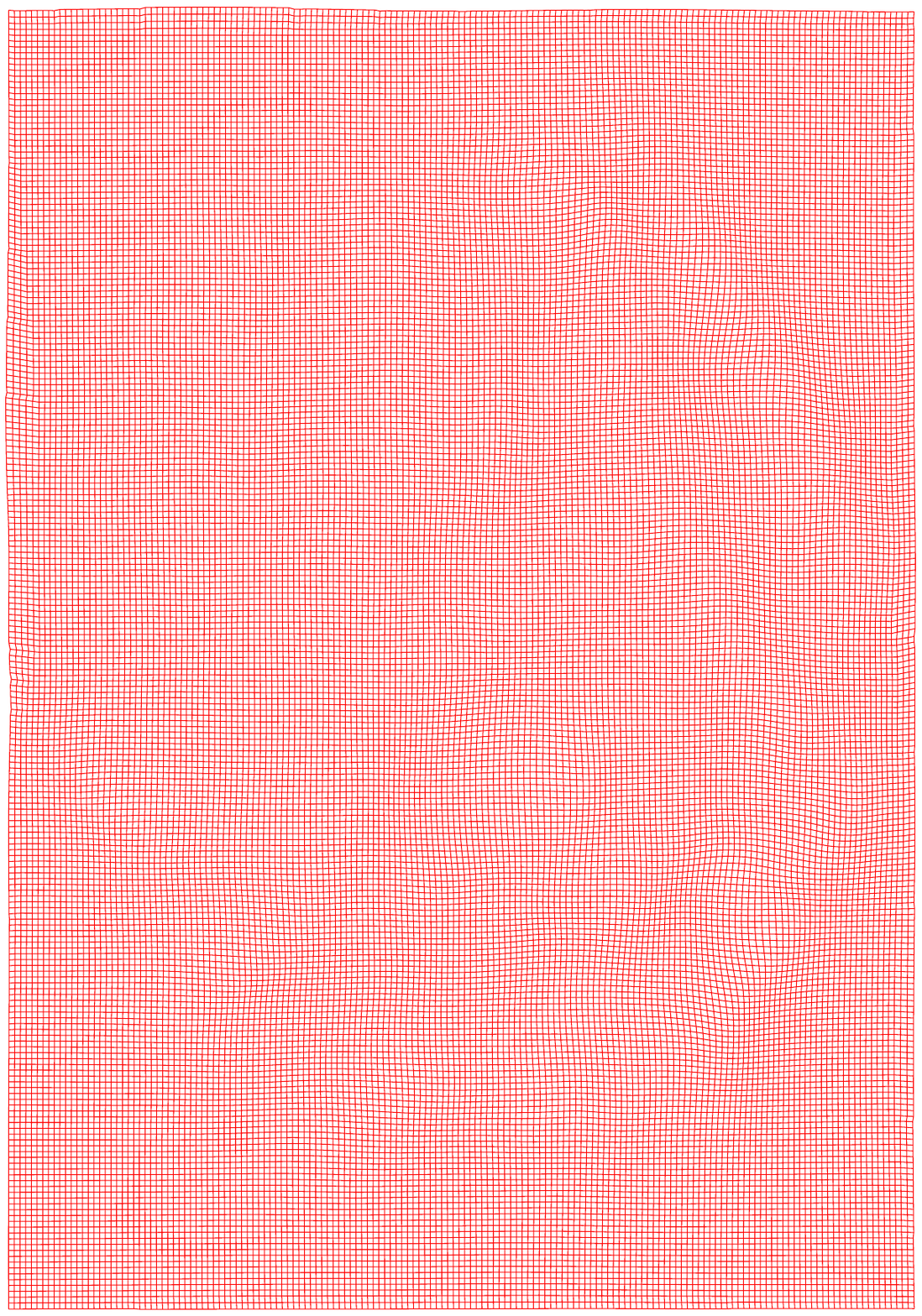}
\includegraphics[width=3.20cm,height=4.1cm]{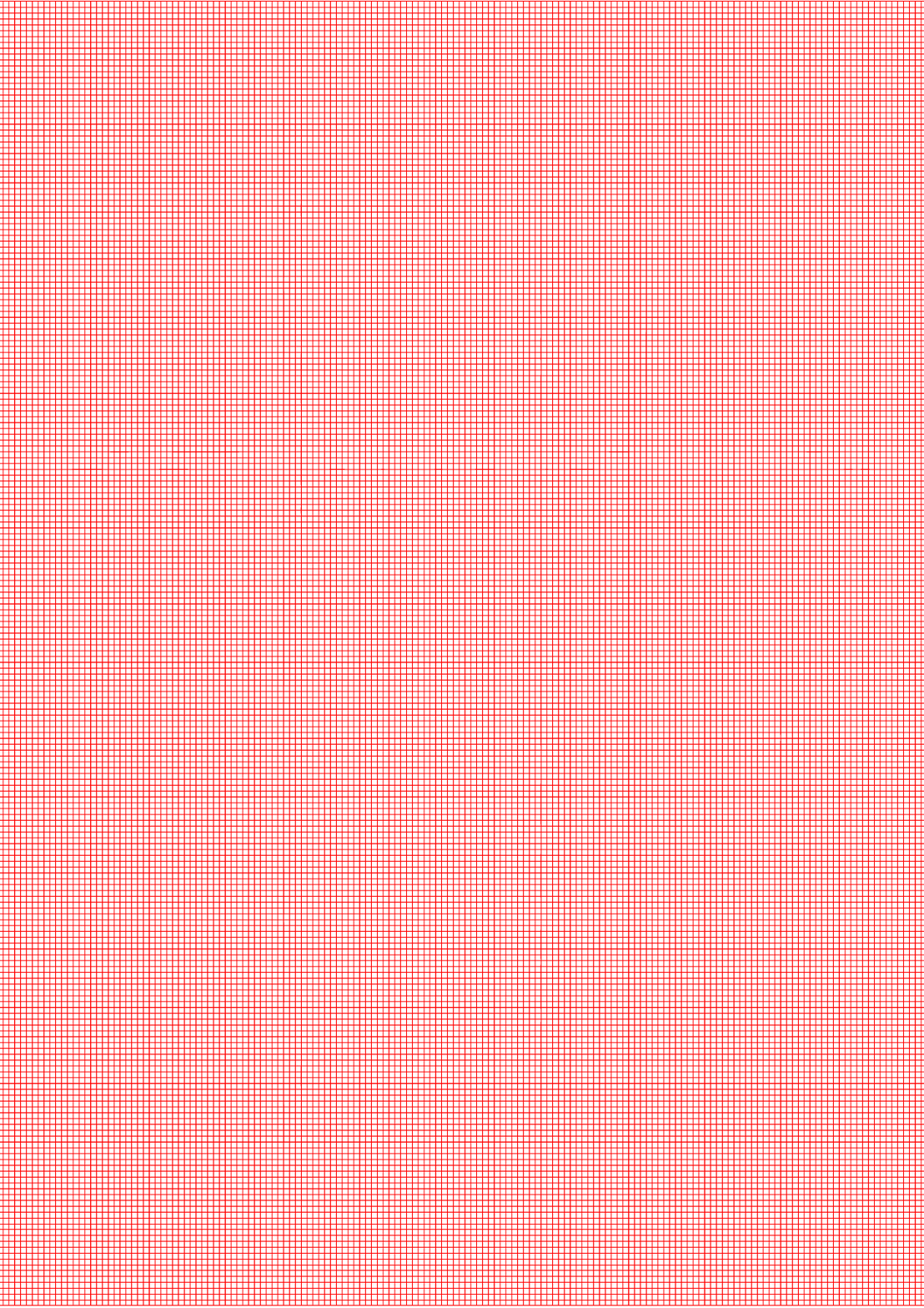}

\vspace{0.1cm}
\includegraphics[width=3.20cm,height=4.1cm]{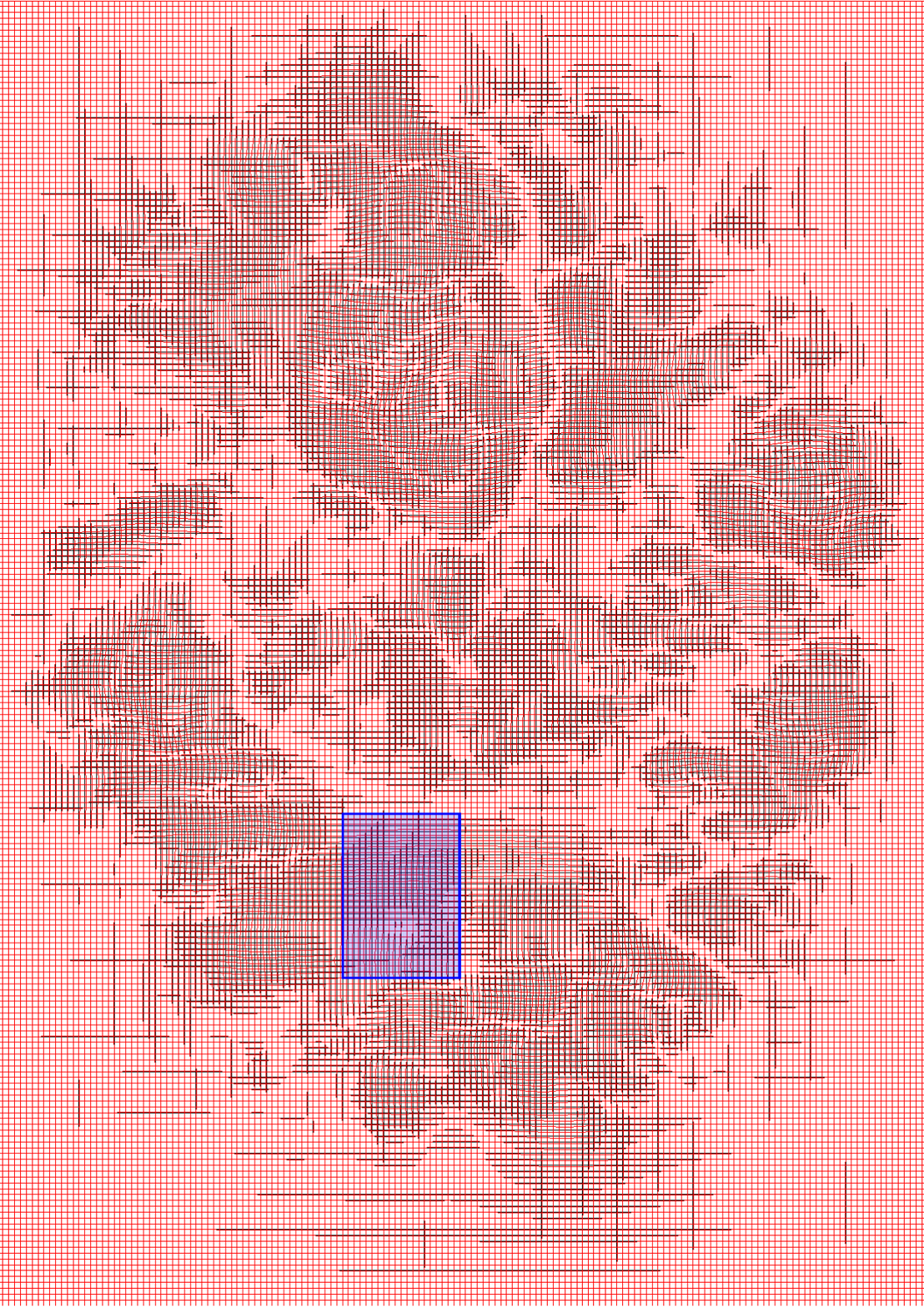}
\includegraphics[width=3.20cm,height=4.1cm]{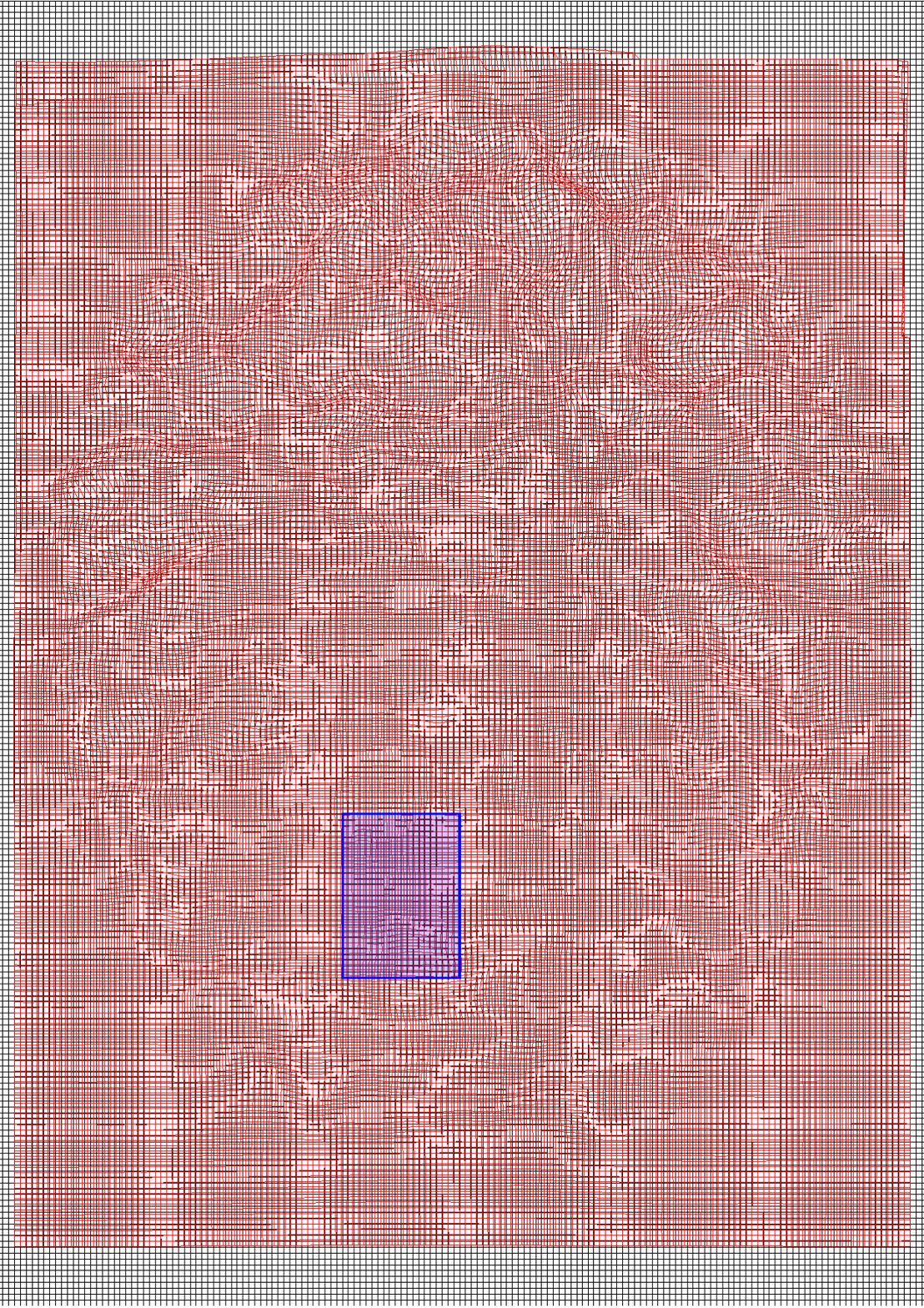}
\includegraphics[width=3.20cm,height=4.1cm]{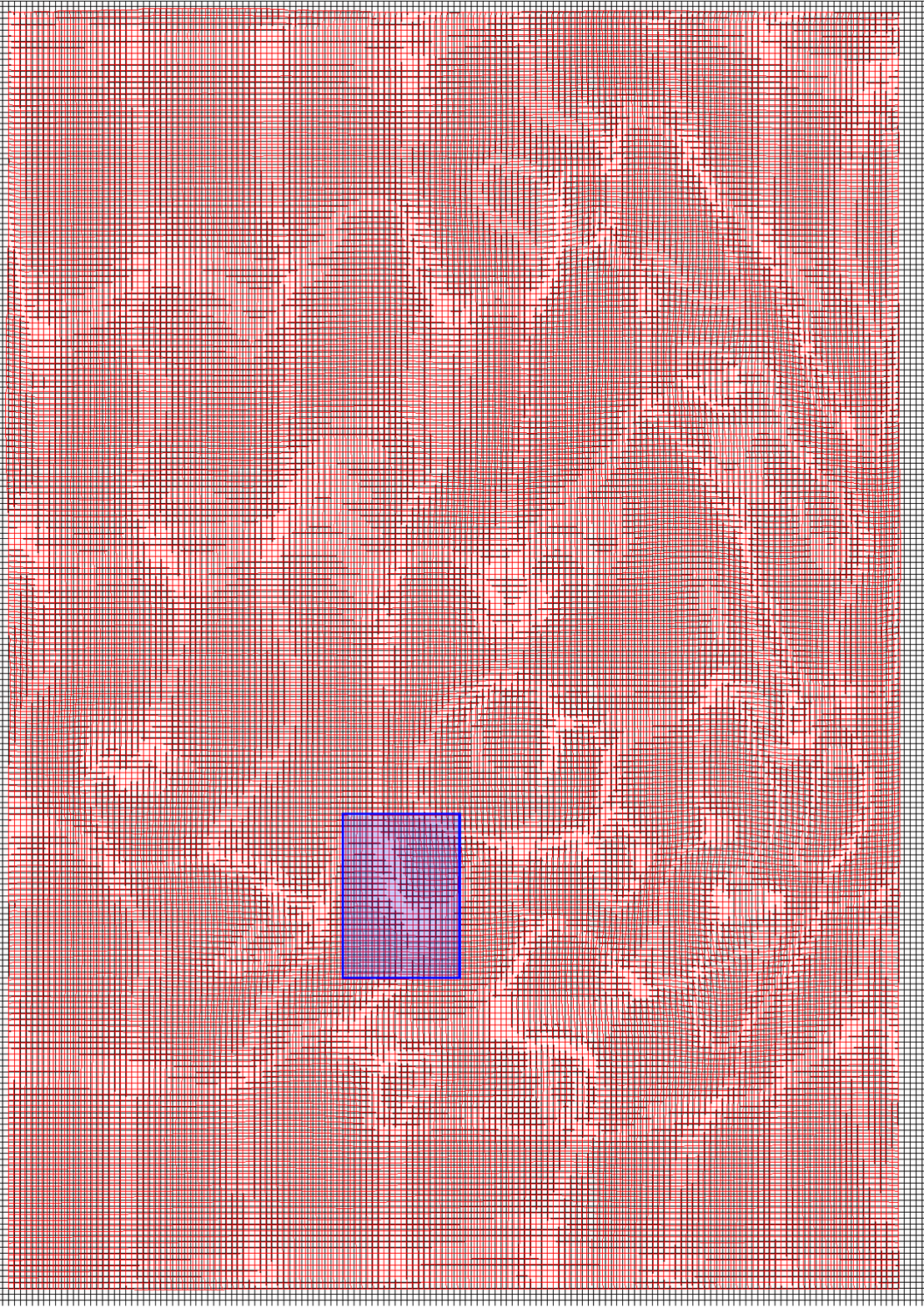}
\includegraphics[width=3.20cm,height=4.1cm]{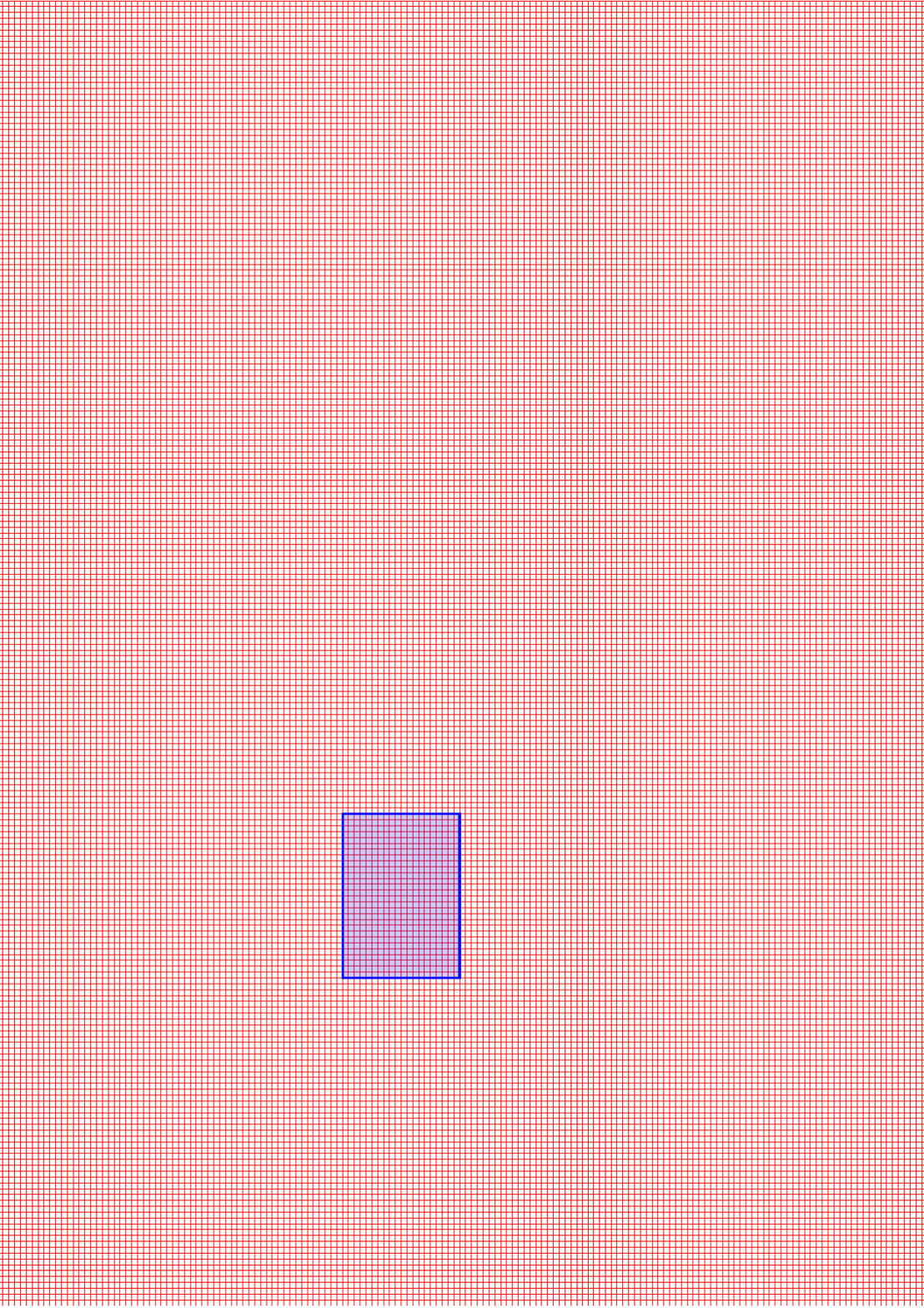}

\vspace{0.1cm}
\includegraphics[width=3.20cm,height=4.1cm]{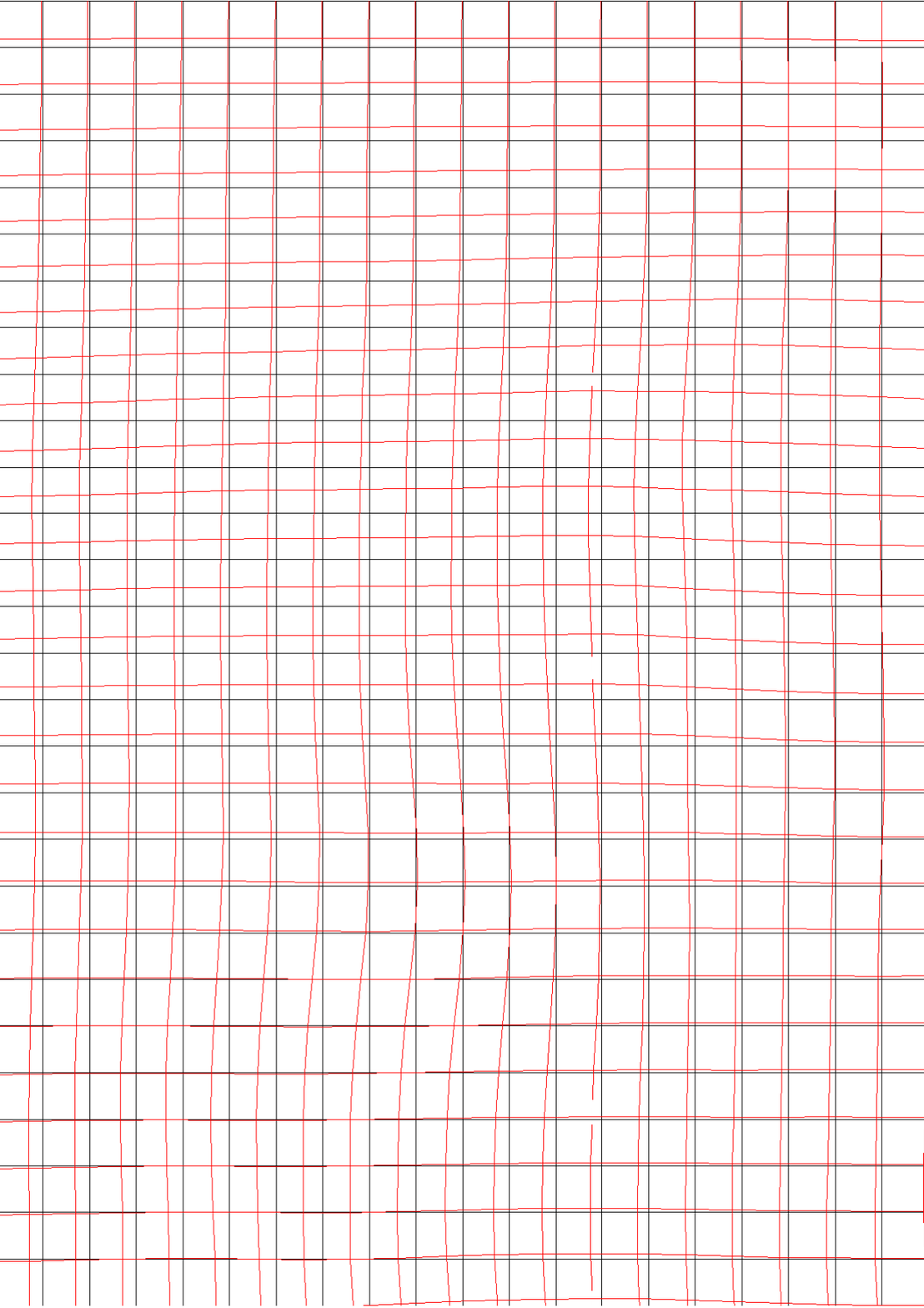}
\includegraphics[width=3.20cm,height=4.1cm]{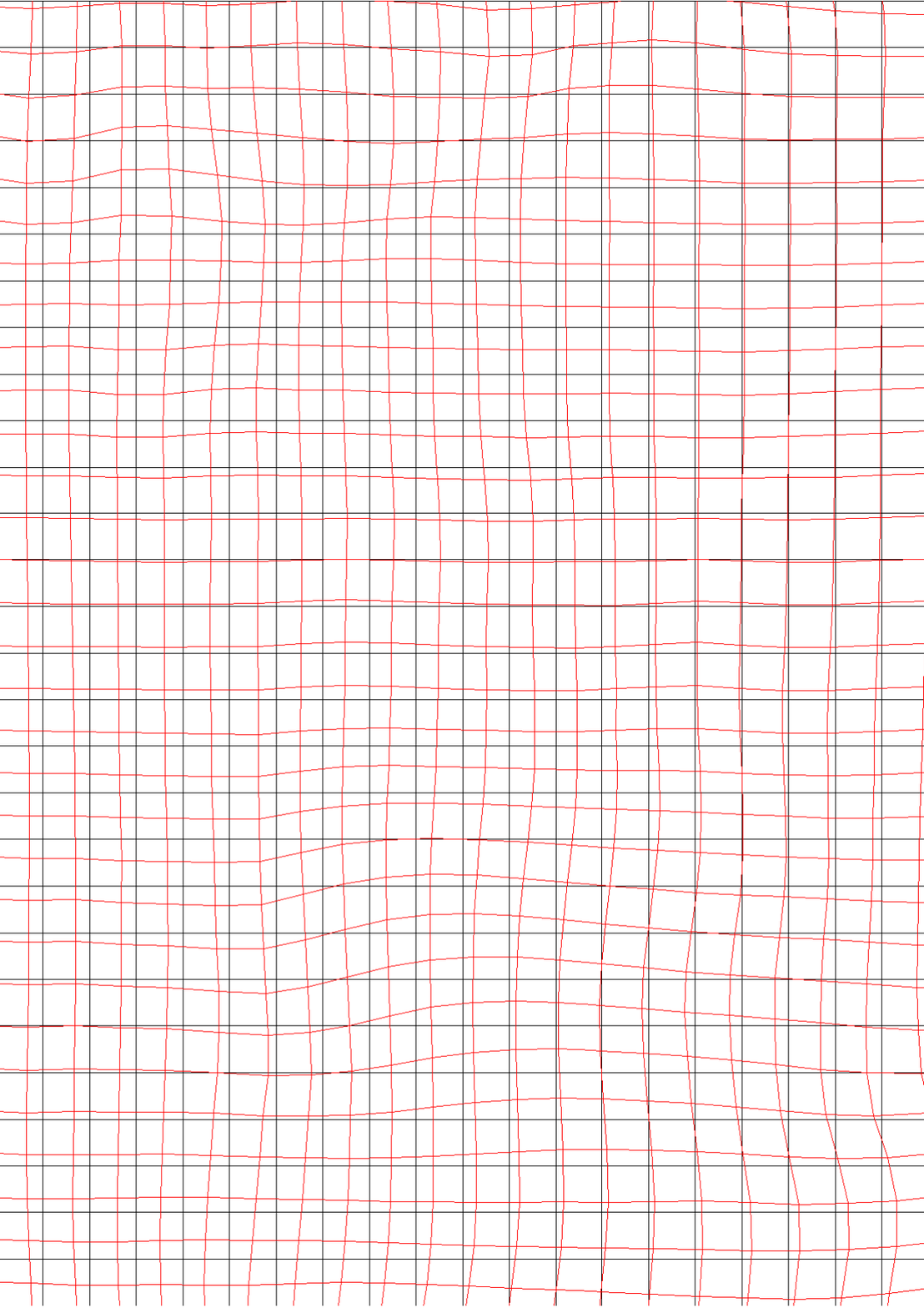}
\includegraphics[width=3.20cm,height=4.1cm]{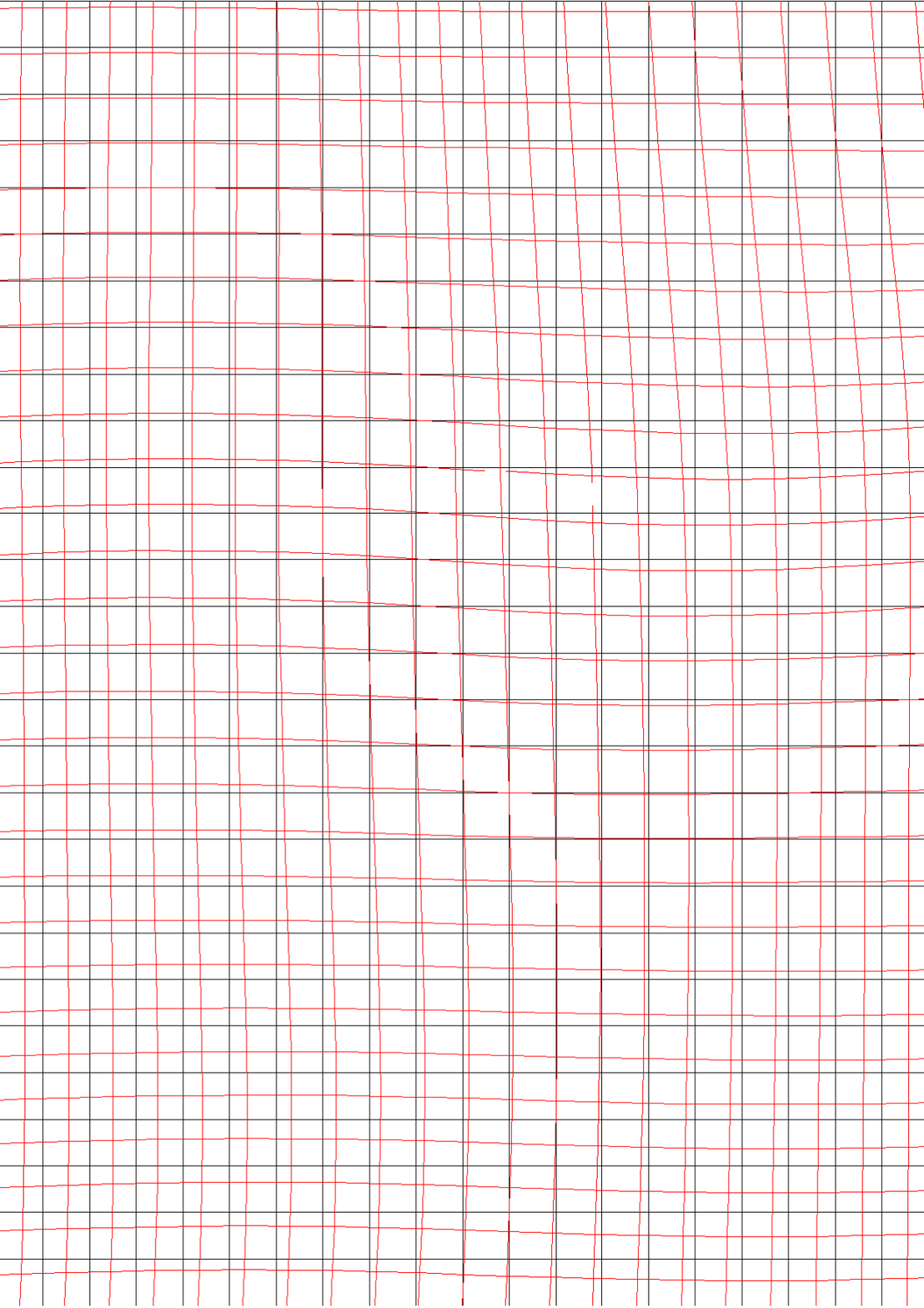}
\includegraphics[width=3.20cm,height=4.1cm]{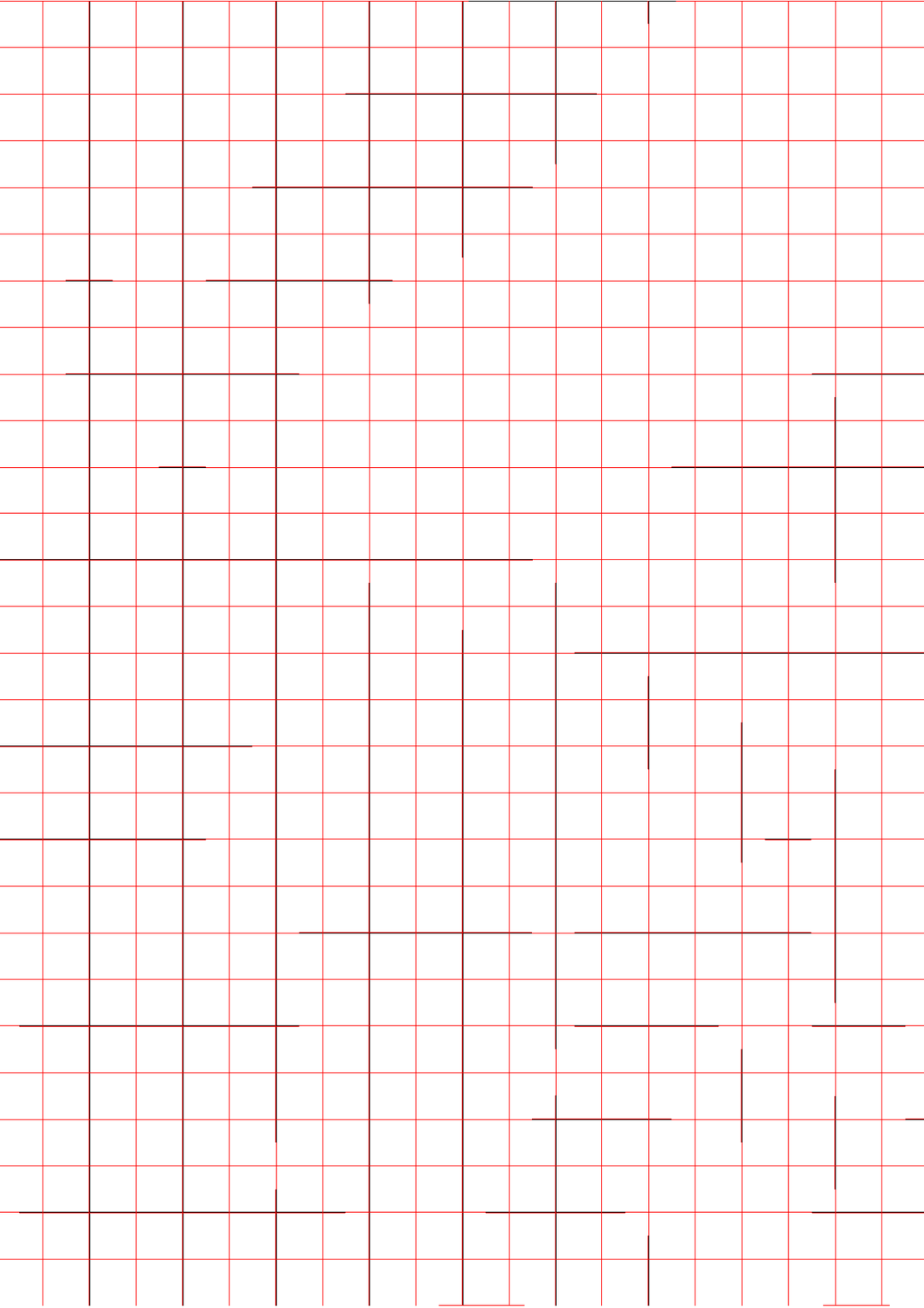}

\caption{Illustration of computed spatial transformations and their inverses. We show the 86-th axial slice for the registration of two representative images from the OASIS-1 dataset (registration from subject-0012 to subject-0140). From left to right (columns): ANTs, Easyreg, Fnirt and VPreg. From top to bottom (rows): ($i$) $\pmb{\phi}$, ($ii$) $\pmb{\phi}^{-1}$, ($iii$) $\pmb{\phi}^{-1}\circ\pmb{\phi}$, ($iv$)  $\pmb{\phi}^{-1}\circ\pmb{\phi}$ vs. $\pmb{\operatorname{id}}_{\Omega}$, and ($v$) zoom-in of row four at $(x,y,z) = (60:80,86,140:168)$ (blue rectangle area). For the comparison of the grids, we show red-grid-lines superimposed on black-grid-lines. Therefore, the less of a black grid is visible, the better the accuracy.}\label{f:mri3dtransvs}
\end{figure}

In a first step, we consider a pair of representative images of the OASIS-1 dataset. We show the registration results in \Cref{f:mri3dvs}.  We illustrate the computed spatial transformation in \Cref{f:mri3dtransvs}. We report the DICE scores for the four ROIs in \Cref{t:subj88converge}. We report additional performance measures in \Cref{t:subj88grid}. We report measures for the difference of the composition of the forward map and its inverse from identity in \Cref{t:subj88sym}.

\begin{table}
\centering\small
\caption{DICE for four ROIs for a representative pair of images from the OASIS-1 dataset. We report results for different registration methods (rows). The results correspond to those shown in~\Cref{f:mri3dvs}.}
\label{t:subj88converge}
\begin{tabular}{llrrrr}
\toprule
& & Cortex & subcortical GM & WM & CSF \\
\midrule
forward map
& ANTs     &   \decc{0.5710} &   \decc{0.8196} &   \decc{0.71032} &   \decc{0.67864}\\
& Easyreg  &   \decc{0.6120} &   \decc{0.8818} &   \decc{0.7591 } &   \decc{0.6767 }\\
& Fnirt    &   \decc{0.6344} &   \decc{0.8729} &   \decc{0.7779 } &   \decc{0.7758 }\\
& VPreg    &\bf\decc{0.6809} &\bf\decc{0.9043} &\bf\decc{0.8050 } &\bf\decc{0.8315 }\\
\midrule
inverse map
& ANTs    &   \decc{0.5710 } &   \decc{0.8178 } &   \decc{0.70949} &   \decc{0.6914 }\\
& Easyreg &   \decc{0.5798 } &   \decc{0.7876 } &   \decc{0.7231 } &   \decc{0.7271 }\\
& Fnirt   &   \decc{0.58211} &   \decc{0.81259} &   \decc{0.72422} &   \decc{0.64136}\\
& VPreg   &\bf\decc{0.6620 } &\bf\decc{0.9003 } &\bf\decc{0.7790 } &\bf\decc{0.8676 }\\
\bottomrule
\end{tabular}
\end{table}

\begin{table}
\centering\small
\caption{Performance measures for the registration of a pair of representative images from the OASIS-1 dataset (registration from subject-0012 to subject-0140). The results correspond to those shown in~\Cref{f:mri3dvs}.}
\label{t:subj88grid}
\begin{tabular}{llrrrrr}
\midrule
& & MSE-ratio & MI-Incr \% & min-JD & max-JD & \%JD$<0$ \\
\midrule
forward map
& ANTs     &    \decc{0.4454} &    \decc{28.94 }\% &   \decc{-0.048791} & \decc{0.67864} & $\approx$0\%     \\
& Easyreg  &    \decc{0.3979} & \bf\decc{33.33 }\% &   \decc{-3.5921  } & \decc{2.6914 } & \decc{0.18186}\% \\
& Fnirt    &    \decc{0.5714} &    \decc{24.485}\% & \decc{ 0.2498  } & \decc{2.9359 } & \bf 0\%          \\
& VPreg   & \bf\decc{0.3948} &    \decc{29.27 }\% &\bf\decc{ 0.06167 } & \decc{24.6421} & \bf 0\%          \\
\midrule
inverse map
& ANTs    &    \decc{0.4604} &    \decc{26.90}\% &    \decc{-0.041939} & \decc{0.6914} & $\approx$0\%       \\
& Easyreg & \bf\decc{0.4040} & \bf\decc{31.83}\% &    \decc{-4.7911  } & \decc{1.2284} & \decc{0.0353}\%    \\
& Fnirt   &    \decc{0.5360} &    \decc{20.44}\% & \decc{ 0.3108  } & \decc{3.5798} & \bf0\%             \\
& VPreg  &    \decc{0.4848} &    \decc{24.46}\% & \bf\decc{ 0.0219  } & \decc{8.70  } & \bf0\%             \\
\bottomrule
\end{tabular}
\end{table}

Based on the results for registration accuracy reported in \Cref{t:subj88converge} and \Cref{t:subj88grid}, we observe that all methods yield competitive results. This is qualitatively confirmed by the visualizations shown in \Cref{f:mri3dvs}. Overall, VPreg yields a slightly better DICE score for the four anatomical regions considered in this study. Easyreg yields the best MSE-ratio and MI-incr. Overall, we conclude that our method is competitive in terms of registration accuracy for this exemplary pair of images.

If we turn to properties of the deformation map, we can observe that VPreg yields well-behaved maps. In particular, VPreg yields diffeomorphic transformations and at the same time provides an accurate approximation of their inverse. This can be seen qualitatively in \Cref{f:mri3dtransvs} and, more importantly, is quantified in \Cref{t:subj88grid} and \Cref{t:subj88sym}. The maps generated by ANTs and Easyreg are not diffeomorphic as judged by the minimal value of the determinant of the Jacobian of the transformation in \Cref{t:subj88grid}. Those generated by Fnirt are. We emphasize that the numerical guarantees on generating diffeomorphic maps VPreg provides do not negatively affect registration accuracy as confirmed by the DICE scores reported in \Cref{t:subj88converge}. These observations are also confirmed by the visualizations shown in~\cref{f:mri3dtransvs}.

Turning to the accuracy for computing the inverse map, we can observe that VPreg does yield inverses that are most consistent with the forward map. This is confirmed by the results reported in \Cref{t:subj88sym}; the composition of the computed map and its inverse $\pmb{\phi}\circ\pmb{\phi}^{-1}$ is much closer to $\pmb{\operatorname{id}}_{\Omega}$ for VPreg than for the other methods.

\begin{table}
\centering\small
\caption{Error in the computation of the inverse map. We report measures that quantify the distance of the computed transformation composed with its inverse from identity, i.e., we verify how close $\pmb{\phi} \circ\pmb{\phi}^{-1}$ is to $\pmb{\operatorname{id}}_{\Omega}$. The number reported third column is the average error (total distance normalized by the number of voxels $|\Omega|$). The results correspond to those shown in~\Cref{f:mri3dtransvs}.}
\label{t:subj88sym}
\begin{tabular}{lrcr}
\toprule
& $\max|$det$\nabla(\pmb{\phi}^{-1}\circ\pmb{\phi})-1|$ & Sum$=\sum|$det$\nabla(\pmb{\phi}^{-1}\circ\pmb{\phi})-1|$& Sum$/|\Omega|$  \\
\midrule
    ANTs    &     \sci{1.2954} &     \sci{1.71e5}&     \sci{2.49e-1} \\
    Easyreg &     \sci{3.3597} &     \sci{1.24e6}&     \sci{1.80e-1}\\
    Fnirt   & \bf \sci{0.5703} &     \sci{6.36e5}&     \sci{9.24e-2} \\
    VPreg   &     \sci{0.6755} & \bf \sci{2.29e2}&    \bf \sci{3.33e-5} \\
\midrule
& max$\|\pmb{\phi}^{-1}\circ\pmb{\phi}-\pmb{\operatorname{id}}_{\Omega}\|$ & Sum$=\sum\|\pmb{\phi}^{-1}\circ\pmb{\phi}-\pmb{\operatorname{id}}_{\Omega}\|$ & Sum$/|\Omega|$\\
\midrule
    ANTs    &     \sci{4.1782}  & \sci{1.16e6}&     \sci{1.17e-1}\\
    Easyreg &     \sci{12.7173} & \sci{4.13e7}&     \sci{6.00e0} \\
    Fnirt   &     \sci{6.3133}  & \sci{2.11e7}&     \sci{3.06e0} \\
    VPreg   & \bf \sci{1.9002}  & \bf \sci{3.47e2}&    \bf \sci{5.04e-5} \\
\bottomrule
\end{tabular}
\end{table}

\subsection{Cohort Study}\label{s:cohort}

Here, we use the first five images of the OASIS-1~\cite{marcus2007:oasis, oasis-web} dataset as the moving image and the next 30 images as the fixed image. We perform 150 distinct registrations using all methods considered in this study.

\begin{figure}
\centering
\includegraphics[width=1.0\textwidth]{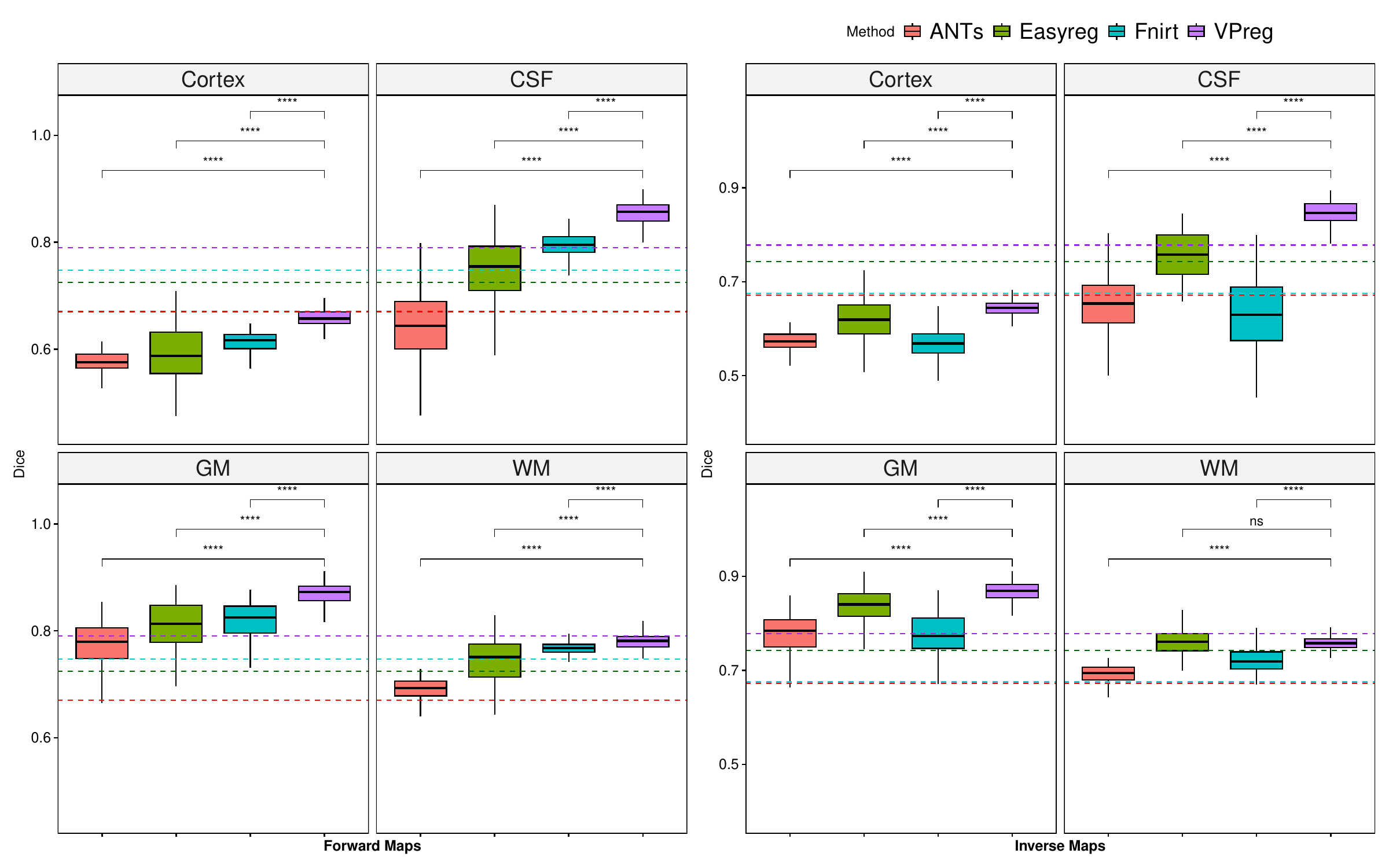}
\caption{Box-whisker plots for the DICE scores for four ROIs of the OASIS-1 dataset summarizing 150 registrations. We report the statistics (box-whisker plots) for ROIs (Cortex, CSF, WM, and GM) for the forward map (left block) and the inverse map (right block). In each plot, we include results for all methods considered in this study: ANTs (red), Easyreg (green), Fnirt (turquoise), and VPreg (purple). A DICE score of one indicates that the anatomical regions are in perfect agreement. A score of zero indicates that the structures are not aligned at all. The horizontal dashed lines are the overall average DICE scores with ANTs (red), Easyreg (green), Fnirt (turquoise), and VPreg (purple), respectively.}
\label{f:seg4}
\end{figure}

\begin{figure}
\centering
\includegraphics[width=1.0\textwidth,height=8.0cm]{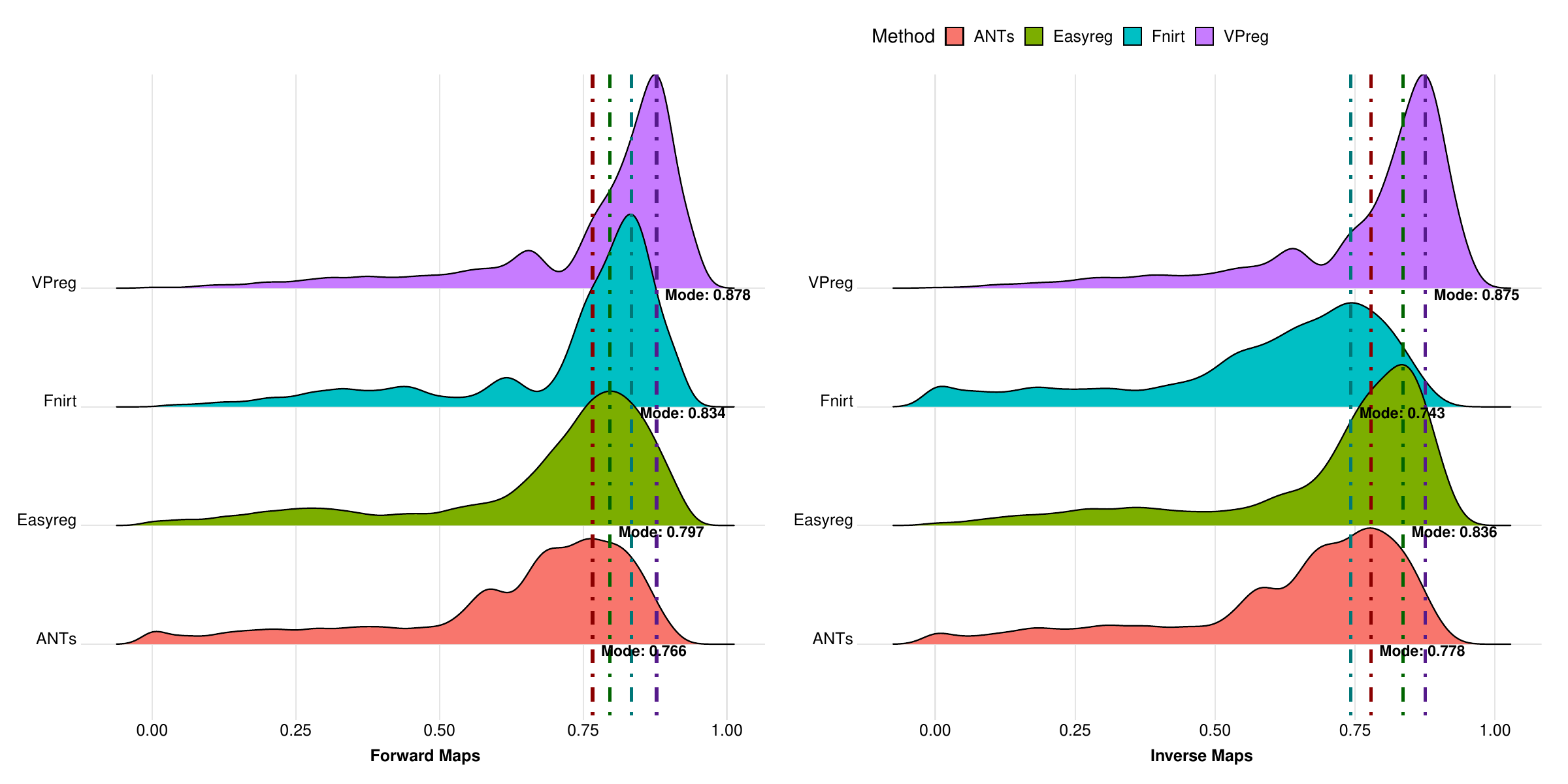}
\caption{We report densities (histograms) for the DICE score for all 35 labels available in the OASIS-1 dataset. We summarize results for 150 registrations. We show results for the forward map on the left and results for the inverse map on the left. Each plot includes the densities obtained for (from top to bottom) VPreg (purple), Fnirt (turquoise), Easyreg (green), and ANTs (red). A score of 1 indicates a perfect alignment. A score of zero indicates that the structures are not aligned at all. The mode values (VPreg: 0.878; Fnirt:0.834; EasyReg:0.797; ANTs: 0.766) are marked by the vertical dot-dashed lines for VPreg (dark purple), Fnirt (dark turquoise), Easyreg (dark green), and ANTs (dark red).}
\label{f:seg35_dens}
\end{figure}

\begin{figure}
\centering
\includegraphics[width=0.97\textwidth,height=0.97\textheight]{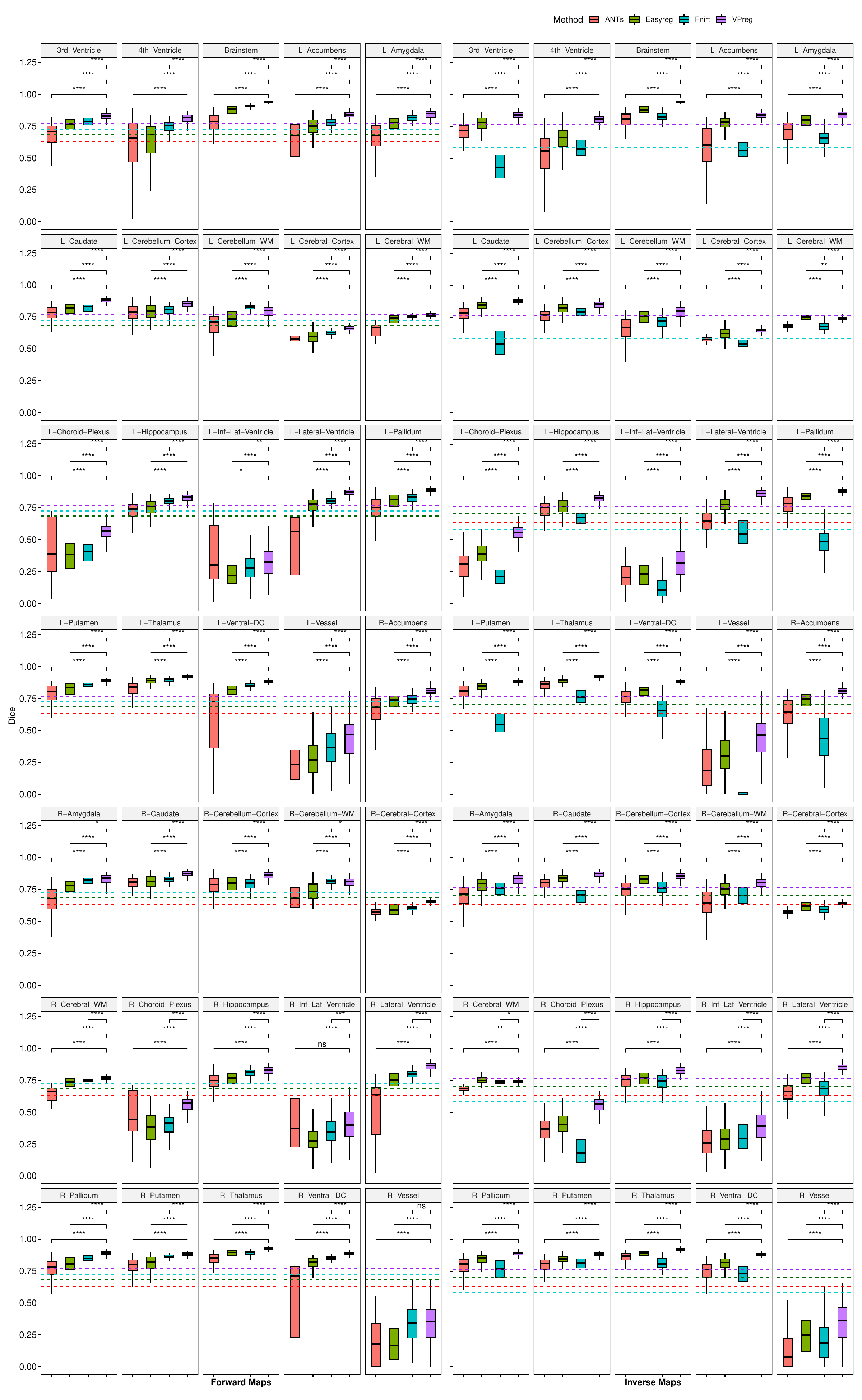}
\caption{Box-whisker plots for the DICE scores for 35 ROIs of the OASIS-1 dataset summarizing 150 registrations. We report statistics for 35 ROIs for the forward map (left block) and the inverse map (right block). A DICE score of one indicates that the anatomical regions are in perfect agreement. A score of zero indicates that the structures are not aligned at all. The horizontal dashed lines are the overall average DICE scores with ANTs (red), Easyreg (green), Fnirt (turquoise), and VPreg (purple), respectively.}
\label{f:seg35}
\end{figure}

\begin{figure}
\centering
\includegraphics[width=1.0\textwidth]{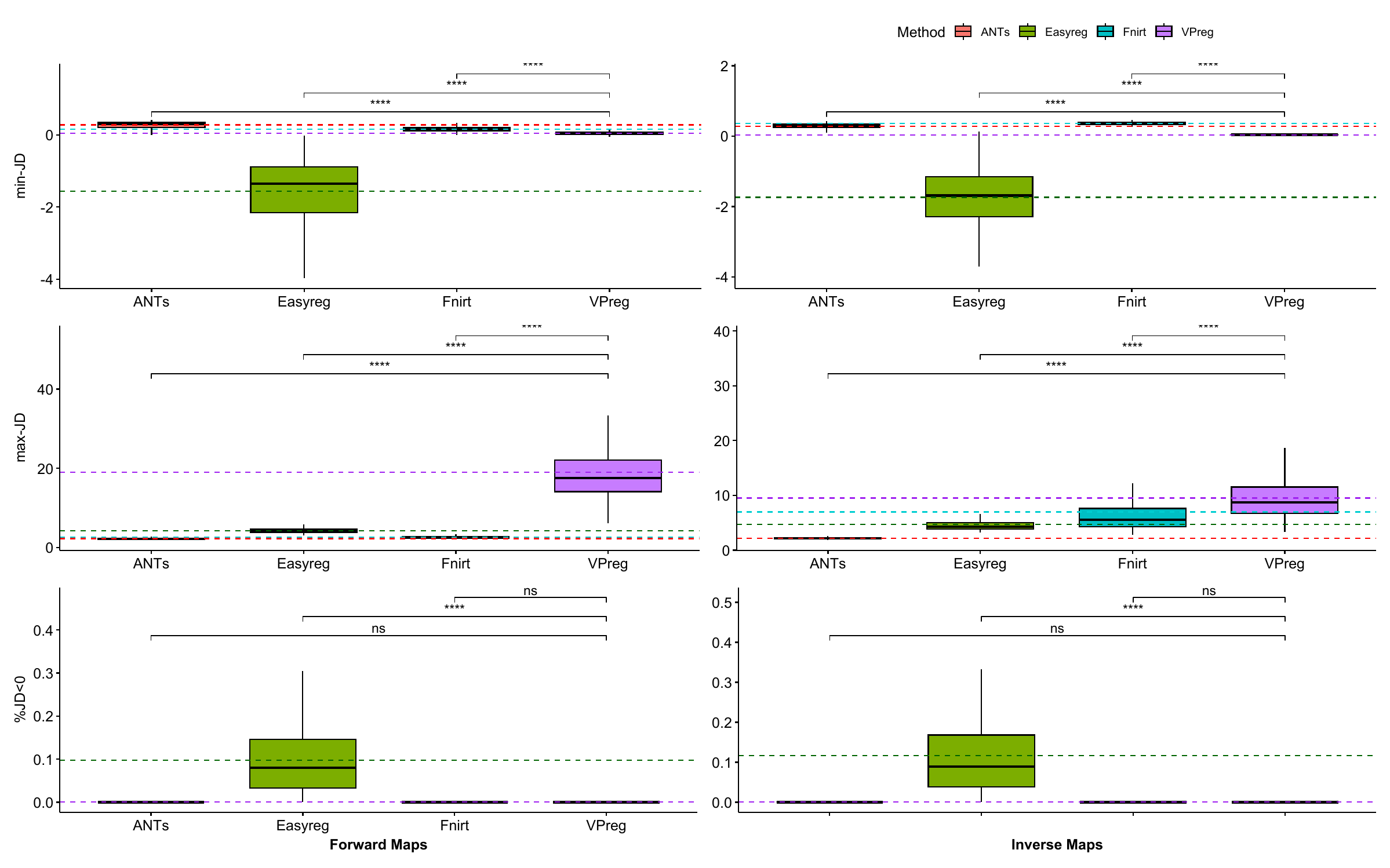}
\caption{Box-whisker plots for the values of the determinant of the Jacobian of the spatial transformations (left) and their inverses (right). We report averages for 150 registrations. Top row: Minimal values. Middle row: Maximum value. Bottom row: Percentage of voxels that have a negative Jacobian determinant. The horizontal dashed lines are the overall average DICE scores with ANTs (red), Easyreg (green), Fnirt (turquoise), and VPreg (purple), respectively.}
\label{f:distrijd_sym}
\end{figure}

\begin{figure}
\centering
\includegraphics[width=1.0\textwidth]{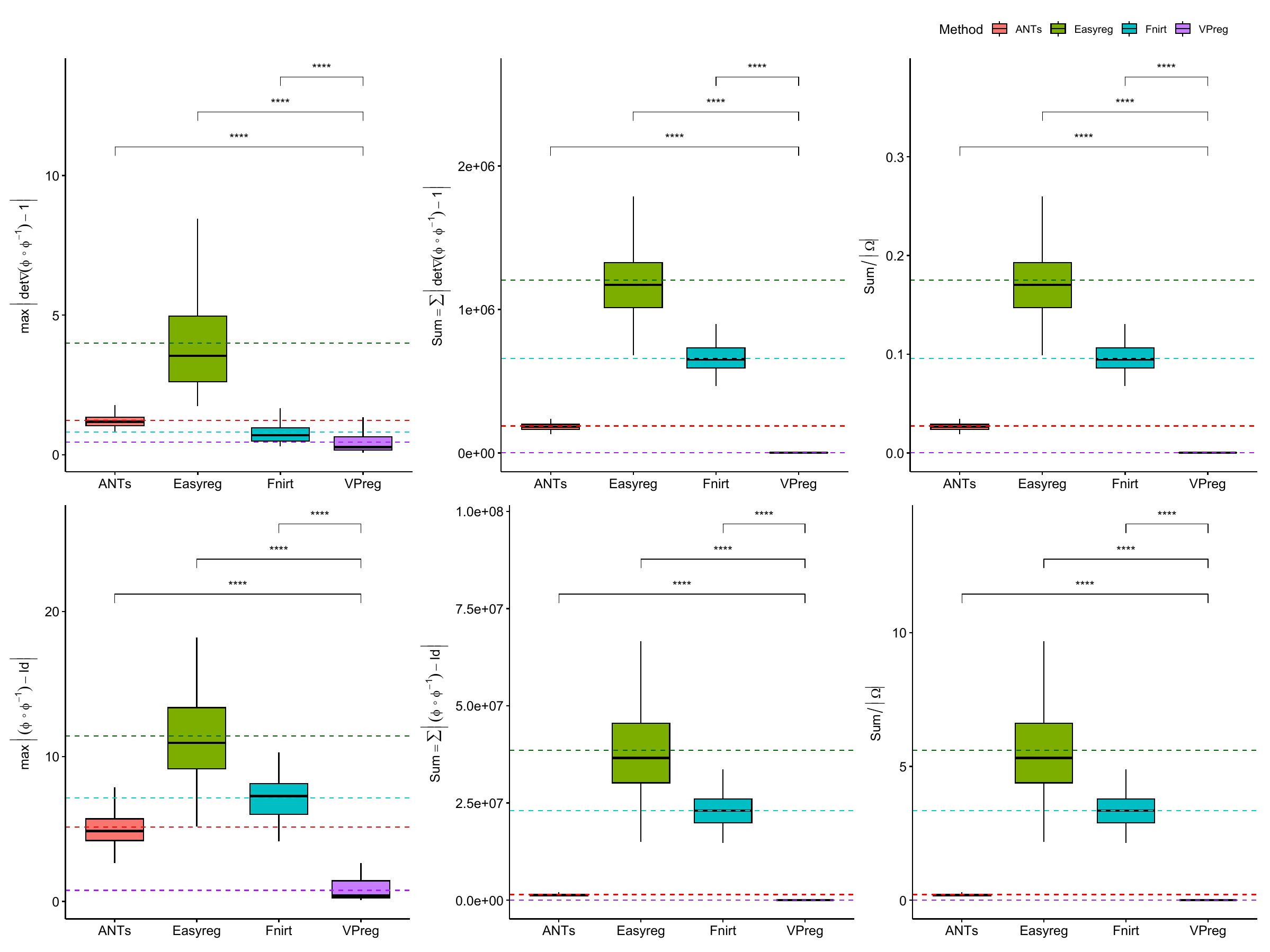}
\caption{Box-whisker plots for the accuracy of the inverse $\pmb{\phi}^{-1}$ of the computed spatial transformation $\pmb{\phi}$. We report results for all methods considered in this work. We report averages for 150 registrations. The top row compares the determinant of the Jacobian of $\pmb{\phi} \circ \pmb{\phi}^{-1}$ to the Jacobian of $\pmb{\operatorname{id}}_{\Omega}$ (maximum values on the left and summation on the right). The bottom row computes the norm between $\pmb{\phi} \circ \pmb{\phi}^{-1}$ and $\pmb{\operatorname{id}}_{\Omega}$. The horizontal dashed lines are the overall average DICE scores with ANTs (red), Easyreg (green), Fnirt (turquoise), and VPreg (purple), respectively.}
\label{f:inverse_map_accuracy}
\end{figure}

We report the statistics of the DICE score for each individual method for four ROIs in \Cref{f:seg4}. We report the densities (histograms) of the DICE score for all 35 regions available in the OASIS-1 data for each individual method in \Cref{f:seg35_dens}. We show box-whisker plot for the DICE score for these 35 regions in \Cref{f:seg35}. We report statistics for the determinant of the Jacobian of the computed registration maps in \Cref{f:distrijd_sym}. We show box-whisker plots for the minimal and maximal values of the determinant of the Jacobian of the transformation as well as the percentage of voxels for which the Jacobian was non-positive (indicating folding). We report values for the associated mean and standard deviations in \Cref{t:distrijd}. We show box-whisker plots to assess the accuracy of the inverse map in \Cref{f:inverse_map_accuracy}. Average values associated with these plots are reported in \Cref{t:sym}.

Based on the results reported in \Cref{f:seg4}, \Cref{f:seg35_dens} and \Cref{f:seg35} we can see that VPreg overall outperforms all methods in terms of the DICE score. The only exception is WM for the inverse transformation; here, the proposed method performs as good as Easyreg (t-test result shows that the two distributions of performance are \emph{not} significantly different). In \Cref{f:distrijd_sym} we can observe that Easyreg produces the most irregular transformations. The remainder of the methods perform quite similar with the exception of the max value; VPreg generates larger maximum values for JD, which indicates VPreg allows a wider range of deformation while keeping the diffeomorphic property. These observations are confirmed by the averages reported in \Cref{t:distrijd}. Based on the results reported in \Cref{f:inverse_map_accuracy} we conclude that VPreg provides the most accurate approximation to the inverse of the computed spatial transformation of all methods considered in the present work. These observations are confirmed by the numbers reported in \Cref{t:sym}.

In \Cref{f:seg4}, \Cref{f:seg35}, \Cref{f:distrijd_sym} and \Cref{f:inverse_map_accuracy}, the significant t-test on group difference is noted as *($P <0.05$); **($P <0.01$); ***($P <0.001$); ****($P <0.0001$); and in the case of not significant, ns($P >0.05$).

\begin{table}
\centering\small
\caption{We report values for the determinant of the Jacobian of the spatial transformation to assess regularity of the computed spatial transformation. We report results averaged across 150 registrations (mean and standard-deviation in bracktes). We include (from left to right) values for the minimum and maximum determinant of the Jacobian of the transformation as well as the percentage of voxels with non-positive values for the determinant of the Jacobian. We include results for all methods considered in this study (rows).}
\label{t:distrijd}
\begin{tabular}{lrrr}
\toprule
& min-JD & max-JD & \textbf{\%}JD$<0$  \\
\midrule
ANTs    & \decc{ 0.27}(\decc{0.10}) & \decc{ 2.17}(\decc{0.20}) & \sci{2.60e-6}(\sci{4.64e-5}) \\
Easyreg & \decc{-1.65}(\decc{0.89}) & \decc{ 4.44}(\decc{1.35}) & \sci{1.07e-1}(\sci{8.42e-2}) \\
Fnirt   & \decc{ 0.14}(\decc{0.07}) & \decc{ 2.57}(\decc{0.43}) & \bf \sci{2.91e-7}(\sci{3.74e-6}) \\
VPreg   & \decc{ 0.04}(\decc{0.04}) & \decc{14.25}(\decc{7.70}) & \sci{2.52e-6}(\sci{2.02e-5}) \\
\bottomrule
\end{tabular}
\end{table}

\begin{table}
\centering\small
\caption{We report values for the deviation $\pmb{\phi}^{-1}\circ\pmb{\phi}$ from $\pmb{\operatorname{id}}_{\Omega}$, i.e., the accuracy and consistency of the computed inverse spatial transformation $\pmb{\phi}^{-1}$ with the forward map $\pmb{\phi}$. We report values averaged across 150 registration (mean and standard deviation in brackets).}
\label{t:sym}
\begin{tabular}{lrrr}
\toprule
& max$|$det$\nabla(\pmb{\phi}^{-1}\circ\pmb{\phi})-1|$ & Sum$=\sum|$det$\nabla(\pmb{\phi}^{-1}\circ\pmb{\phi})-1|$ &Sum$/|\Omega|$\\
\midrule
ANTs    &    \decc{1.23}(\decc{0.30}) &     \sci{1.88e5}(\sci{3.45e4})&     \sci{2.73e-2}(\sci{5.03e-3})\\
Easyreg &    \decc{3.99}(\decc{1.66}) &     \sci{1.20e6}(\sci{2.58e5})&     \sci{1.73e-1}(\sci{3.93e-2})\\
Fnirt   &    \decc{2.19}(\decc{0.73}) &     \sci{1.19e6}(\sci{9.07e4})&     \sci{9.64e-2}(\sci{1.61e-2})\\
VPreg   & \bf\decc{0.44}(\decc{0.37}) & \bf \sci{5.09e2}(\sci{2.34e2})&    \bf \sci{7.18e-4}(\sci{7.90e-3})\\
\midrule
& max$\|\pmb{\phi}^{-1}\circ\pmb{\phi}-\pmb{\operatorname{id}}_{\Omega}\|$ & Sum$=\sum\|\pmb{\phi}^{-1}\circ\pmb{\phi}-\pmb{\operatorname{id}}_{\Omega}\|$ &Sum$/|\Omega|$\\
\midrule
ANTs    &     \decc{ 5.15}(\decc{1.40}) &     \sci{1.44e6}(\sci{4.20e5})&     \sci{2.09e-1}(\sci{6.118e-2})\\
Easyreg &     \decc{11.42}(\decc{2.99}) &     \sci{3.85e7}(\sci{1.15e7})&     \sci{5.55e0}(\sci{1.717e0})\\
Fnirt   &     \decc{17.40}(\decc{3.28}) &     \sci{3.74e7}(\sci{2.81e6})&     \sci{3.37e0}(\sci{7.272e1})\\
VPreg   & \bf \decc{ 0.77}(\decc{0.71}) & \bf \sci{1.54e3}(\sci{1.00e3})&   \bf\sci{2.20e-2}(\sci{2.668e-1})\\
\bottomrule
\end{tabular}
\end{table}

\section{Conclusions}\label{s:conclusion}

In this paper we introduced a novel diffeomorphic image registration method that is based on the variational principle grid generation method~\cite{zhou2023:vpgrid,zhou2024:construction}. This work extends on our past research on diffeomorphic image registration~\cite{zhou2022:recent} in several ways. Aside from extending our methodology, we also include additional material for our past work not presented elsewhere. Our framework allows us to precisely control the properties of the diffeomorphic transformation by controlling JD and its curl. The most important observations of this work are as follows:
\begin{itemize}
\item VPreg generates smooth diffeomorphic transformations with a controlled JD of the spatial transformation without sacrificing registration accuracy.
\item VPreg provides maps that are guaranteed to be diffeomorphic as judged by the values of JD.
\item VPreg yields a registration accuracy that is competitive with existing methods. It gave a DICE score that is superior to all methods considered in the present work.
\item VPreg does not only provide the forward map but also its inverse. The computed inverse is consistent with the forward map; our numerical results suggest that the inverses produces by our approach are significantly more accurate than those generated by existing methods.
\end{itemize}

With these properties, VPreg is an excellent candidate for neuroimaging VBM, DBM and TBM studies. VPreg is currently based on a Matlab prototype implementation. Our next step is to port the software to another programming language to speed up the time-to-solution.

\begin{appendix}

\section{Corresponding Mathematical Derivations and Algorithms in pseudo-code}\label{s:append}

In this appendix, we derive necessary conditions for the different problem formulations that either predate or constitute components in VPreg. These derivations are formal only. We also include the settings used for the baseline methods considered in our work.

\subsection{Settings and Parameters for Baseline Methods}\label{a:baseline}

All baseline methods are implemented using generic settings and parameters from their tutorials, where only the non-linear registration mode is activated. Sample codes that are used for comparisons are listed in the following.
\begin{itemize}
\setlength{\itemindent}{-.3in}
\item ANTs (\hyperref{https://github.com/ANTsX/ANTs/wiki/Forward-and-inverse-warps-for-warping-images,-pointsets-and-Jacobians}{link}{}{\textcolor{blue}{ANTs}}):
\begin{lstlisting}[style=bashstyle]
antsRegistrationSyN.sh -d 3 -g 0.05 -m mov_img -f fixed_img -t so -o moved_img
\end{lstlisting}

\item Easyreg (\hyperref{https://surfer.nmr.mgh.harvard.edu/fswiki/EasyReg}{link}{}{\textcolor{blue}{FreeSurfer}}):
\begin{lstlisting}[style=bashstyle]
mri_easyreg --ref fixed_img --ref_seg fixed_img_seg \
       --ref_reg inversed_fixed_img_seg --fwd_field forward \
       --flo mov_img  --flo_seg mov_img_seg \
       --flo_reg moved_img --bak_field inverse --threads 12 \
\end{lstlisting}

\item Fnirt (\hyperref{https://web.mit.edu/fsl_v5.0.10/fsl/doc/wiki/FNIRT(2f)UserGuide.html}{link}{}{\textcolor{blue}{FSL}}):
\begin{lstlisting}[style=bashstyle]
#forward map
fnirt --ref=fixed_img --in=mov_img --iout=forward --cout=forward_coef \
    --intmod=global_non_linear
#inverse map
invwarp --ref=mov_img --warp=foward --out=inverse
\end{lstlisting}
\end{itemize}

\subsection{Necessary Optimality Conditions for Constrained Formulation}\label{s:appendvpcont}

To derive the variational gradient of \cref{e:mse} with respect to control $\pmb{C}$, we need to first derive the variational gradient of \cref{e:mse} with respect to $\pmb{\phi}$. We have
\begin{equation}\label{e:msedel}
\delta_{\pmb{\phi}}\text{MSE}(\pmb{\phi})
=
\frac{\text{d}}{\text{d}\epsilon} \text{MSE}(\pmb{\phi}+\epsilon\delta\pmb{\phi})\biggr\rvert_{\epsilon=0}
=
\dfrac{1}{|\Omega|}\int_{\Omega} (M(\pmb{\phi}) - F) \nabla M(\pmb{\phi}) \cdot\delta \pmb{\phi}\,\text{d}\pmb{\omega},
\end{equation}

\noindent which implies that
\[
\frac{\partial \text{MSE}}{\partial \pmb{\phi}} = (M(\pmb{\phi}) - F) \nabla M(\pmb{\phi}).
\]

The variational gradient of MSE in \cref{e:mse} with respect to $\pmb{C}$ can be derived based on \cref{e:mseC}. Since the partial derivatives are interchangeable with variations in the sense of variational calculus, we have $\delta \pmb{C} = \delta \Delta\pmb{\phi}$. Since images $M$ and $F$ are assumed to be sufficiently smooth we can stipulate $\pmb{b}$ satisfies $\Delta\pmb{b}=(M(\pmb{\phi}) - F)\nabla M(\pmb{\phi})$. Assuming that $\delta\pmb{C}$ vanishes on $\partial\Omega$ we can replace $(M(\pmb{\phi}) - F)\nabla M(\pmb{\phi})$ in \cref{e:msedel} by $\Delta\pmb{b}$ to obtain
\[
\delta_{\pmb{\phi}} \text{MSE}(\pmb{\phi})
= \dfrac{1}{|\Omega|}\int_{\Omega} \Delta\pmb{b} \cdot \delta \pmb{\phi}\,\text{d}\pmb{\omega}.
\]

\noindent Since $\pmb{\phi}$ is modeled as $\pmb{\operatorname{id}}_{\mathrm{\Omega}} + \pmb{u}$ and $\delta \pmb{\phi} = \pmb{0}$ on $\partial\Omega$, we have $\delta\pmb{\phi} = \delta(\pmb{\operatorname{id}}_{\mathrm{\Omega}} + \pmb{u}) = \delta\pmb{u} = \pmb{0}$. Therefore, by \emph{Green's identities} with fixed boundary conditions we obtain
\begin{equation}\label{e:msedelc}
\delta_{\pmb{\phi}} \text{MSE}(\pmb{\phi})
=\dfrac{1}{|\Omega|}\int_{\Omega} \pmb{b} \cdot \Delta \delta \pmb{\phi}\,\text{d}\pmb{\omega}
=\dfrac{1}{|\Omega|}\int_{\Omega} \pmb{b} \cdot \delta \Delta \pmb{\phi}\,\text{d}\pmb{\omega}
=\dfrac{1}{|\Omega|}\int_{\Omega} \pmb{b} \cdot \delta \pmb{C}\,\text{d}\pmb{\omega}=\delta_{\pmb{C}} \text{MSE},
\end{equation}

\noindent which implies that
\begin{equation}\label{e:msedelc2}
\frac{\partial \text{MSE}}{\partial \pmb{C}}=\pmb{b}.
\end{equation}

This trick also allows for the derivation of the partial gradients of MSE \cref{e:mse} with respect to control functions $f$ and $\pmb{g}$. Using \cref{e:mseC} we obtain $\delta \pmb{C} = \delta(\nabla f-\nabla \times \pmb{g})$. Inserting this expression into \cref{e:msedelc} yields
\begin{equation}
\delta_{\pmb{C}} \text{MSE}
= \int_{\Omega} \pmb{b} \cdot \delta \pmb{C}\,\text{d}\pmb{\omega}
= \int_{\Omega} \pmb{b} \cdot \delta( \nabla  f -\nabla \times  \pmb{g}) \,\text{d}\pmb{\omega}
= \int_{\Omega} \pmb{b} \cdot  \nabla \delta f \,\text{d}\pmb{\omega}
- \int_{\Omega} \pmb{b} \cdot\nabla \times \delta \pmb{g} \,\text{d}\pmb{\omega}.
\end{equation}

Applying \emph{Green's identities} with fixed boundary conditions yields
\begin{equation}
  \int_{\Omega} - \nabla \cdot\pmb{b}\delta f \,\text{d}\pmb{\omega}
+ \int_{\Omega} \nabla \times \pmb{b}\cdot \delta \pmb{g}\,\text{d}\pmb{\omega}
= \delta_{f} \text{MSE} + \delta_{\pmb{g}} \text{MSE}.
\end{equation}

\noindent It follows that
\begin{equation}\label{e:msedelfg}
\frac{\partial \text{MSE}}{\partial f}=-\nabla\cdot\pmb{b}
\quad\text{and}\quad
\frac{\partial \text{MSE}}{\partial \pmb{g}}=\nabla \times \pmb{b}.
\end{equation}

We summarize the associated algorithm in \Cref{alg:vpcont}. The algorithm consists of two stages, one for the update associated with the auxiliary variable $\pmb{C}$ and one for the update associated with the controls $f$ and $\pmb{g}$. That is, we decompose the map $\pmb{\phi}$ into two maps: The map $\pmb{\phi}_{\text{global}}$ and $\pmb{\phi}_{\text{local}}$. The map $\pmb{\phi}_{\text{global}}$ is computed based on the iterative procedure associated with $\pmb{C}$. The map $\pmb{\phi}_{\text{local}}$ is associated with the iterative procedure for updating $f$ and $\pmb{g}$. The sought after map $\pmb{\phi}$ is given by the composition of $\pmb{\phi}_{\text{global}}$ and $\pmb{\phi}_{\text{local}}$, i.e., $\pmb{\phi} = \pmb{\phi}_{\text{global}} \circ \pmb{\phi}_{\text{local}}$. This algorithm has been implemented in \cite{zhou2022:recent}. We provide a justification for this two-stage structure in the main part of this manuscript (see \Cref{s:vpcontrol}).

We note that the algorithm requires several elliptic PDE solves, i.e., the inversion of $\Delta$. We do so using FFTs, i.e., we use a pseudo-spectral method with a Fourier basis. The step size of our gradient descent algorithm is denoted by $t$.

\begin{algorithm}
\caption{Diffeomorphic image registration based on two-stage algorithm VP-control implemented in \cite{zhou2022:recent}.}
\label{alg:vpcont}
\begin{algorithmic}[1]
\STATE {\bf input:} $M$, $F$
\STATE {\bf Stage-1: Global Stage}
\STATE {\bf initialize:} $\pmb{\phi}_{\text{global}} \gets \pmb{\operatorname{id}}_{\Omega}$, $\pmb{C} \gets \pmb{0}$, $t \gets 1$, better $\gets$ true, converged $\gets$ false
\WHILE{$\neg$ \text{converged}}
    \IF{\text{better}}
    \STATE compute $\pmb{b} = \frac{\partial \text{MSE}}{\partial\pmb{C}}$ by solving $\mathrm{\Delta}\pmb{b} = (M(\pmb{\phi}_{\text{global}}) - F) \nabla M(\pmb{\phi}_{\text{global}})$
    \STATE $\pmb{C}_{\text{new}} \gets \pmb{C}-t \, \pmb{b}$
    \ENDIF
    \STATE compute $\pmb{\phi}_{\text{temp}}$ by solving $\Delta\pmb{\phi}_{\text{temp}} = \pmb{C}_{\text{new}}$
    \STATE $\pmb{\phi}_{\text{new}} \gets \pmb{\phi}_{\text{temp}}(\pmb{\phi}_{\text{global}} )$
    \IF{$\text{MSE}(M,F,\pmb{\phi}_{\text{new}})$ in \eqref{e:mse} decreases}
        \STATE $\pmb{\phi}_\text{global} \gets \pmb{\phi}_{\text{new}}\quad and \quad \text{better} \gets \text{true} \quad and \quad \text{increase } t  $
        \STATE $\pmb{C} \gets \pmb{C}_{\text{new}}$
    \ELSE
        \STATE reduce $t$
    \ENDIF
\ENDWHILE
\STATE {\bf Stage-2: Local Stage}
\STATE {\bf initialize:} $\pmb{\phi}_\text{local}\gets\pmb{\operatorname{id}}_{\Omega}$, $f\gets 1$, $\pmb{g}\gets\pmb{0}$, $t \gets 1$, better $\gets$ true, $M_{\text{global}} \gets M(\pmb{\phi}_{\text{global}})$ converged $\gets$ false
\WHILE{$\neg$ \text{converged}}
    \IF{\text{better}}
        \STATE compute $\pmb{b}$ by solving $\Delta\pmb{b} = (M_{\text{global}}(\pmb{\phi}_\text{local}) - F) \nabla M_{\text{global}}(\pmb{\phi}_\text{local})$
        \STATE compute $\frac{\partial \text{MSE}}{\partial f}$ and $\frac{\partial \text{MSE}}{\partial\pmb{g}}$ from $\pmb{b}$ via \cref{e:msedelfg}
        \STATE $f_{\text{new}} \gets f -  t\,\frac{\partial\text{MSE}}{\partial f}$ and $\pmb{g}_{\text{new}} \gets \pmb{g} - t \, \frac{\partial\text{MSE}}{\partial\pmb{g}}$
    \ENDIF
    \STATE compute $\pmb{\phi}_{\text{new}}$ by solving $\Delta\pmb{\phi}_{\text{new}} = \nabla\cdot f_{\text{new}} -  \nabla\times \pmb{g}_{\text{new}}$
    \STATE $\pmb{\phi}_\text{new} \gets \pmb{\phi}_{\text{new}}(\pmb{\phi}_\text{local})$
    \IF{$\text{MSE}(M_{\text{global}},F,\pmb{\phi}_{\text{temp}})$ in \cref{e:mse} decreases}
        \STATE $\pmb{\phi}_{\text{local}} \gets \pmb{\phi}_{\text{new}} \quad and \quad \text{better} \gets \text{true} \quad and \quad \text{increase } t$
        \STATE  $ f \gets f_{\text{new}} \quad and \quad \pmb{g} \gets \pmb{g}_{\text{new}}$
    \ELSE
        \STATE reduce $t$
    \ENDIF
\ENDWHILE
\STATE $M_{\pmb{\phi}} \gets$ interpolate $M \circ \pmb{\phi}$
\STATE $\pmb{\phi} \gets \pmb{\phi}_{\text{global}}\circ \pmb{\phi}_{\text{local}}$
\STATE {\bf output:} $\pmb{\phi}$ $and$ $M_{\pmb{\phi}}$
\end{algorithmic}
\end{algorithm}

\subsection{The Euler-Lagrange Equations for Penalty Formulation}\label{s:appendvpregel}

Next, we are deriving the Euler-Lagrange equation for \cref{e:vpregeltarget}. First, we compute variations with respect to $f_t$ and $g_t$ in the directions $\delta f_t$ and $\delta \pmb{g}_t$, respectively. We obtain
\begin{equation}\label{e:var-soft-penalty}
\begin{aligned}
\delta_{f_t} \mathcal{V}(\pmb{\phi},f,\pmb{g},f_t,\pmb{g}_t)
& = \int_{\Omega} \delta f_t (\nabla\cdot\pmb{\phi} - f_t - 2 - f_t + f)\,\text{d}\pmb{\omega},\\
\delta_{\pmb{g}_t} \mathcal{V}(\pmb{\phi},f,\pmb{g},f_t,\pmb{g}_t)
& = \int_{\Omega} \delta \pmb{g}_t \cdot(\nabla \times\pmb{\phi} - \pmb{g}_t-\pmb{g}_t+\pmb{g})\,\text{d}\pmb{\omega},
\end{aligned}
\end{equation}

\noindent for all $\delta f_t$ and $\delta \pmb{g}_t$ on $\Omega$. The strong form of the optimality conditions are given by
\begin{equation}\label{e:optimalityeq}
f_t = \frac{1}{2}(\nabla\cdot\pmb{\phi}-f-2)
\quad\text{and}\quad
\pmb{g}_t = \frac{1}{2}(\nabla \times\pmb{\phi}-\pmb{g}),
\end{equation}

\noindent respectively. Since the optimality equations \cref{e:optimalityeq} are necessary conditions for $\mathcal{V}$ to be optimal, we can substitute \cref{e:optimalityeq} back into \cref{e:mse} to eliminate $f_t$ and $\pmb{g}_t$ from \cref{e:mse}. This leads to the reduced form
\begin{equation}\label{e:vpcontrolel}
\mathcal{V}(\pmb{\phi},f,\pmb{g}) = \text{MSE}(\pmb{\phi})
+\frac{1}{2}\int_{\Omega}(\nabla\cdot\pmb{\phi} - f-2)^{2}\,\mathrm{d}\pmb{\omega}
+\frac{1}{2}\int_{\Omega} \|\nabla \times\pmb{\phi} - \pmb{g}\|^{2}\,\mathrm{d}\pmb{\omega}.
\end{equation}

\noindent With this the unknown target functions $f_t$ and $\pmb{g}_t$ become implicit as the optimality equations \cref{e:optimalityeq} are part of the necessary conditions. Next, we derive the Euler-Lagrange equations of \cref{e:vpcontrolel}. We formally compute variations with respect to $\pmb{\phi} \in \operatorname{Sol}(\Omega)$ in the direction of $\delta\pmb{\phi} \in \operatorname{Sol}(\Omega)$. We have
\begin{equation}\label{e:variations-soft-const}
	\begin{aligned}
		\delta_{\pmb{\phi}}\mathcal{V}(\pmb{\phi},f,\pmb{g})
		&= \frac{1}{|\Omega|}
		\int_{\Omega}(M(\pmb{\phi}) - F) \nabla M(\pmb{\phi})\cdot \delta\pmb{\phi}\,\text{d}\pmb{\omega}+\int_{\Omega} (\nabla\cdot\pmb{\phi} - f-2) (\nabla\cdot\delta\pmb{\phi})\,\text{d}\pmb{\omega} \\
		& \qquad +\int_{\Omega}(\nabla \times\pmb{\phi}- \pmb{g})\cdot(\nabla \times\delta\pmb{\phi})\,\text{d}\pmb{\omega}.
	\end{aligned}
\end{equation}

Next, we apply Green's identities to the second and third term. For the second term, we have
\begin{equation*}
\begin{aligned}
& \int_{\Omega} [(\nabla\cdot\pmb{\phi} - f -2)\nabla\cdot\delta\pmb{\phi}]\,\text{d}\pmb{\omega} = \int_{\Omega}(\nabla\cdot\pmb{\phi} - f+2)(\delta\phi_{1x} + \delta\phi_{2y} + \delta\phi_{3z}) \,\text{d}\pmb{\omega}\\
&=\int_{\mathrm{\Omega}}
\begin{pmatrix}
\nabla\cdot\pmb{\phi} - f-2\\ 0 \\ 0
\end{pmatrix}
\!\cdot\!\nabla \delta \phi_{1}
+
\begin{pmatrix}
0 \\ \nabla\cdot\pmb{\phi} - f-2 \\ 0
\end{pmatrix}
\!\cdot\!\nabla \delta \phi_{2}
+
\begin{pmatrix}
0 \\ 0 \\ \nabla\cdot\pmb{\phi} - f-2
\end{pmatrix}
\!\cdot\! \nabla \delta \phi_{3}\,\text{d}\pmb{\omega}\\
&=-\int_{\Omega}
\begin{pmatrix}
\phi_{1xx}+\phi_{2yx}+\phi_{3zx}- f_x \\
\phi_{1xy}+\phi_{2yy}+\phi_{3zy}- f_y \\
\phi_{1xz}+\phi_{2yz}+\phi_{3zz}- f_z \end{pmatrix}\cdot\delta \pmb{\phi}\,\text{d}\pmb{\omega}.
\end{aligned}
\end{equation*}

For the third term, we obtain
\begin{equation*}
\begin{aligned}
&\int_{\Omega}(\nabla \times\pmb{\phi} - \pmb{g})\cdot\nabla \times\delta\pmb{\phi}\,\text{d}\pmb{\omega}
=\int_{\Omega}\left(\begin{pmatrix}
	\phi_{3y}-\phi_{2z}\\
	-\phi_{3x}+\phi_{1z} \\
	\phi_{2x}-\phi_{1y}
\end{pmatrix} - \pmb{g}\right)
\cdot
\begin{pmatrix}
	\delta\phi_{3y}-\delta\phi_{2z}\\
	-\delta\phi_{3x}+\delta\phi_{1z} \\
	\delta\phi_{2x}-\delta\phi_{1y}
\end{pmatrix}\,\text{d}\pmb{\omega}
\\
&=\int_{\Omega}
\!\begin{pmatrix}
	0 \\
	-\phi_{2x} +\phi_{1y}+g_3 \\
	-\phi_{3x} +\phi_{1z} -g_2
\end{pmatrix}
\!\cdot\!\nabla\delta\phi_{1}
+
\!\begin{pmatrix}
	\phi_{2x}-\phi_{1y}-g_3 \\
	0 \\
	-\phi_{3y}+\phi_{2z}+g_1
\end{pmatrix}\!\cdot\!\nabla \delta \phi_{2}
+
\!\begin{pmatrix}
	\phi_{3x}-\phi_{1z}+g_2 \\
	\phi_{3y}-\phi_{2z}-g_1 \\
	0
\end{pmatrix}\!\cdot\!\nabla \delta \phi_{3}\,\text{d}\pmb{\omega}
\\
&=-\int_{\Omega}\begin{pmatrix}
	-\phi_{2xy} +\phi_{1yy} +g_{3y} -\phi_{3xz} +\phi_{1zz} -g_{2z}\\
	\phi_{2xx}-\phi_{1yx}-g_{3x} -\phi_{3yz}+\phi_{2zz}+g_{1z}\\
	\phi_{3xx}-\phi_{1zx}+g_{2x}+\phi_{3yy}-\phi_{2zy}-g_{1y}
\end{pmatrix}\cdot \delta\pmb{\phi}\,\text{d}\pmb{\omega}.
\end{aligned}
\end{equation*}

\noindent Adding up the resulting expressions for the second and third term, we obtain
\begin{equation*}
\begin{aligned}
&-\int_{\Omega}
\left(
\begin{pmatrix}
	-\phi_{2xy} +\phi_{1yy} +g_{3y} -\phi_{3xz} +\phi_{1zz} -g_{2z}\\
	\phi_{2xx}-\phi_{1yx}-g_{3x} -\phi_{3yz}+\phi_{2zz}+g_{1z}\\
	\phi_{3xx}-\phi_{1zx}+g_{2x}+\phi_{3yy}-\phi_{2zy}-g_{1y}
\end{pmatrix}
+
\begin{pmatrix}
	\phi_{1xx}+\phi_{2yx}+\phi_{3zx}- f_x \\
	\phi_{1xy}+\phi_{2yy}+\phi_{3zy}- f_y \\
	\phi_{1xz}+\phi_{2yz}+\phi_{3zz}- f_z \end{pmatrix}
\right)
\cdot\delta \pmb{\phi}
\,\text{d}\pmb{\omega}\\
&=-\int_{\Omega}
\begin{pmatrix}
	\phi_{1xx}+\phi_{2yx}+\phi_{3zx}- f_x -\phi_{2xy} +\phi_{1yy} +g_{3y} -\phi_{3xz} +\phi_{1zz} -g_{2z}\\
	\phi_{1xy}+\phi_{2yy}+\phi_{3zy}- f_y +\phi_{2xx}-\phi_{1yx}-g_{3x} -\phi_{3yz}+\phi_{2zz}+g_{1z} \\
	\phi_{1xz}+\phi_{2yz}+\phi_{3zz}- f_z +\phi_{3xx}-\phi_{1zx}+g_{2x}+\phi_{3yy}-\phi_{2zy}-g_{1y}\end{pmatrix}\cdot\delta \pmb{\phi}\,\text{d}\pmb{\omega}\\
&=\int_{\Omega}
\left(-\begin{pmatrix}
	\phi_{1xx}+\phi_{1yy} +\phi_{1zz} \\
	\phi_{2xx}+\phi_{2yy}+\phi_{2zz} \\
	\phi_{3xx}+\phi_{3yy}+\phi_{3zz}
\end{pmatrix}
+
\begin{pmatrix}
	f_x\\
	f_y \\
	f_z
\end{pmatrix}
-
\begin{pmatrix}
	g_{3y} -g_{2z}\\
	g_{1z} -g_{3x}\\
	g_{2x}-g_{1y}
\end{pmatrix}\right)\cdot\delta \pmb{\phi}\,\text{d}\pmb{\omega}
\\
&=\int_{\Omega} (-\Delta \pmb{\phi} + \nabla f -\nabla\times \pmb{g})
\cdot\delta \pmb{\phi}\,\text{d}\pmb{\omega}.
\end{aligned}
\end{equation*}

\noindent Combining this result with the first term in \cref{e:variations-soft-const} yields
\begin{equation}
\delta_{\pmb{\phi}}\mathcal{V}(\pmb{\phi},f,\pmb{g})=\int_{\Omega}
\left(
(M(\pmb{\phi}) - F) \nabla M(\pmb{\phi})
-\Delta \pmb{\phi} + \nabla f -\nabla\times \pmb{g}
\right)\cdot\delta \pmb{\phi}\,\text{d}\pmb{\omega}
\end{equation}

\noindent for all $\delta \pmb{\phi} \in \operatorname{Sol}(\Omega)$. It follows that
\begin{equation}\label{e:vpconteldel}
\frac{\partial\mathcal{V}}{\partial\pmb{\phi}} =
(M(\pmb{\phi}) - F) \nabla M(\pmb{\phi})
-\Delta \pmb{\phi} + \nabla f -\nabla\times \pmb{g}.
\end{equation}

Setting the expression in \cref{e:vpconteldel} to zero, we obtain the Euler-Lagrange equation for $\mathcal{V}$ in~\cref{e:vpcontrolel} given by
\begin{equation}\label{e:vpconteleq}
\Delta \pmb{\phi} = (M(\pmb{\phi}) - F) \nabla M(\pmb{\phi}) + \nabla f -\nabla\times \pmb{g}.
\end{equation}

This system is a fixed point equation; the non-linearity appears on the right hand side. A straightforward strategy to compute a minimizer $\pmb{\phi}^\star$ that satisfies \cref{e:vpconteleq} is based on the iterative fixed-point scheme
\begin{equation}\label{e:fixed-point}
\pmb{\phi}^{(k+1)} = \Delta^{-1}\left((M(\pmb{\phi}^{(k)}) - F) \nabla M(\pmb{\phi}^{(k)}) + \nabla f^{(k)} -\nabla\times \pmb{g}^{(k)}\right)
\end{equation}

\noindent for given trial controls $f^{(k)}$ and $\pmb{g}^{(k)}$ at iteration $k \in \mathbb{N}_0$ with initial guess $\pmb{\phi}^{(0)}=\pmb{\operatorname{id}}_{\Omega}$. To find $f^{(k)}$ and $\pmb{g}^{(k)}$ we require additional equations. The proposed algorithm uses the original definition of the control functions $f$ and $\pmb{g}$. That is, we set $f^{(k)}$ and $\pmb{g}^{(k)}$ to $f^{(k)} = \det \nabla \pmb{\phi}^{(k)}$ and $\pmb{g}^{(k)} = \nabla \times \pmb{\phi}^{(k)}$, respectively. We note that, for $\pmb{\phi}^{(0)}=\pmb{\operatorname{id}}_{\Omega}$, we have $f^{(0)} = \det \nabla \pmb{\operatorname{id}}_{\Omega} = 1$ and  $\pmb{g}^{(k)} = \nabla \times \pmb{\operatorname{id}}_{\Omega} = \pmb{0}$, respectively; both satisfy the conditions in \cref{e:fg}. Instead of directly updating $\pmb{\phi}^{(k)}$ by solving \cref{e:fixed-point}, we use a convex combination of the solution of \cref{e:fixed-point} and the current iterate $\pmb{\phi}^{(k)}$ to find $\pmb{\phi}^{(k+1)}$. That is
\begin{equation}\label{e:homo}
\pmb{\phi}^{(k+1)}= (1-\tau)\,\pmb{\phi}^{(k)} + \tau\,\pmb{\phi}_{\text{temp}},
\end{equation}

\noindent where $\pmb{\phi}_{\text{temp}}$ solves \eqref{e:fixed-point} with parameter $\tau \in (0,1]$. The homotopy structure with $\tau$ prevents the transformation being far from $\pmb{\operatorname{id}}_{\Omega}$ so that $\pmb{\phi}^{(k+1)}$ is more likely to remain diffeomorphic. We summarize the resulting scheme in~\Cref{alg:vpcontelalg}.

We note that the algorithm requires several elliptic PDE solves, i.e., the inversion of $\Delta$. We do so using a FFT (i.e., we discretize the Laplacian operator using a pseudo-spectral method with a Fourier basis). The step size of our gradient descent algorithm is denoted by $t$.

\begin{algorithm}
\caption{Diffeomorphic image registration based on two-stage algorithm based on the Euler-Lagrange equations.}
\label{alg:vpcontelalg}
\begin{algorithmic}[1]
\STATE {\bf input:} $M$, $F$
\STATE {\bf Stage-1: Global Stage}
\STATE {\bf initialize:} $\pmb{\phi}_{\text{global}} \gets \pmb{\operatorname{id}}_{\Omega}$, $f=1$, $\pmb{g}\gets\pmb{0}$, better $\gets$ true
\WHILE{\text{better}}
    \STATE $f\gets \det\nabla\pmb{\phi}_{\text{global}}$ and $\pmb{g} \gets \nabla\times\pmb{\phi}_{\text{global}}$
    \STATE compute $\pmb{\phi}_{\text{new}}$ by solving $\Delta\pmb{\phi}_{\text{new}}=(M(\pmb{\phi}_{global})-F)\nabla M(\pmb{\phi}_\text{global}) + \nabla f-\nabla\times\pmb{g}$
    \STATE $\pmb{\phi}_{\text{temp}}\gets (1-\tau)\pmb{\phi}_\text{global} + \tau\pmb{\phi}_{\text{new}}$
    \IF{$\text{MSE}(M,F,\pmb{\phi}_{\text{temp}})$ in \eqref{e:mse} decreases}
        \STATE $\pmb{\phi}_\text{global} \gets \pmb{\phi}_{\text{temp}}$ \quad and \quad better $\gets$ true \quad and \quad increase $\tau$
    \ELSE
        \STATE better $\gets$ false
    \ENDIF
\ENDWHILE
\STATE {\bf Stage-2: Local Stage}
\STATE {\bf initialize:} $\pmb{\phi}_\text{local}\gets\pmb{\operatorname{id}}_{\Omega}$, $f\gets 1$, $\pmb{g}\gets\pmb{0}$, $t \gets 1$, better $\gets$ true, $M_{\text{global}} \gets M(\pmb{\phi}_{\text{global}})$ converged $\gets$ false
\WHILE{$\neg$ \text{converged}}
    \IF{\text{better}}
        \STATE compute $\pmb{b}$ by solving $\Delta\pmb{b} = (M_{\text{global}}(\pmb{\phi}_\text{local}) - F) \nabla M_{\text{global}}(\pmb{\phi}_\text{local})$
        \STATE compute $\frac{\partial \text{MSE}}{\partial f}$ and $\frac{\partial \text{MSE}}{\partial\pmb{g}}$ from $\pmb{b}$ via \cref{e:msedelfg}
        \STATE $f_{\text{new}} \gets f -  t\,\frac{\partial\text{MSE}}{\partial f}$ and $\pmb{g}_{\text{new}} \gets \pmb{g} - t \, \frac{\partial\text{MSE}}{\partial\pmb{g}}$
    \ENDIF
    \STATE compute $\pmb{\phi}_{\text{new}}$ by solving $\Delta\pmb{\phi}_{\text{new}} = \nabla\cdot f_{\text{new}} -  \nabla\times \pmb{g}_{\text{new}}$
    \STATE $\pmb{\phi}_\text{new} \gets \pmb{\phi}_{\text{new}}(\pmb{\phi}_\text{local})$
    \IF{$\text{MSE}(M_{\text{global}},F,\pmb{\phi}_{\text{temp}})$ in \cref{e:mse} decreases}
        \STATE $\pmb{\phi}_{\text{local}} \gets \pmb{\phi}_{\text{new}} \quad and \quad \text{better} \gets \text{true} \quad and \quad \text{increase } t$
        \STATE  $ f \gets f_{\text{new}} \quad and \quad \pmb{g} \gets \pmb{g}_{\text{new}}$
    \ELSE
        \STATE reduce $t$
    \ENDIF
\ENDWHILE
\STATE $M_{\pmb{\phi}} \gets$ interpolate $M \circ \pmb{\phi}$
\STATE $\pmb{\phi} \gets \pmb{\phi}_{\text{local}}\circ \pmb{\phi}_{\text{global}}$
\STATE {\bf output:} $\pmb{\phi}$ $and$ $M_{\pmb{\phi}}$
\end{algorithmic}
\end{algorithm}

\subsection{Necessary Optimality Conditions for Grid Generation}\label{s:appendlmvp}

To obtain the necessary conditions for minimizing \cref{e:lmvp}, we require the first variations of $\mathcal{L}$ in \cref{e:lmvp} with respect to the Lagrange multipliers $\delta\lambda_{f}$, $\delta\pmb{\lambda}_{\pmb{g}}$, the control functions $\delta f$, $\delta\pmb{g}$, and the diffeomorphic transformation $\delta\pmb{\phi}$. The first variations of $\mathcal{L}$ in the direction of $\delta \lambda_{f}$ are given by
\begin{equation*}
\delta_{\lambda_{f}} \mathcal{L}
= \frac{\mathrm{d}}{\mathrm{d}\epsilon} \mathcal{L}(\lambda_{f}+\epsilon\delta\lambda_{f})\biggr\rvert_{\epsilon=0}
= \int_{\Omega} \delta\lambda_{f} (\det\nabla\pmb{\phi} - f)\,\mathrm{d}\pmb{\omega}
\end{equation*}

\noindent for all $\delta\lambda_{f}$ on $\Omega$. This implies that $\frac{\partial \mathcal{L}}{\partial\lambda_{f}} = \det\nabla\pmb{\phi} - f$.

The first variations of $\mathcal{L}$ in the direction of $\delta \pmb{\lambda}_{\pmb{g}}$ are given by
\begin{equation*}
\delta_{\pmb{\lambda}_{\pmb{g}}} \mathcal{L}
= \frac{\mathrm{d}}{\mathrm{d}\epsilon} \biggr\rvert_{\epsilon=0}\mathcal{L}(\pmb{\lambda}_{\pmb{g}}+\epsilon\delta\pmb{\lambda}_{\pmb{g}})
= \int_{\Omega}\delta\pmb{\lambda}_{\pmb{g}}\cdot(\nabla \times\pmb{\phi}- \pmb{g}) \,\mathrm{d}\pmb{\omega},
\end{equation*}

\noindent for all $\delta\pmb{\lambda}_{\pmb{g}}$ on $\Omega$, which implies
\[
\frac{\partial\mathcal{L}}{\partial\pmb{\lambda}_{\pmb{g}}}= ((\phi_{3y} - \phi_{2z} - g_{1})  , ( -\phi_{3x} + \phi_{1z}  - g_{2}), ( \phi_{2x} - \phi_{1y} -g_{3})) = \nabla \times\pmb{\phi}- \pmb{g}.
\]

Setting these variations to zero we obtain the state equations
\begin{equation}\label{e:lmvpstateeq}
f =\det\nabla\pmb{\phi}
\quad \text{and} \quad
\pmb{g} = \nabla \times\pmb{\phi}.
\end{equation}

Next, we provide the first variations with respect to the control variables $f$ and $\pmb{g}$. We have
\begin{equation*}
\delta_{f} \mathcal{L}
= \frac{\mathrm{d}}{\mathrm{d}\epsilon} \mathcal{L}(f+\epsilon\delta f)\biggr\rvert_{\epsilon=0}
= \int_{\Omega} ( -\lambda_{f} + f)\,\delta f\,\mathrm{d}\pmb{\omega}
\end{equation*}

\noindent for all $\delta f$ on $\Omega$, which implies $\frac{\partial \mathcal{L}}{\partial f} = -\lambda_{f} + f$. Moreover,
\begin{equation*}
\delta_{\pmb{g}} \mathcal{L}
= \frac{d}{d\epsilon} \mathcal{L}(\pmb{g}+\epsilon\delta\pmb{g})\biggr\rvert_{\epsilon=0}
= \int_{\Omega} \delta\pmb{g}\cdot(-\pmb{\lambda}_{\pmb{g}} + \pmb{g})\, \mathrm{d}\pmb{\omega},
\end{equation*}

\noindent for all $\delta\pmb{g}$ on $\Omega$. This implies
\[
\frac{\partial \mathcal{L}}{\partial\pmb{g}} = (-\lambda_{1}+ g_{1},-\lambda_{2}+ g_{2},-\lambda_{3}+ g_{3}) = -\pmb{\lambda}_{\pmb{g}} + \pmb{g}.
\]

\noindent Setting these first variations to zero yields the control equations $\lambda_{f} = f$ and $\pmb{\lambda}_{\pmb{g}} = \pmb{g}$, respectively.

Lastly, we compute variations with respect to the unknown $\pmb{\phi}_{m}\in \operatorname{Sol}(\Omega)$. We have
\begin{equation}\label{e:variations_phi_m}
\begin{aligned}
\delta_{\pmb{\phi}_{m}} \mathcal{L}
& =
\frac{\mathrm{d}}{\mathrm{d}\epsilon}\mathcal{L}(\pmb{\phi}_{m}+\epsilon\delta\pmb{\phi}_{m}) \biggr\rvert_{\epsilon=0}\\
&= \int_{\Omega}(\pmb{\phi}_{m}\circ\pmb{\phi}_{o}-\pmb{\phi}_{t}) \cdot \delta\pmb{\phi}_{m}\,\mathrm{d}\pmb{\omega}
+ \int_{\Omega} \lambda_{f}\delta\det\nabla (\pmb{\phi}_{m} \circ \pmb{\phi}_{o})\,\mathrm{d}\pmb{\omega}\\
&\quad\quad
+ \int_{\Omega}\pmb{\lambda}_{\pmb{g}}\cdot\delta\nabla \times(\pmb{\phi}_{m} \circ \pmb{\phi}_{o})\,\mathrm{d}\pmb{\omega}.
\end{aligned}
\end{equation}

The second integral in \cref{e:variations_phi_m} yields
\begin{equation*}
\begin{aligned}
&\int_{\Omega} \lambda_{f}\delta\det\nabla(\pmb{\phi}) \,\mathrm{d}\pmb{\omega}
= \int_{\Omega} \lambda_{f}\delta(\det\nabla\pmb{\phi}_{m} \det\nabla\pmb{\phi}_{o})\,\mathrm{d}\pmb{\omega}
= \int_{\Omega} \lambda_{f}\det\nabla\pmb{\phi}_{o} \delta\det\nabla\pmb{\phi}_{m}\,\mathrm{d}\pmb{\omega}\\
&\quad=\int_{\Omega}
\Big[\lambda_{f}\det\nabla\pmb{\phi}_{o}\,\delta\Big(\phi_{m1x}(\phi_{m2y}\phi_{m3z} -\phi_{m2z}\phi_{m3y})-\phi_{m1y}(\phi_{m2x}\phi_{m3z} -\phi_{m2z}\phi_{m3x})\\
&\qquad\qquad+\phi_{m1z}(\phi_{m2x}\phi_{m3y} -\phi_{m2y}\phi_{m3x})\Big)\Big]\,\mathrm{d}\pmb{\omega}
\\
&\quad=\int_{\Omega}
\Big[\lambda_{f}\det\nabla\pmb{\phi}_{o}
\Big(\delta \phi_{m1x}\phi_{m2y}\phi_{m3z}+ \phi_{m1x}\delta \phi_{m2y}\phi_{m3z}+\phi_{m1x}\phi_{m2y}\delta \phi_{m3z}\\
&\qquad\qquad-\delta\phi_{m1x}\phi_{m2z}\phi_{m3y}-\phi_{m1x}\delta\phi_{m2z}\phi_{m3y}-\phi_{m1x}\phi_{m2z}\delta\phi_{m3y}-\delta\phi_{m1y}\phi_{m3z}\phi_{m2x}\\
&\qquad\qquad-\phi_{m1y}\delta \phi_{m3z}\phi_{m2x}-\phi_{m1y}\phi_{m3z}\delta \phi_{m2x}+\delta\phi_{m1y} \phi_{m3x}\phi_{m2z} + \phi_{m1y}\delta\phi_{m3x}\phi_{m2z} \\
&\qquad\qquad+\phi_{m1y}\phi_{m3x}\delta \phi_{m2z}+\delta \phi_{m1z}\phi_{m2x}\phi_{m3y}+\phi_{m1z}\delta \phi_{m2x}\phi_{m3y} + \phi_{m1z}\phi_{m2x}\delta \phi_{m3y}\\
&\qquad\qquad-\delta\phi_{m1z}\phi_{m2y}\phi_{m3x}-\phi_{m1z}\delta \phi_{m2y}\phi_{m3x}-\phi_{m1z}\phi_{m2y}\delta \phi_{m3x}\Big)\Big]\,\mathrm{d}\pmb{\omega}
\\
&\quad=\int_{\Omega}\lambda_{f}\det\nabla\pmb{\phi}_{o}\Big(
    \begin{pmatrix}
        \phi_{m2y}\phi_{m3z}-\phi_{m3y}\phi_{m2z} \\
        \phi_{m3x}\phi_{m2z}-\phi_{m2x}\phi_{m3z} \\
        \phi_{m2x}\phi_{m3y}-\phi_{m2y}\phi_{m3x}
    \end{pmatrix} \cdot  \nabla \delta \phi_{m1} \\
&\qquad\qquad
    +\begin{pmatrix}
    \phi_{m3y}\phi_{m1z}-\phi_{m1y}\phi_{m3z} \\
    \phi_{m1x}\phi_{m3z}-\phi_{m1z}\phi_{m3x} \\
    \phi_{m3x}\phi_{m1y}-\phi_{m1x}\phi_{m3y}
    \end{pmatrix}\cdot  \nabla \delta \phi_{m2}
    +\begin{pmatrix}
        \phi_{m1y}\phi_{m2z}-\phi_{m2y}\phi_{m1z} \\
        \phi_{m2x}\phi_{m1z}-\phi_{m1x}\phi_{m2z} \\
        \phi_{m1x}\phi_{m2y}-\phi_{m2x}\phi_{m1y}
    \end{pmatrix}\cdot \nabla \delta \phi_{m3}\Big)\,\mathrm{d}\pmb{\omega}.
\end{aligned}
\end{equation*}

Introducing the short hand notation $\pmb{v}_i$, $i = 1,2,3$, for the vectors that depend on derivatives of components of $\pmb{\phi}_m$ scaled by $\lambda_{f}\det\nabla\pmb{\phi}_{o}$ we obtain the final result
\begin{equation}\label{e:lmvp_second_term}
\int_{\Omega} \lambda_{f}\delta\det\nabla(\pmb{\phi}) \,\mathrm{d}\pmb{\omega} =
\int_{\Omega} (\pmb{v}_{1}\cdot\nabla \delta \phi_{m1}+\pmb{v}_{2}\cdot  \nabla \delta \phi_{m2}+\pmb{v}_{3}\cdot \nabla \delta \phi_{m3})\,\mathrm{d}\pmb{\omega}.
\end{equation}

The third integral in \cref{e:variations_phi_m} yields
\begin{equation}\label{e:lmvp_third_term}
\begin{aligned}
&\int_{\Omega}\pmb{\lambda}_{\pmb{g}}\cdot\delta\nabla \times(\pmb{\phi}_{m} \circ \pmb{\phi}_{o})\,\mathrm{d}\pmb{\omega}
=\int_{\mathrm{\Omega}} \pmb{\lambda}_{\pmb{g}}\cdot\delta
    \begin{pmatrix}
        \nabla\phi_{m3}\cdot(\pmb{\phi}_{o})_{y}-\nabla\phi_{m2}\cdot (\pmb{\phi}_{o})_{z}\\
        -\nabla\phi_{m3}\cdot(\pmb{\phi}_{o})_{x}+\nabla\phi_{m1}\cdot(\pmb{\phi}_{o})_{z} \\
        \nabla\phi_{m2}\cdot(\pmb{\phi}_{o})_{x}-\nabla\phi_{m1}\cdot(\pmb{\phi}_{o})_{y}
    \end{pmatrix}\,\mathrm{d}\pmb{\omega}
\\
&\quad = \int_{\Omega}
\begin{pmatrix}
    \lambda_{\pmb{g}1} \\
    \lambda_{\pmb{g}2}  \\
    \lambda_{\pmb{g}3}
\end{pmatrix}\cdot
\begin{pmatrix}
    \delta(\nabla\phi_{m3}\cdot(\pmb{\phi}_{o})_{y})-\delta(\nabla\phi_{m2}\cdot(\pmb{\phi}_{o})_{z})\\
    -\delta(\nabla\phi_{m3}\cdot(\pmb{\phi}_{o})_{x})+\delta(\nabla\phi_{m1}\cdot(\pmb{\phi}_{o})_{z}) \\
    \delta(\nabla\phi_{m2}\cdot(\pmb{\phi}_{o})_{x})-\delta(\nabla\phi_{m1}\cdot(\pmb{\phi}_{o})_{y})
\end{pmatrix}\,\mathrm{d}\pmb{\omega}
\\
&\quad=\int_{\Omega}
    \begin{pmatrix}
        \lambda_{\pmb{g}2}\phi_{o1z}-\lambda_{\pmb{g}3}\phi_{o1y} \\
        \lambda_{\pmb{g}2}\phi_{o2z}-\lambda_{\pmb{g}3}\phi_{o2y} \\
        \lambda_{\pmb{g}2}\phi_{o3z}-\lambda_{\pmb{g}3}\phi_{o3y}
    \end{pmatrix}
    \cdot
    \begin{pmatrix}
        \delta(\phi_{m1})_{x} \\
        \delta(\phi_{m1})_{y} \\
        \delta(\phi_{m1})_{z}
    \end{pmatrix}
    +
    \begin{pmatrix}
        -\lambda_{\pmb{g}1}\phi_{o1z}+ \lambda_{\pmb{g}3}\phi_{o1x} \\
        -\lambda_{\pmb{g}1}\phi_{o2z}+ \lambda_{\pmb{g}3}\phi_{o2x} \\
        -\lambda_{\pmb{g}1}\phi_{o3z}+ \lambda_{\pmb{g}3}\phi_{o3x}
    \end{pmatrix}
    \cdot
    \begin{pmatrix}
        \delta(\phi_{m2})_{x} \\
        \delta(\phi_{m2})_{y} \\
        \delta(\phi_{m2})_{z}
    \end{pmatrix}
\\
&\qquad\qquad+
    \begin{pmatrix}
        \lambda_{\pmb{g}1}\phi_{o1y}- \lambda_{\pmb{g}2}\phi_{o1x} \\
        \lambda_{\pmb{g}1}\phi_{o2y}- \lambda_{\pmb{g}2}\phi_{o2x} \\
        \lambda_{\pmb{g}1}\phi_{o3y}- \lambda_{\pmb{g}2}\phi_{o3x}
    \end{pmatrix}\cdot  \begin{pmatrix}
        \delta(\phi_{m3})_{x} \\
        \delta(\phi_{m3})_{y} \\
        \delta(\phi_{m3})_{z}
        \end{pmatrix}\,\mathrm{d}\pmb{\omega}
\\
&\quad =
    \int_{\Omega} \pmb{w}_{1}\cdot  \nabla \delta \phi_{m1}
            +\pmb{w}_{2}\cdot  \nabla \delta \phi_{m2}
            +\pmb{w}_{3}\cdot  \nabla \delta \phi_{m3}\,\mathrm{d}\pmb{\omega}
\end{aligned}
\end{equation}

\noindent with short hand notations $\pmb{w}_i$, $i = 1,2,3$, for the vectors that contain the partial derivatives of $\pmb{\phi}_o$ and the components of $\pmb{\lambda}_{\pmb{g}}$. Adding the expressions in \cref{e:lmvp_second_term} and \cref{e:lmvp_third_term} and applying \emph{Green's identities} we obtain
\begin{equation*}
\begin{aligned}
&\int_{\Omega}(\pmb{v}_{1} + \pmb{w}_{1})\cdot  \nabla \delta \phi_{m1} +(\pmb{v}_{2} + \pmb{w}_{2})\cdot  \nabla \delta \phi_{m2}+ (\pmb{v}_{3} + \pmb{w}_{3})\cdot  \nabla \delta \phi_{m3}\,\mathrm{d}\pmb{\omega}\\
&=-\int_{\mathrm{\Omega}} \nabla \cdot  (\pmb{v}_{1} + \pmb{w}_{1})\delta \phi_{m1} + \nabla \cdot  (\pmb{v}_{2} + \pmb{w}_{2})\delta \phi_{m2} + \nabla \cdot  (\pmb{v}_{3} + \pmb{w}_{3}) \delta \phi_{m3}\,\mathrm{d}\pmb{\omega}.\\
\end{aligned}
\end{equation*}

Combining all expressions derived above we arrive at
\[
\delta_{\pmb{\phi}_{m}} \mathcal{L}
= \int_{\Omega}\Big(
\pmb{\phi}_{m}\circ\pmb{\phi}_{o}-\pmb{\phi}_{t}
-
\begin{pmatrix}
\nabla \cdot (\pmb{v}_{1} + \pmb{w}_{1}) \\
\nabla \cdot (\pmb{v}_{2} + \pmb{w}_{2}) \\
\nabla \cdot (\pmb{v}_{3} + \pmb{w}_{3}) \\
\end{pmatrix}\Big)
\cdot \delta\pmb{\phi}_{m}\,\mathrm{d}\pmb{\omega}.
\]

Setting the variations to zero yields the strong form of the optimality conditions given by
\begin{equation}\label{e:lmvpdel}
\frac{\partial \mathcal{L}}{\partial\pmb{\phi}_{m}} = \pmb{\phi}_{m}\circ\pmb{\phi}_{o}-\pmb{\phi}_{t}
- \begin{pmatrix}
    \nabla \cdot  (\pmb{v}_{1} + \pmb{w}_{1}) \\
    \nabla \cdot  (\pmb{v}_{2} + \pmb{w}_{2}) \\
    \nabla \cdot  (\pmb{v}_{3} + \pmb{w}_{3})
\end{pmatrix} = \pmb{0}.
\end{equation}

\begin{algorithm}
\caption{Algorithm to compute the inverse of a given map $\pmb{\phi}$ modified from its general form introduced in \cite{zhou2024:construction}.}\label{alg:lmvpalg}
\begin{algorithmic}
    \STATE {\bf input:} $\pmb{\phi}_{t}\gets\pmb{\operatorname{id}}_{\Omega}$, $\pmb{\phi}_{o}\gets\pmb{\phi}$ ($\pmb{\phi}$ is the output from \Cref{alg:vpcontelalg})
    \STATE {\bf initialize:} $\pmb{\phi}_{m}=\pmb{\operatorname{id}}_{\Omega}$, better $\gets$ true
    \WHILE{$\neg$ \text{converged}}
    \IF{\text{better}}
        \STATE evaluate \cref{e:lmvpdel} to get $\frac{\partial\mathcal{L}}{\partial\pmb{\phi}_{m}}$
        \STATE $\pmb{\phi}_{\text{new}} \gets \pmb{\phi}_{m}- t\frac{\partial\mathcal{L}}{\partial\pmb{\phi}_{m}}$
        \STATE $\pmb{\phi}_{\text{temp}} \gets \pmb{\phi}_{\text{new}}(\pmb{\phi}_{o})$
    \ENDIF
    \IF{$\int_{\Omega} \|\pmb{\phi}_{\text{temp}}-\pmb{\phi}_{t}||^{2}\,\text{d}\pmb{\omega}$ in \cref{e:lmvp} decreases}
        \STATE $\pmb{\phi}_{m} \gets \pmb{\phi}_{\text{new}}$ \quad and \quad \text{better} $\gets$ true  \quad and increase $t$
    \ELSE
        \STATE reduce $t$
    \ENDIF
    \ENDWHILE
    \STATE {\bf output:} $\pmb{\phi}_{m}$ $and$ $\pmb{\phi}\gets\pmb{\phi}_{m}(\pmb{\phi}_{o})$
\end{algorithmic}
\end{algorithm}

\end{appendix}

\end{document}